\documentclass{article}

\usepackage{hyperref}

\usepackage[accepted]{mlsys2023}

\usepackage{latexsym}
\usepackage{amsmath}
\usepackage{amssymb}
\usepackage{ifthen}
\usepackage{enumitem}
\setlist{nosep}
\usepackage{graphicx}
\usepackage{listings}
\usepackage{booktabs}
\usepackage{subcaption}

\usepackage[dvipsnames]{xcolor}
\usepackage[T1]{fontenc}
\usepackage[scaled=0.85]{beramono}

\usepackage[utf8]{inputenc}

\usepackage{microtype}

\ifthenelse{\isundefined{\definition}}{\newtheorem{definition}{Definition}}{}
\ifthenelse{\isundefined{\assumption}}{\newtheorem{assumption}{Assumption}}{}
\ifthenelse{\isundefined{\hypothesis}}{\newtheorem{hypothesis}{Hypothesis}}{}
\ifthenelse{\isundefined{\proposition}}{\newtheorem{proposition}{Proposition}}{}
\ifthenelse{\isundefined{\theorem}}{\newtheorem{theorem}{Theorem}}{}
\ifthenelse{\isundefined{\lemma}}{\newtheorem{lemma}{Lemma}}{}
\ifthenelse{\isundefined{\corollary}}{\newtheorem{corollary}{Corollary}}{}
\ifthenelse{\isundefined{\alg}}{\newtheorem{alg}{Algorithm}}{}
\ifthenelse{\isundefined{\example}}{\newtheorem{example}{Example}}{}

\usepackage{xspace}

\newcommand\tldrDone[1]{}

\newcommand\gptdavinci{\texttt{OpenAI/davinci}\xspace}

\newcommand\coherexl{\texttt{Cohere/XL v20220609}\xspace}
\newcommand\anthropic{\texttt{Anthropic/v4-s3}\xspace}
\newcommand\mtnlg{\texttt{Microsoft+NV/TNLG v2}\xspace}
\newcommand\jurassicjumbo{\texttt{AI21/J1-Jumbo v1}\xspace}
\newcommand\jurassicgrande{\texttt{AI21/J1-Grande v1}\xspace}
\newcommand\jurassiclarge{\texttt{AI21/J1-Large v1}\xspace}

\newcommand\gptj{\texttt{EleutherAI/GPT-J}\xspace}
\newcommand\bloom{\texttt{BigScience/BLOOM}\xspace}
\newcommand\yalm{\texttt{Yandex/YaLM}\xspace}

\lstdefinelanguage{runtime_estimation}{
 keywords={output_generation_cost, prompt_encoding_cost},
 keywordstyle=\color{NavyBlue}\bfseries,
 ndkeywords={},
 ndkeywordstyle=\color{BrickRed}\bfseries,
 identifierstyle=\color{black},
 sensitive=false,
 comment=[l]{\/\/},
 morecomment=[s]{/*}{*/},
 commentstyle=\color{ForestGreen}\itshape\ttfamily,
 stringstyle=\color{red}\ttfamily,
}

\begin{document}

\twocolumn[
\mlsystitle{Cheaply Evaluating Inference Efficiency Metrics for Autoregressive Transformer APIs}

\begin{mlsysauthorlist}
\mlsysauthor{Deepak Narayanan}{msr}
\mlsysauthor{Keshav Santhanam}{stan}
\mlsysauthor{Peter Henderson}{stan}
\mlsysauthor{Rishi Bommasani}{stan}
\mlsysauthor{Tony Lee}{stan}
\mlsysauthor{Percy Liang}{stan}
\end{mlsysauthorlist}

\mlsysaffiliation{stan}{Stanford University}
\mlsysaffiliation{msr}{Microsoft Research}

\mlsyscorrespondingauthor{Deepak Narayanan}{dnarayanan@microsoft.com}

\vskip 0.3in

\begin{abstract}

Large language models (LLMs) power many state-of-the-art systems in natural language processing. However, these models are extremely computationally expensive, even at inference time, raising the natural question: when is the extra cost of deploying a larger model worth the anticipated boost in capabilities?
Better understanding this tradeoff fundamentally could benefit from an inference efficiency metric that is both (i) easily comparable across models from different providers, and (ii) representative of the true cost of running queries in an isolated performance environment.
Unfortunately, access to LLMs today is largely restricted to black-box text generation APIs and
raw runtimes measured through this interface do not satisfy these desiderata: model providers can apply various software and hardware optimizations orthogonal to the model, and models served on shared infrastructure are susceptible to performance contention. 
To circumvent these problems, we propose a new metric for comparing inference efficiency across models. This metric puts models on equal footing as though they were served (i) on uniform hardware and software, and (ii) without performance contention.
We call this metric the \emph{idealized runtime}, and we propose a methodology to efficiently estimate this metric for autoregressive Transformer models.
We also propose cost-aware variants that incorporate the number of accelerators needed to serve the model. 
Using these metrics, we compare ten state-of-the-art LLMs to provide the first analysis of inference efficiency-capability tradeoffs; we make several observations from this analysis, including the fact that the superior inference runtime performance of certain APIs is often a byproduct of optimizations within the API rather than the underlying model. Our methodology also facilitates the efficient comparison of different software and hardware stacks.

\end{abstract}

]

\printAffiliationsAndNotice{}
\section{Introduction}
\label{sec:introduction}

Large language models~\citep[LLMs;][]{devlin2018bert,brown2020language,rae2021scaling,lieberjurassic,black2022gpt,smith2022using,chowdhery2022palm,gpt4} have grown by almost four orders of magnitude in recent years, achieving state-of-the-art performance on traditional tasks like question answering and summarization~\cite{zellers2019hellaswag, hendrycks2020measuring}. 
LLMs display many new capabilities like reasoning about the physical world~\cite{bisk2020piqa}, solving grade-school math problems~\cite{cobbe2021training}, and generating code~\cite{chen2021evaluating}, to name a few.
To capitalize on these capabilities, several organizations offer access to LLMs through black-box text generation APIs~\cite{openai_api, ai21_api, cohere_api} and many companies are deploying LLM-powered products at scale like ChatGPT, Bing, \texttt{jasper.ai}, Github Copilot and OpenAI Playground~\cite{chatgpt, bing, scalegenerativeaiindex}.

When building models, both users and developers must balance the benefits of new capabilities against the costs of scale.
Recent efforts have begun to systematically evaluate and compare the downstream task accuracies of LLMs~\cite{brown2020language, rae2021scaling, srivastava2022beyond}, while others have examined the massive energy, financial, and computational costs of model training~\citep[][\S5.3]{cao2020towards, henderson2020towards, strubell2019energy, bender2021dangers, patterson2021carbon, bommasani2021opportunities}. However, few have considered the trade-offs of \textbf{inference efficiency vs. capability improvements}. This is important given that model inference costs might outweigh training costs for certain applications (e.g., ChatGPT).

Inference efficiency metrics are hard to estimate with black-box APIs. 
\textbf{Raw runtimes} of inference queries are not inherently \emph{comparable} across model providers since the API can include optimizations orthogonal to the model (e.g., caching,  customized hardware, etc.) and be susceptible to performance variance (e.g., in our experiments, we found that heavy load can worsen raw runtime by up to 2$\times$ for certain model providers).
This makes it hard to gauge the inference efficiency of models on a level playing field, which can be important for \emph{model creators and researchers} to understand the full \emph{long-term costs} of various training decisions (e.g., model architecture / size). Raw latency is still a good metric for \emph{end users} who are directly impacted by slow (or fast) predictions.
Another efficiency metric often used is the \textbf{model size}~\cite{wei2022emergent}, but this is hard to interpret and completely disregards practical deployment considerations (e.g., two models with the same size can have vastly different inference runtimes~\cite{fedus2021switch,jeon2018constructing,henderson2020towards, scao2022what}).

\begin{figure}
  \centering
  \includegraphics[width=\columnwidth]{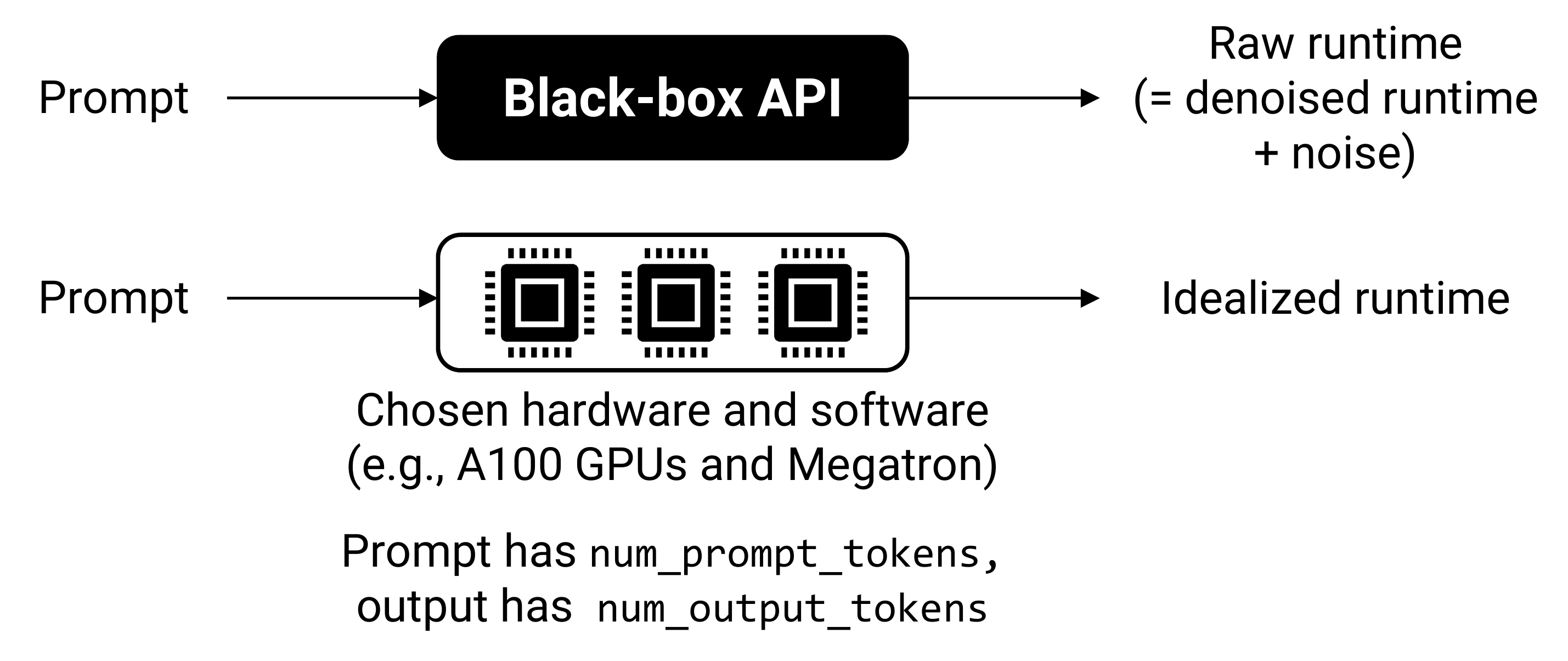}
  \caption{Comparison of raw runtime to the two runtime metrics proposed in this work for a given prompt size and number of output tokens.}
  \label{fig:metrics_comparison}
  \vspace{-0.1in}
\end{figure}

In this paper, we propose inference efficiency metrics that facilitate apples-to-apples comparisons across models.
The main metric we propose is the \textbf{idealized runtime}, which is the runtime of an inference query if run on \textbf{a specified software and hardware stack}.
The idealized runtime can be extended to calculate the idealized energy and dollar cost as well to take into account the number and type of accelerators used to serve the model. To measure the idealized runtime, we only require details on the model architecture used, even if the model parameters are not available.

The idealized runtime for a query can be estimated by passing the query through a standalone system instantiated with the chosen hardware and software; however, this is expensive for thousands of queries. Instead, we make the observation that runtime for autoregressive text generation using Transformer models is the sum of a linear function of the number of output tokens and a piecewise linear function of the number of prompt tokens; these functions are parameterized by $(m, s, h)$-specific parameters ($m$: model, $s$: software, $h$: hardware). This allows us to efficiently estimate the idealized runtime by fitting a linear regression model to the runtimes of a small set of ``configuration'' queries. This procedure also allows us to efficiently compare different software and hardware implementations for a given model: for example, we can quantify the speedup produced by FlashAttention~\cite{dao2022flashattention} on all inference queries in a benchmark, or determine if it is \emph{cheaper} to run a workload on older hardware (e.g., V100 GPUs).\footnote{While our method for efficient estimation is confined to Transformer models, we believe this is a reasonable compromise for now given that modern text generation APIs are powered almost exclusively by Transformer models. The same metrics can be measured for other model architectures, but na\"ive implementations will incur much higher profiling overheads.}

Using these metrics, we conduct a novel analysis of inference efficiency-capability tradeoffs for various Transformer models that are only available through black-box APIs (\S\ref{sec:tradeoffs}). We found that the idealized metrics can be a useful tool for model creators and researchers to understand the true inference costs that result from a particular training process and model architecture. For example, the vanilla \gptdavinci model is often on the Pareto frontier of the efficiency-capability trade-off landscape when using raw runtime as the efficiency metric on a set of 4 NLP scenarios covering sentiment analysis, question answering and classification. However, this efficiency appears to come from optimizations within the API rather than inherent efficiency in the model itself. When we compare models using idealized runtime, the set of models on the Pareto frontier is different, with \gptdavinci consistently not in it. 

\section{Transformer Models}
\label{sec:transformer_computation}

LLM APIs almost exclusively use Transformer models~\cite{vaswani2017attention}. In this section, we first provide important background on these models.

Transformer models consist of many Transformer layers, which themselves are composed of a self-attention layer and a two-layer FFN in traditional formulations. The input to a Transformer layer is a sequence of vector embeddings of \emph{tokens}. At a high level, the Transformer layer measures the importance of tokens on each other through the self-attention layer, and uses this cross-token importance to influence the model's output.
Unlike most models, Transformer models feature different compute patterns for training and inference (apart from the absence of a backward pass during inference); consequently, we describe these separately.

\begin{figure}
  \centering
  \includegraphics[width=\columnwidth]{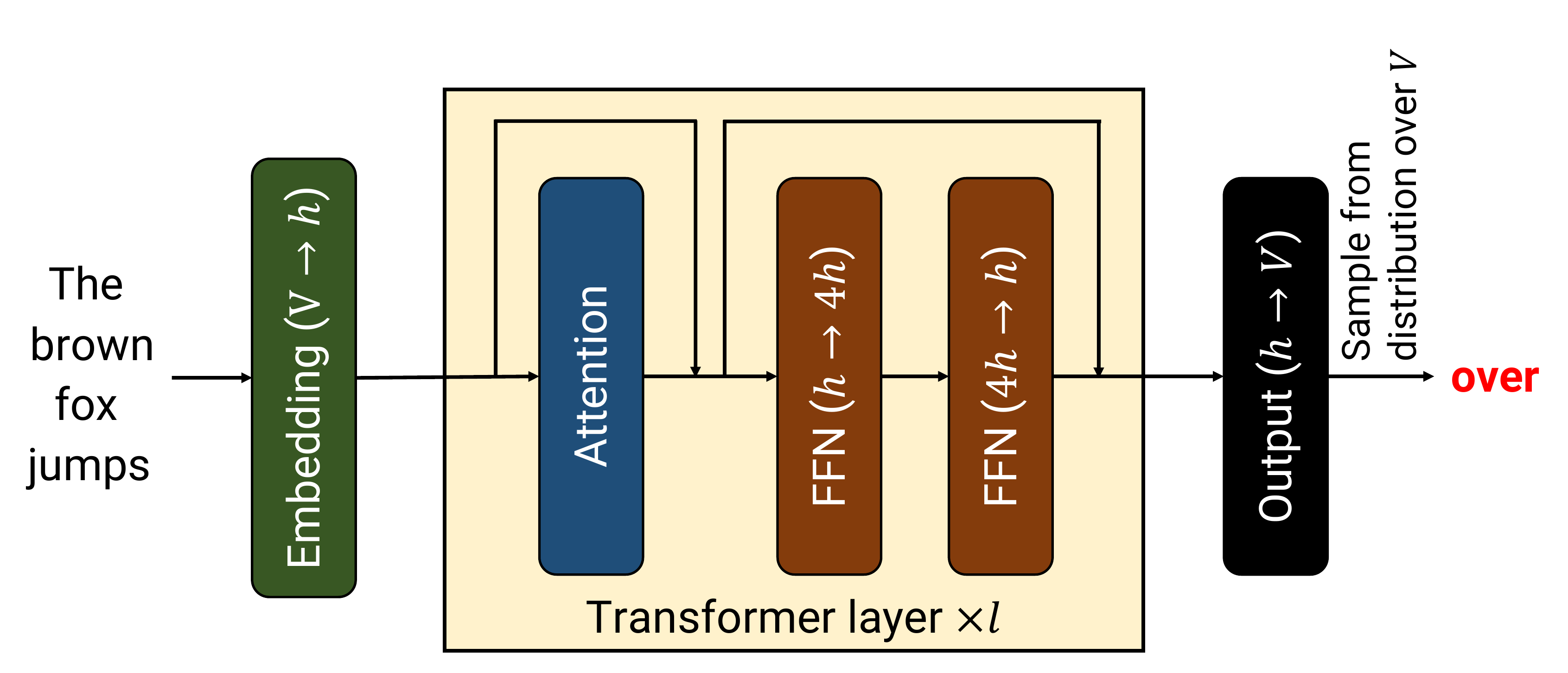}
  \caption{High-level schematic of a Transformer model with $l$ Transformer layers generating text at inference time given a prompt ``The brown fox jumps''.}
  \label{fig:transformer_schematic}
  \vspace{-0.1in}
\end{figure}

\subsection{Training} \label{sec:transformer_computation_training}

In this paper, we focus on language applications for Transformer models, where the input to the model is text. The input text is first preprocessed into a sequence of tokens (e.g., words) through a process called tokenization. Feature representations for each token (obtained by passing one-hot representations of the tokens through an embedding layer) are passed through multiple Transformer layers. Inputs to each Transformer layer are typically 3-dimensional tensors of shape $(b, s, h)$ where $b$ is the microbatch size (number of sequences), $s$ is the sequence length (number of tokens in each sequence), and $h$ is the hidden size (dimensionality of the model). For simplicity, we denote inputs as $X$. 

Transformer layers in language models use self-attention to allow tokens to ``interact'' with each other. Self-attention is composed of the following operations:
\begin{itemize}
    \item \textbf{Attention key ($K$), value ($V$), query ($Q$) transformations.} Given input $X$, we perform matrix multiplications $K = X \times W^K$, $V = X \times W^V$, and $Q = X \times W^Q$. $W^K$, $W^V$, and $W^Q$ are learned parameters.
    \item \textbf{Attention score computation.} Matrix multiplication $Q \times K^T$, followed by application of the softmax function to obtain score tensor $Z$. Each element $Z_{ij}$ is an importance score between query token $i$ and key token $j$. This is the primary mechanism that allows interaction across tokens in a sequence.
    \item \textbf{Attention over value computation.} Matrix multiplication of scores $Z$ by values $V$.
\end{itemize}

The subsequent two-layer feed forward network (FFN) consists of two linear layers (implemented as matrix multiplications). For most models, this involves multiplying the output of the self-attention layer by a matrix with dimension $h \times 4h$ and then multiplying the resulting output (after other operators like layer norm) by a matrix with dimension $4h \times h$. Figure~\ref{fig:transformer_schematic} shows how these operators are connected in a typical ``decoder-only'' Transformer model.

In aggregate, a forward pass through the Transformer layer of the model results in $24bsh^2\left(1 + \frac{s}{6h}\right)$ floating-point operations~\cite{narayanan2021efficient}, which scales linearly with the sequence length $s$ and quadratically with the hidden size $h$ if $s \ll 6h$, which is true for most LLMs. For a detailed explanation of this formula, see \S\ref{sec:transformer_operators_in_training} in the Appendix.

\subsection{Autoregressive Inference}
\label{sec:transformer_inference}

Auto-regressive language models like GPT-3~\cite{brown2020language} estimate the conditional probability $\Pr (x_i | x_{1:i-1})$ of a token $x_i$ given prefix tokens $x_1, x_2, \ldots, x_{i-1}$. During training, where we know all tokens in the training input a priori, the conditional probabilities $\Pr (x_1 | \emptyset), \Pr (x_2 | x_{1:1}), \Pr (x_3 | x_{1:2}), \ldots, \Pr (x_s | x_{1:s-1})$ can be estimated in parallel, and thus only a single forward pass needs to be executed in every iteration before the backward pass. However, at inference time, outputs of the model need to be fed back in as inputs to generate subsequent outputs. In particular, a token $x_i$ is sampled from the conditional probability distribution obtained by running a forward pass through the model. Different sampling approaches can be used to obtain the token $x_i$ from the conditional probability distribution $\Pr (x_i | x_{1:i-1})$; common approaches include greedy sampling, random sampling with temperature annealing, nucleus sampling, and beam search. The process then needs to be repeated for the next token $x_{i+1}$ and so on. Consequently, inference through an auto-regressive language model needs to perform \emph{multiple} forward passes. This entire procedure is very different to traditional inference for other models: for example, image classification for a ResNet-50 model involves just a single forward pass through the model.

Requests to language models are seeded with a prompt, which is a set of initial tokens $x_1, x_2, \ldots, x_p$ (we assume that the prompt has $p$ tokens). The conditional distribution $\Pr (x_{p+1} | x_{1:p})$ can then be computed through a forward pass. We call this the ``prompt encoding'' phase. Each subsequent generated token (sampled from $\Pr (x_{i+1} | x_{1:i})$ where $i > p$) needs its own forward pass through the model, which is the ``token-at-a-time generation'' phase.

\section{A Paremeterization of Autoregressive Inference Runtime}

We now derive a parameterized closed-form expression for the runtime of autoregressive inference given a prompt of size $p$ tokens and number of generated output tokens $o$.

\subsection{Closed-Form Expression for Runtime}

To generate $o$ tokens, $o-1$ additional forward passes are needed (the first token is generated during the prompt encoding phase).
The runtime of generating $o$ tokens given a prompt with $p$ tokens can be expressed as:
\begin{eqnarray}
& t(\text{prompt size } p, \text{number of output tokens } o) = \nonumber \\
& \textsf{\small prompt\_encoding\_time}(p) + \textsf{\small output\_generation\_time}(o). \nonumber
\end{eqnarray}

\subsubsection{Number of Floating-Point Operations}

To derive an expression for the end-to-end runtime of autoregressive inference, we first derive expressions for the number of floating-point operations required for each of the two steps, and then use these to derive expressions for runtime. We assume that the costs of projecting into vocabulary space in the output layer of the model and sampling the next token given the distribution $\Pr (x_{i+1} | x_{1:i})$ are cheap compared to the computation in the Transformer layers of this model, an observation made in previous work~\cite{narayanan2021efficient}.

\textbf{Prompt encoding.}
As outlined in \S\ref{sec:transformer_computation_training}, the total number of operations that need to be run in the prompt encoding phase for a single prompt of size $p$ is $24bph^2l\left(1 + \frac{p}{6h}\right)$, where $l$ is the number of Transformer layers in the model. $p \ll 6h$, so the number of compute operations needed to encode prompts simplifies to $24bph^2l$, or more simply $\theta_{pe} \cdot p$ (a linear function of $p$) for a given model with fixed $h$ and $l$.

\textbf{Output token generation.}
When using language models autoregressively to generate new text, the computations described in \S\ref{sec:transformer_computation_training} must be performed incrementally in the token generation phase. Concretely, the key, query, and value transformations need to be performed for just the new token, and self-attention scores need to be computed between the new token and all previous tokens.

We can compute the number of floating-point operations needed per Transformer layer to perform these computations. Let $i$ be the number of tokens generated so far (i.e., we are trying to generate the $(i+1)^\text{th}$ token, including the prompt). The total number of compute operations needed to generate the $(i+1)^\text{th}$ token is $24bh^2l + 4bihl = 24bh^2l\left(1 + \frac{i}{6h}\right)$ (see \S\ref{sec:transformer_operators_in_inference} in the Appendix for details).
If $i \ll 6h$, which is largely true in practice (e.g., for \gptdavinci, the maximum context length is 2048 and $h = 12288$), the floating-point operations to generate a new token is roughly independent of the number of tokens generated so far (we denote this by $\theta_{og}$).

\subsubsection{End-to-End Runtime}
Runtimes for each of these stages can be expressed as the ratio of the number of floating-point operations and the corresponding throughputs:
\begin{eqnarray}
\textsf{\small prompt\_encoding\_time}(p) = \dfrac{\theta_{pe} \cdot p}{\text{throughput}_{pe}(p)} \label{eqn:prompt_encoding_time}
\end{eqnarray}
\begin{eqnarray}
\textsf{\small output\_generation\_time}(o) = \sum_{\text{token } i} \dfrac{\theta_{og}}{\text{throughput}_{og}(i)} \label{eqn:output_generation_time}
\end{eqnarray}
Usually, $\text{throughput}_{og}$ is a constant independent of the token being generated meaning \textsf{\small output\_generation\_time} is a linear function of $o$; we will show this empirically next.

\begin{figure}[t!]
    \centering
    \begin{subfigure}[c]{\columnwidth}
        \centering
        \includegraphics[keepaspectratio=1.0,width=0.75\columnwidth]{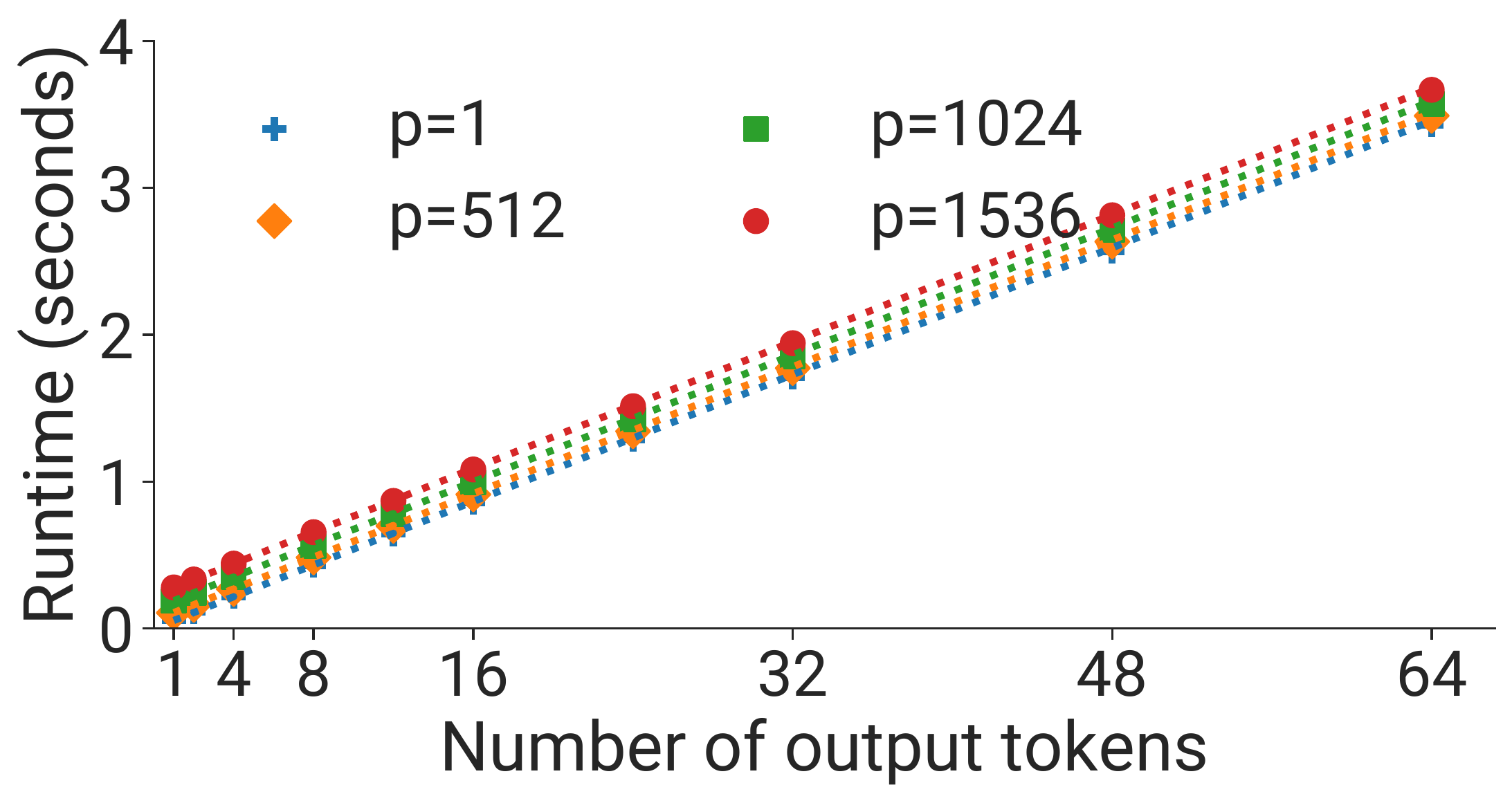}
        \caption{\anthropic.}
    \end{subfigure}
    \centering
    \begin{subfigure}[c]{\columnwidth}
        \centering
        \includegraphics[keepaspectratio=1.0,width=0.75\columnwidth]{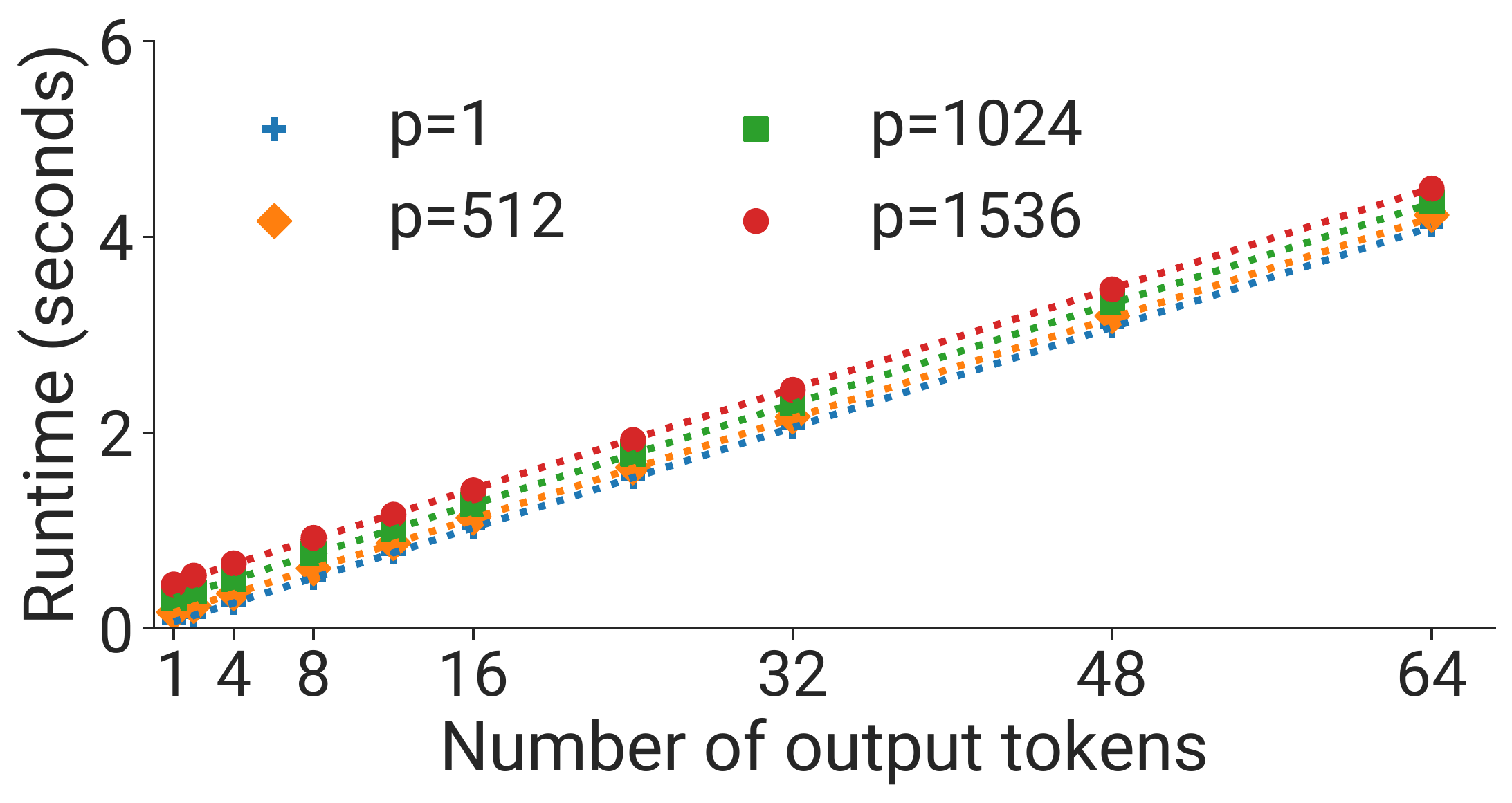}
        \caption{\jurassicjumbo.}
    \end{subfigure}
    \centering
    \begin{subfigure}[c]{\columnwidth}
        \centering
        \includegraphics[keepaspectratio=1.0,width=0.75\columnwidth]{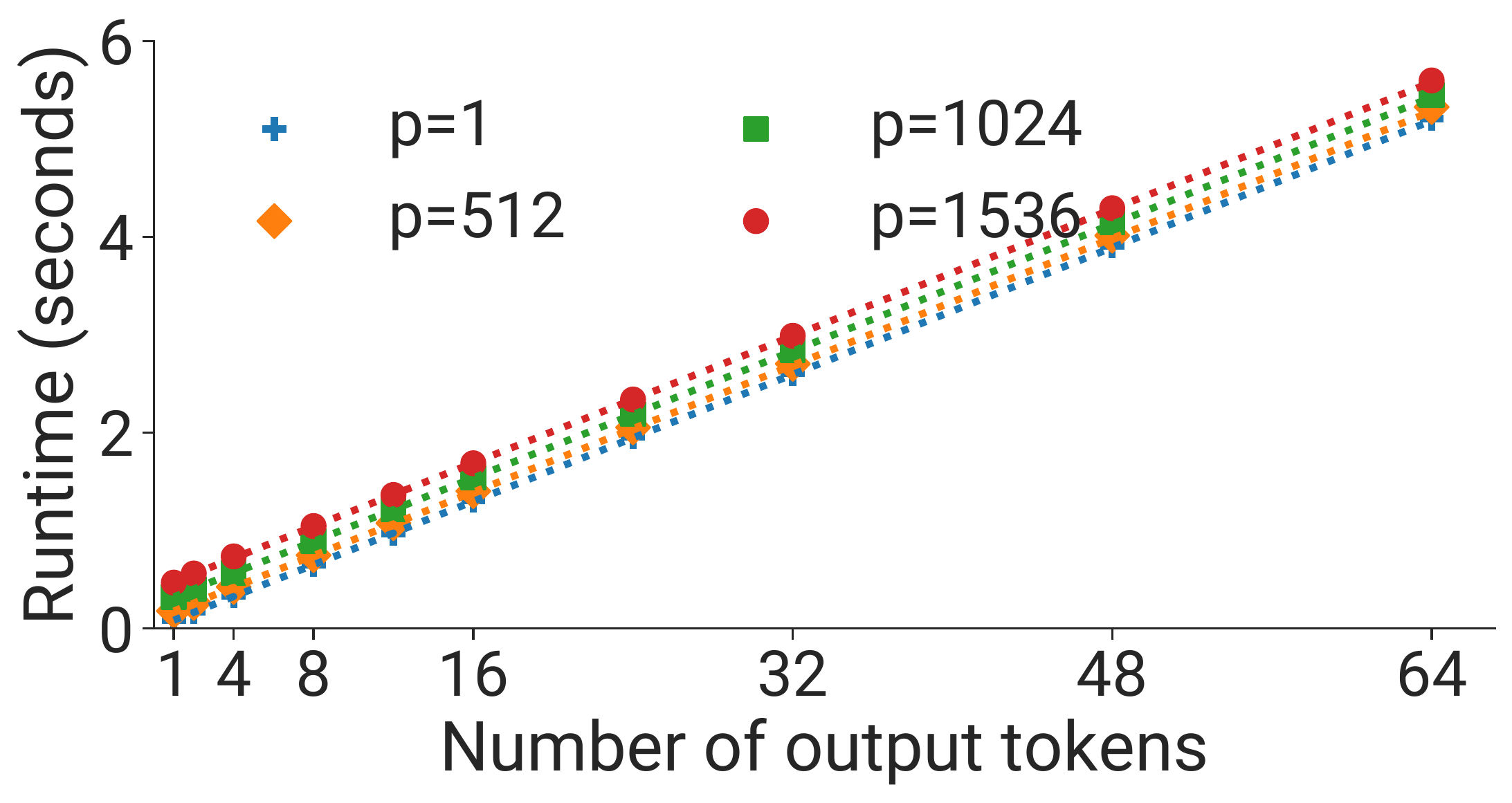}
        \caption{\gptdavinci.}
    \end{subfigure}
    \centering
    \begin{subfigure}[c]{\columnwidth}
        \centering
        \includegraphics[keepaspectratio=1.0,width=0.75\columnwidth]{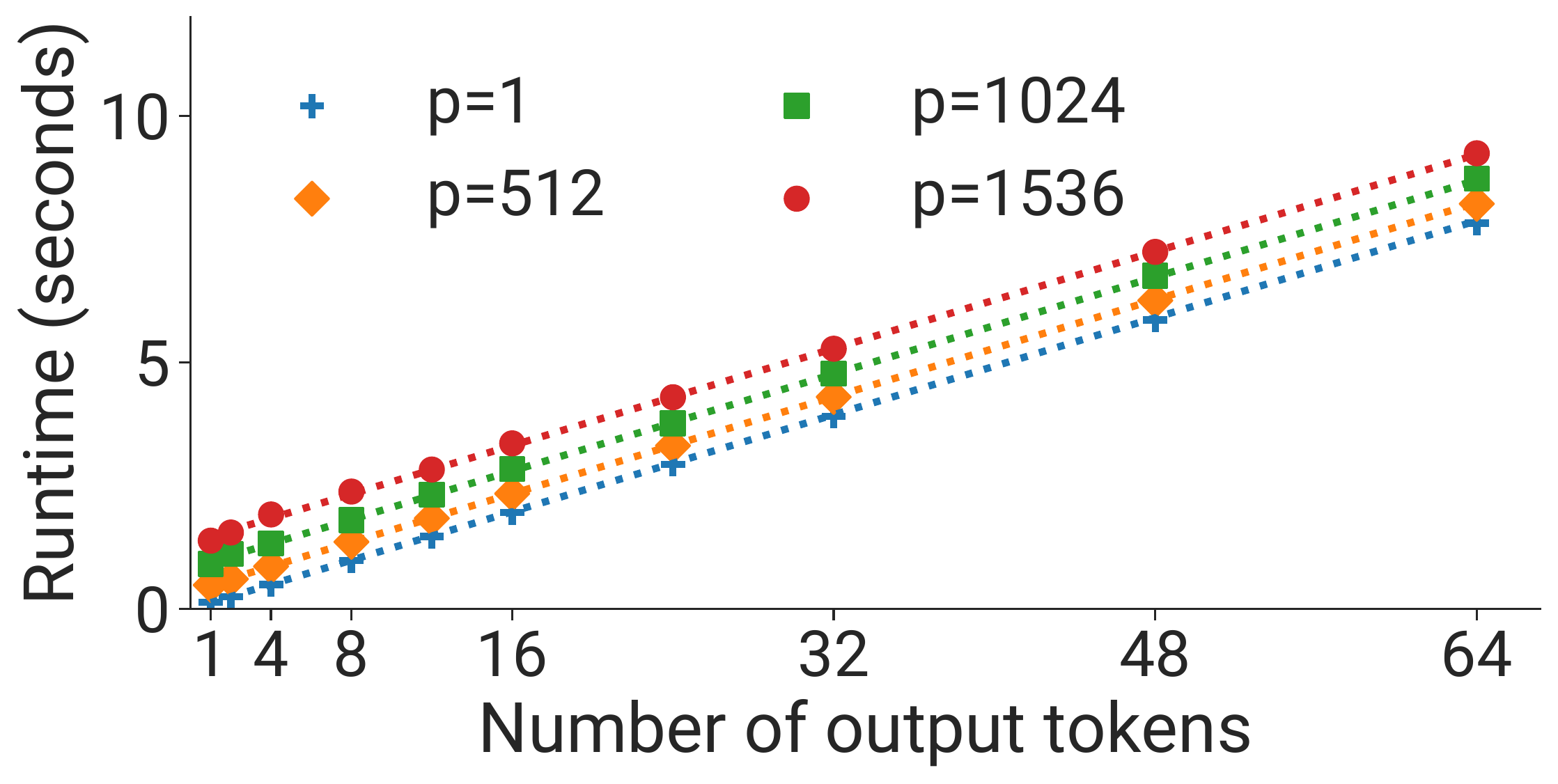}
        \caption{\mtnlg.}
    \end{subfigure}
    \caption{
        End-to-end runtimes for different prompt sizes (shown in legend in terms of number of tokens) as the number of generated output tokens is varied using Megatron.
    }
    \label{fig:runtime_vs_num_output_tokens_megatron}
    \vspace{-0.1in}
\end{figure}

\begin{figure}[t!]
    \centering
    \begin{subfigure}[c]{0.48\columnwidth}
        \centering
        \includegraphics[keepaspectratio=1.0,width=0.83\columnwidth]{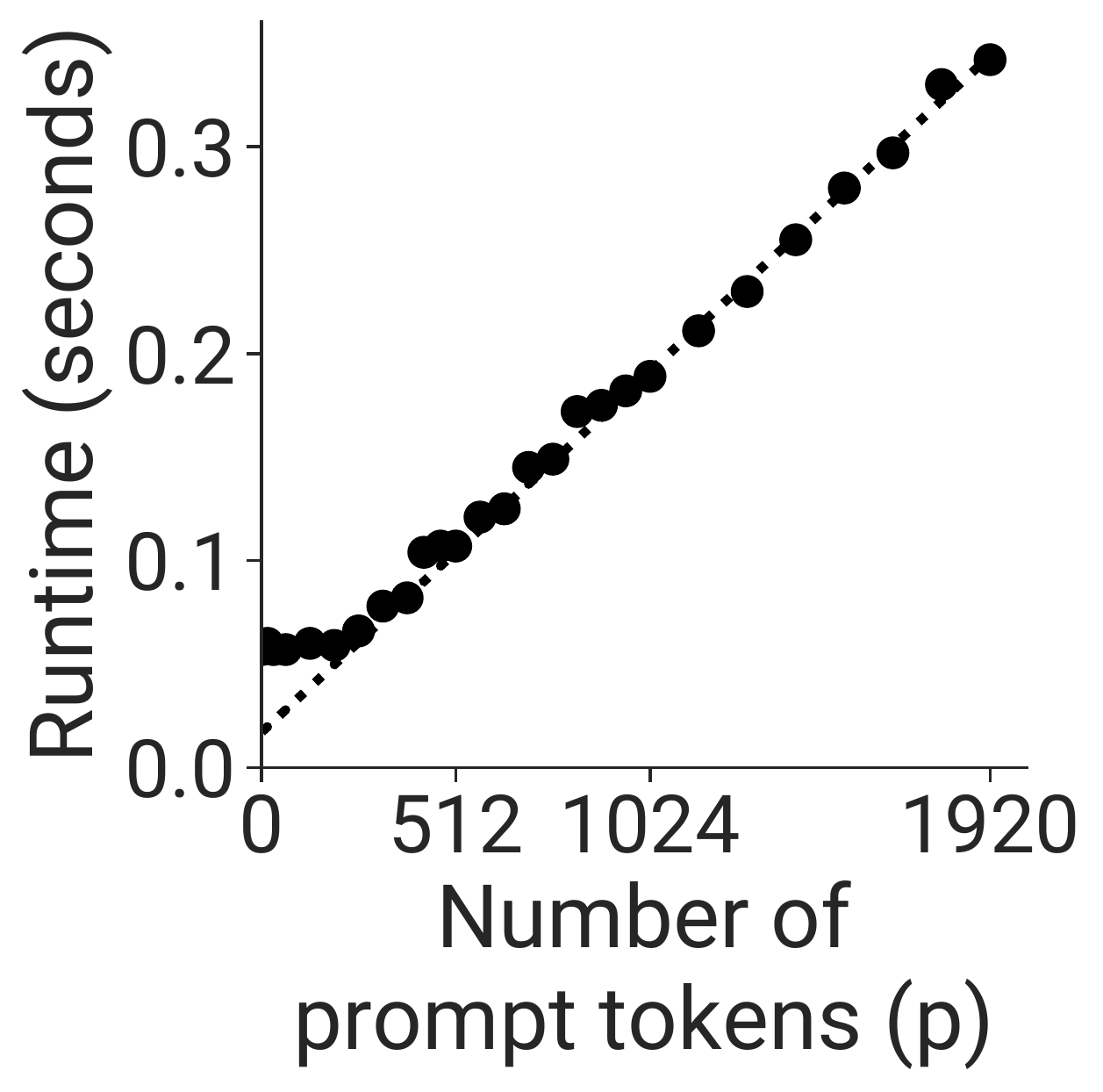}
        \caption{\anthropic.}
    \end{subfigure}
    \centering
    \begin{subfigure}[c]{0.48\columnwidth}
        \centering
        \includegraphics[keepaspectratio=1.0,width=0.83\columnwidth]{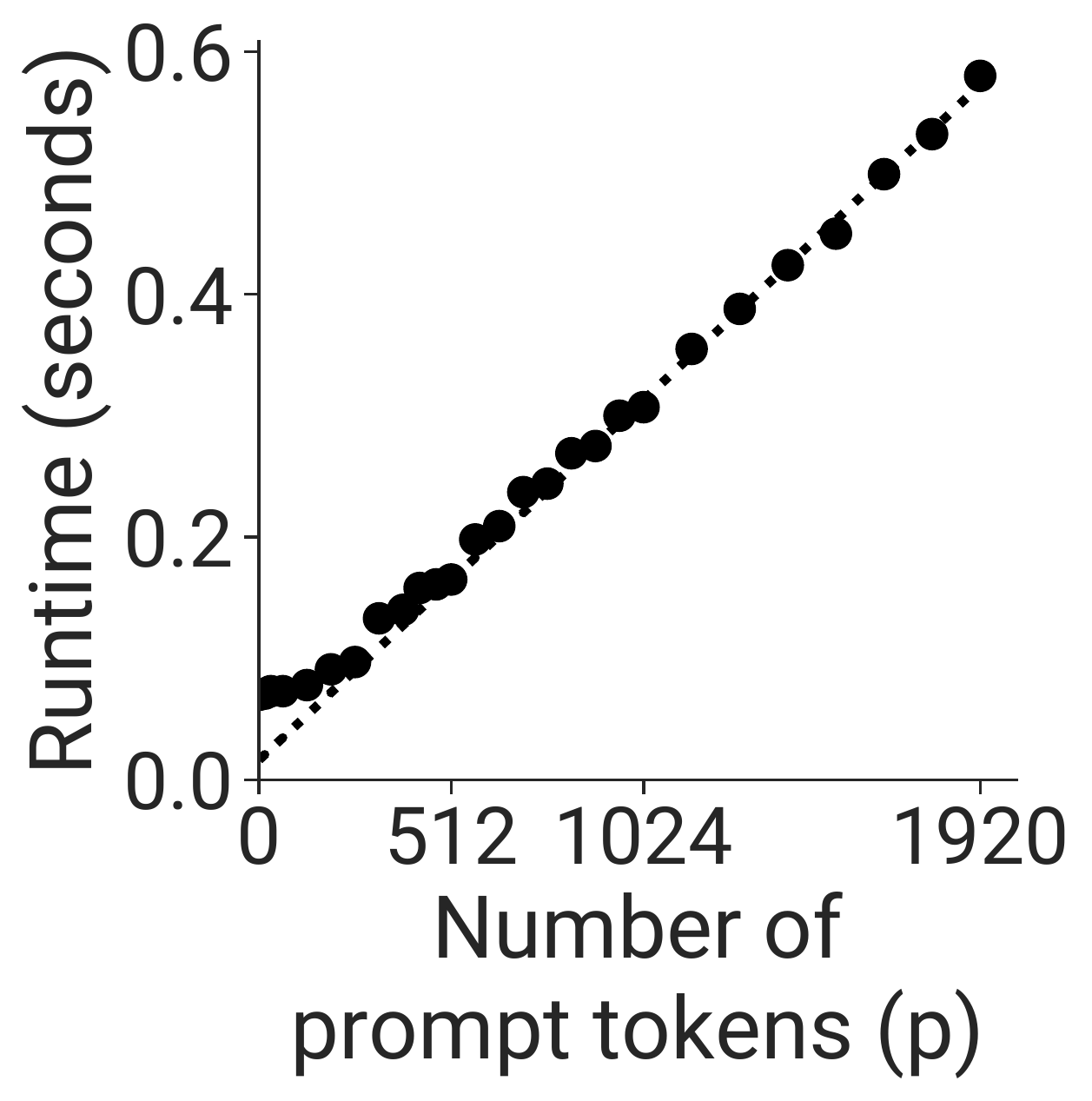}
        \caption{\jurassicjumbo.}
    \end{subfigure}
    \centering
    \begin{subfigure}[c]{0.48\columnwidth}
        \centering
        \includegraphics[keepaspectratio=1.0,width=0.83\columnwidth]{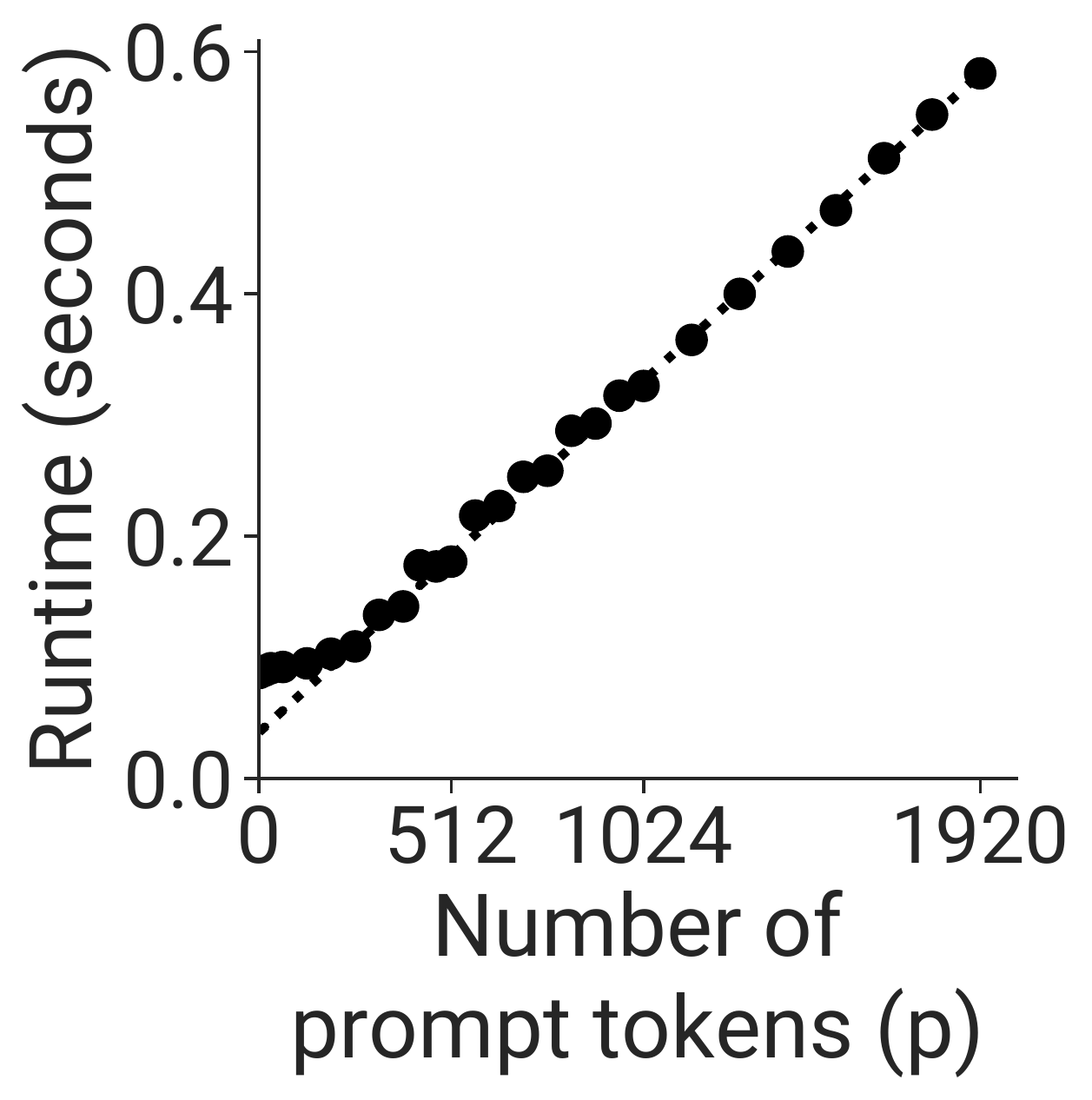}
        \caption{\gptdavinci.}
    \end{subfigure}
    \centering
    \begin{subfigure}[c]{0.48\columnwidth}
        \centering
        \includegraphics[keepaspectratio=1.0,width=0.83\columnwidth]{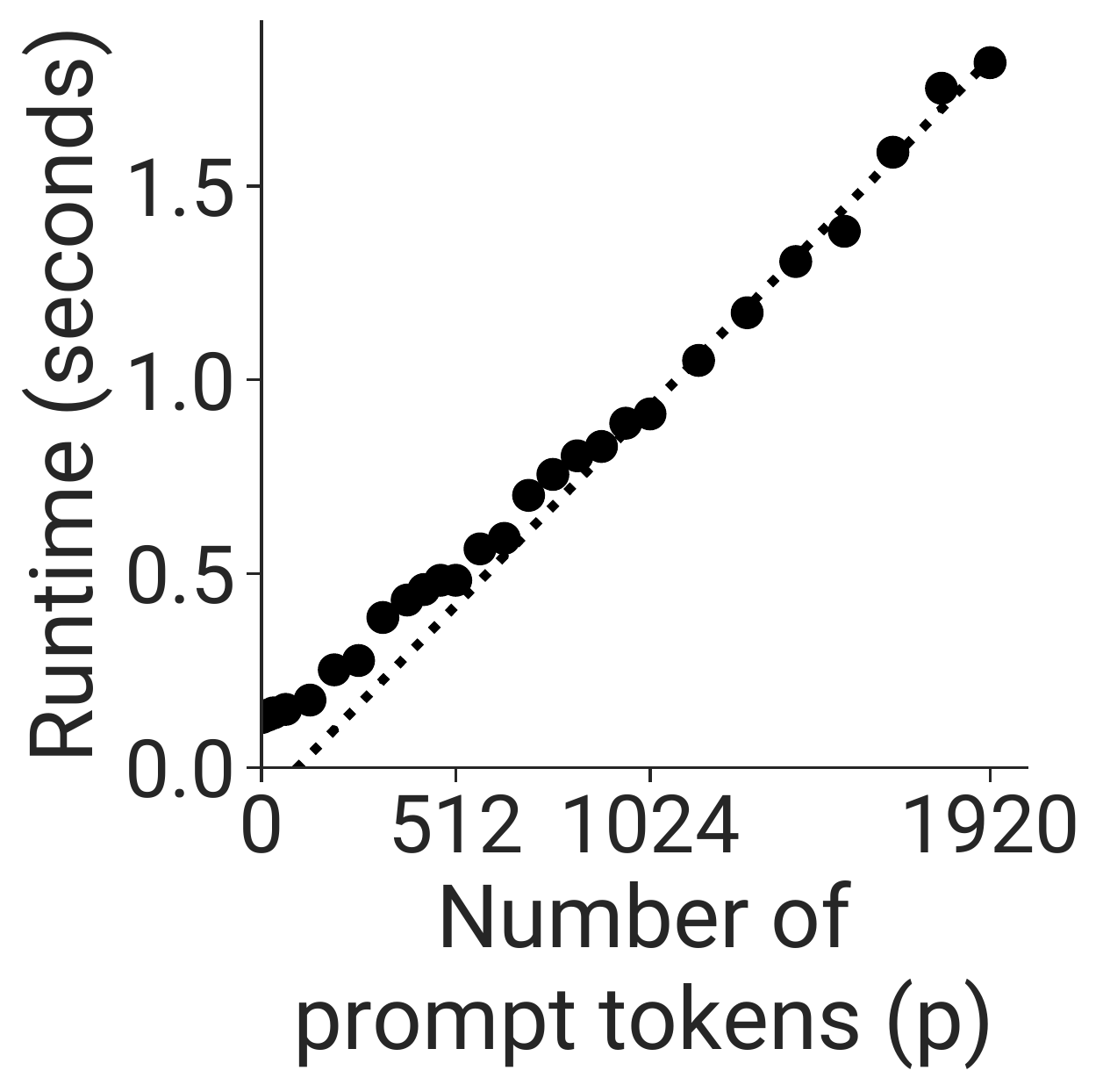}
        \caption{\mtnlg.}
    \end{subfigure}
    \caption{
        End-to-end runtimes versus prompt sizes for various models. We also show a dotted best-fit line.
    }
    \label{fig:runtime_vs_num_prompt_tokens_megatron}
    \vspace{-0.1in}
\end{figure}

\begin{table*}[t!]
\centering
\begin{tabular}{lccccccc}
\toprule
Model (owner/name) & Provider & $h$ & $l$ & $n$ & \# Parameters (billion) & \# GPUs$\times$GPU type \\
\toprule
\gptdavinci & OpenAI & 12288 & 96 & 96 & 175 & 8$\times$80GB-A100 \\ 
\jurassiclarge  & AI21 Labs & 4096 & 32 & 32 & 6.7 & 1$\times$80GB-A100 \\
\jurassicgrande  & AI21 Labs & 5120 & 50 & 40 & 17 & 1$\times$80GB-A100 \\
\jurassicjumbo  & AI21 Labs & 13824 & 76 & 96 & 178 & 8$\times$80GB-A100 \\
\coherexl & Cohere & 8192 & 64 & 64 & 52 & 4$\times$80GB-A100 \\
\midrule
\anthropic & Anthropic & 8192 & 64 & 64 & 52 & 4$\times$80GB-A100 \\
\mtnlg & Microsoft & 20480 & 105 & 128 & 530 & 24$\times$80GB-A100 \\
\midrule
\gptj & Together & 4096 & 28 & 16 & 6 & 1$\times$80GB-A100 \\
\yalm & Together & 10240 & 80 & 128 & 100 & 4$\times$80GB-A100 \\
\bloom & Together & 14336 & 70 & 112 & 176 & 8$\times$80GB-A100 \\
\bottomrule
\end{tabular}
\vspace{-0.05in}
\caption{Models studied in this paper. We also specify the number of GPUs / GPU type used to estimate the default idealized runtimes (different configurations are used with 32GB-V100 GPUs).}
\vspace{-0.1in}
\label{table:models}
\end{table*}

\subsection{Empirical Results}
\label{sec:empirical_runtimes}

We can validate the above equations empirically.

\textbf{Models.}
In this paper, we study 10 state-of-the-art LLMs. Each of these is a Transformer model, but with different hyperparameters that control the size of the model.

\textbf{Setup.} We use Megatron's (a high-performance GPU implementation) Transformer and autoregressive inference functionality. We also use the minimum number of GPUs necessary.
For example, \gptdavinci ~cannot fit on a single 80-GB A100 GPU; we use tensor model parallelism~\cite{shoeybi2019megatron} to ensure that the model parameters fit in GPU memory in such cases.
Tensor model parallelism is optimal within a multi-GPU server~\cite{narayanan2021efficient} since expensive all-to-all communication is limited to fast high-bandwidth NVLink. For even larger models like \mtnlg, we need other forms of parallelism like pipeline model parallelism in order to fit the model in GPU memory without poor scaling. We use A100 GPUs because they are the fastest widely available GPU right now. Other accelerators like the TPU~\cite{jouppi2017datacenter} or the NVIDIA H100 GPU could also be used. Table~\ref{table:models} shows the exact hardware configurations used.

\textbf{Results.}
Figure~\ref{fig:runtime_vs_num_output_tokens_megatron} shows the end-to-end runtime versus number of generated output tokens for different prompt sizes and models.
We instantiate models based on reported architectures, but without trained parameters, as we only care about estimating runtime on the dedicated hardware, and runtime is independent of the model's parameters given a prompt size and number of output tokens. 

For each prompt size $p$, we can compute a best-fit line using linear regression.
We observe that the coefficients of determination ($R^2$) for the resulting time estimates are very close to 1.0 ($>0.999$) for all models. Consequently, we see empirically that runtime shows a linear relationship with the number of output tokens for each prompt size, indicating that $\textsf{\small output\_generation\_time}(o)$ is a linear function of $o$ (i.e., throughput$_{og}$ is independent of the token being generated).

Runtime also increases empirically with prompt size. Figure~\ref{fig:runtime_vs_num_prompt_tokens_megatron} shows the prompt encoding time versus prompt size ($p$) for the same set of 4 models. We see that runtime and the prompt size have a roughly linear relationship, especially at large prompt sizes. However, this linear relationship breaks down at smaller prompt sizes. We can see why this is the case when looking at Equation~\ref{eqn:prompt_encoding_time}; the prompt-encoding throughput (throughput$_{pe}$) initially increases as $p$ increases (arithmetic intensity~\cite{williams2009roofline} of the computation increases with $p$) but eventually plateaus. Consequently, we observe that $\textsf{\small prompt\_encoding\_time}$ is piecewise linear.

\subsection{Final Parameteric Form}

We conclude that the end-to-end runtime of autoregressive inference with a Transformer model is the sum of a piecewise linear function of $p$ and a linear function of $o$ (for simplicity, we will continue to denote the function for the runtime of prompt encoding as \textsf{\small prompt\_encoding\_time} since piecewise linear functions are clunky to write out fully):
\begin{eqnarray}
& t(\text{prompt size } p, \text{number of output tokens } o) = \nonumber \\
& \textsf{\small prompt\_encoding\_time}(p) + (o - 1) \cdot g. \label{eqn:runtime_as_linear_function}
\end{eqnarray}

\subsection{Estimation Procedure}
\label{sec:estimation}

Equation~\ref{eqn:runtime_as_linear_function} provides a parameterization of the end-to-end runtime for autoregressive Transformer LLMs for arbitrary prompt size $p$ and number of generated tokens $o$, and suggests an efficient way of estimating the runtime of a query with given prompt size and number of output tokens on a \emph{target system} instead of running each query multiple times.

For each model and target system, we follow a two-step process. First, for a given prompt size $p$, we profile the autoregressive Transformer LLM with different numbers of output tokens, and then fit a linear regression model to the end-to-end runtimes. The resulting $y$-intercept gives us the prompt encoding time for that $p$. We repeat this procedure for all prompt sizes that we want to explore.
For example, if $\text{max\_context\_length} =$ 2048, then one possible range of prompt sizes to explore is $P = \{$1, 256, 512, 1024, 1536$\}$. In practice, the number of tokens $p$ in a prompt of a query might not be in the set of prompt sizes explored, in which case we can interpolate between known data points, since $\textsf{\small prompt\_encoding\_time}$ is piecewise linear.

Equipped with these prompt encoding runtimes, we can leverage the fact that $\textsf{\small total\_runtime}(p, o) - \textsf{\small prompt\_encoding\_time}(p)$ is a linear function in $o$: we can fit a single linear regression model with $y = \text{runtime difference}$ and $x = o$ to obtain an estimate for the slope $g$, the runtime cost of generating the next output token for this model and target system.

\subsection{Empirical Results with Black-Box APIs}
\label{sec:empirical_runtimes_blackbox_apis}

We can run a similar experiment using black-box APIs. The runtime for text generation using a black-box API can be expressed by Equation~\ref{eqn:runtime_as_linear_function} with a small modification:
\begin{eqnarray}
& t(\text{prompt size } p, \text{number of output tokens } o) = \nonumber \\
& \textsf{\small prompt\_encoding\_time}(p) + (o-1) \cdot g + \textsf{\small overhead} \nonumber.
\end{eqnarray}
In the above equation, $\textsf{\small overhead}$ captures the fixed costs of using an API to serve model predictions instead of using accelerators locally (e.g., round-trip latency of communicating with a remote API server) and performance variability (e.g., queuing delay or performance interference across requests).

\begin{figure*}[t!]
    \centering
    \begin{subfigure}[c]{0.65\columnwidth}
        \centering
        \includegraphics[keepaspectratio=1.0,width=0.95\columnwidth]{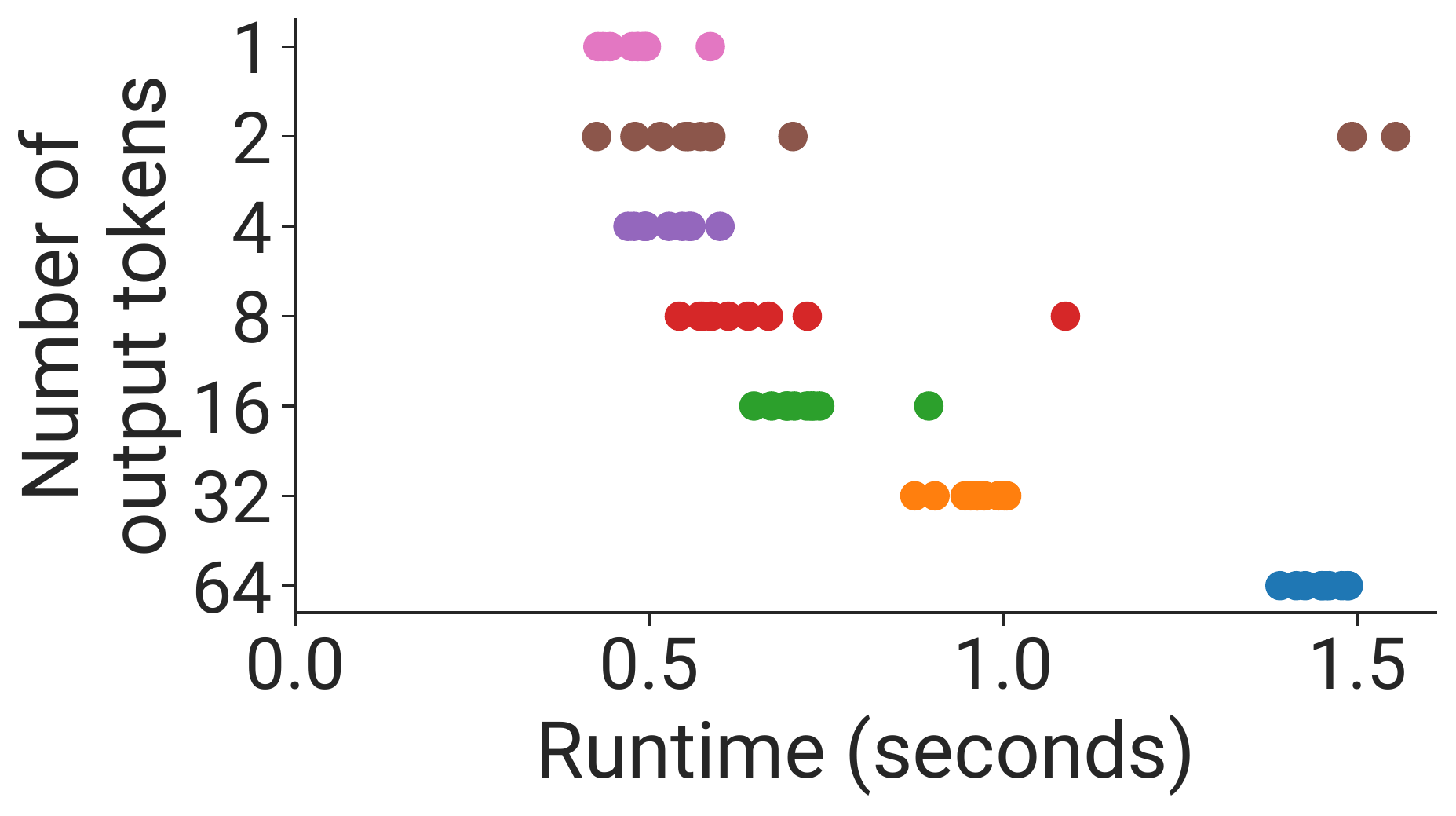}
        \caption{\jurassiclarge.}
    \end{subfigure}
    \centering
    \begin{subfigure}[c]{0.65\columnwidth}
        \centering
        \includegraphics[keepaspectratio=1.0,width=0.95\columnwidth]{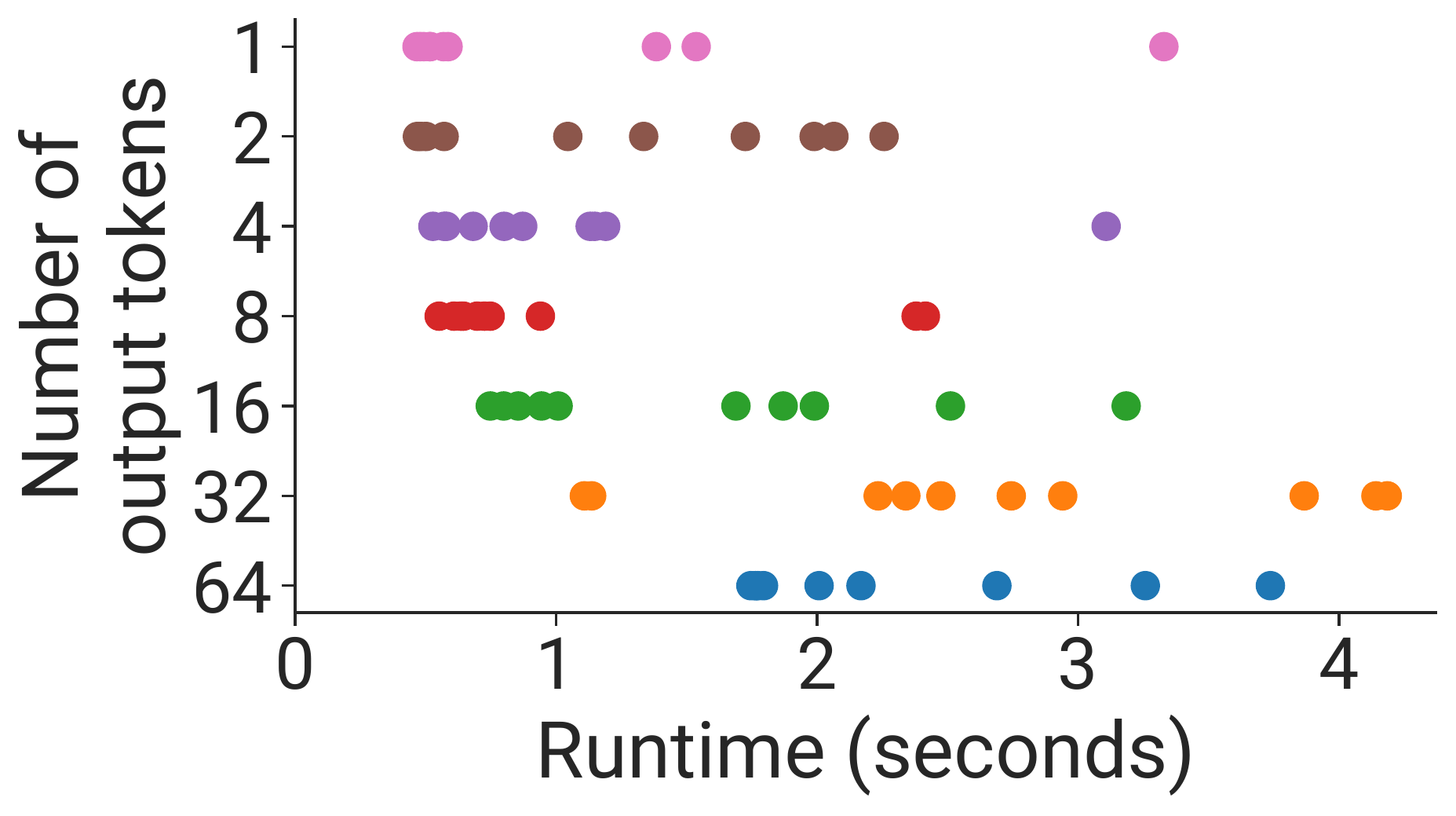}
        \caption{\jurassicgrande.}
    \end{subfigure}
    \centering
    \begin{subfigure}[c]{0.65\columnwidth}
        \centering
        \includegraphics[keepaspectratio=1.0,width=0.95\columnwidth]{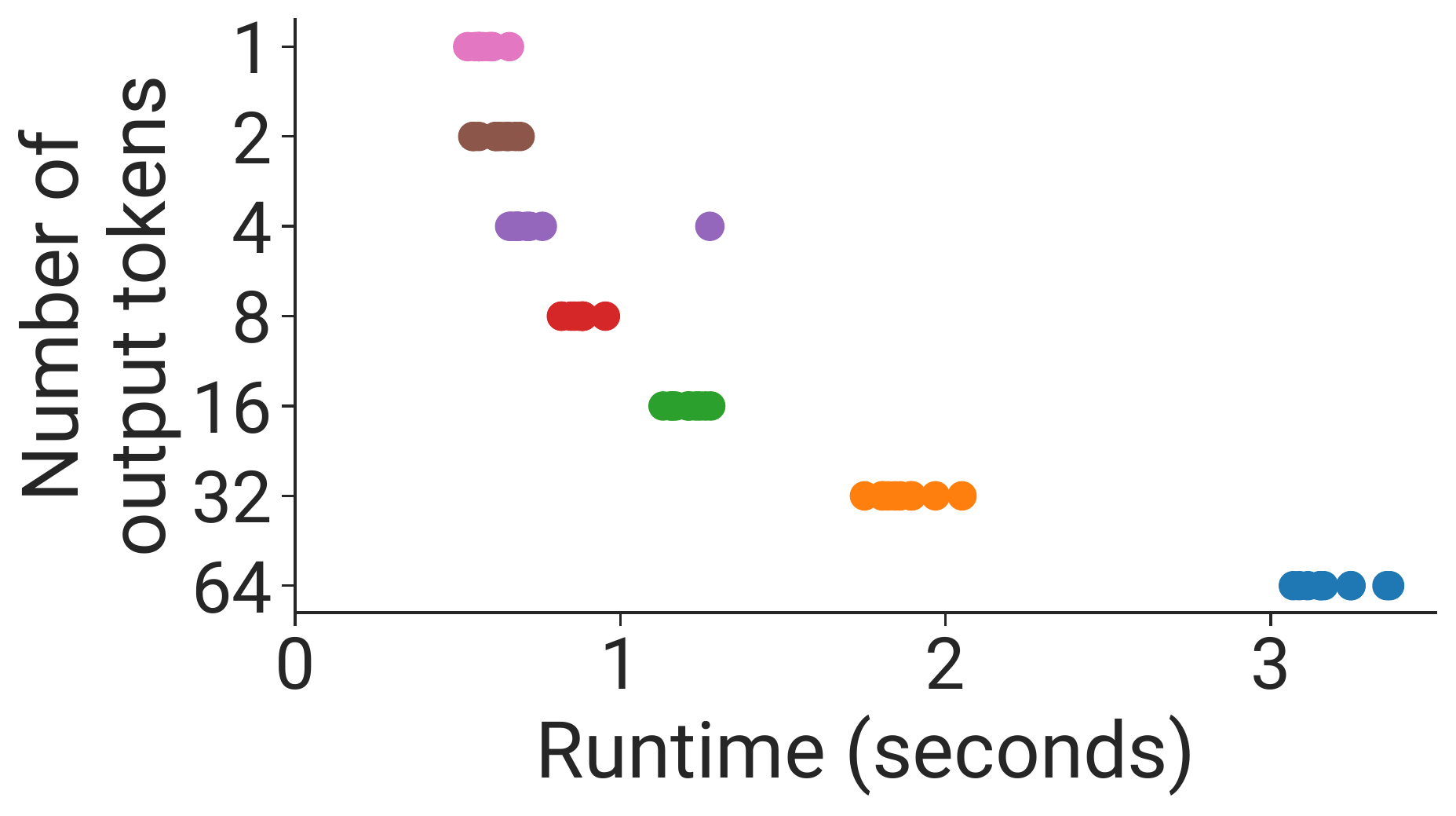}
        \caption{\jurassicjumbo.}
    \end{subfigure}
    \caption{
        Per-instance runtimes using black-box APIs to access LLMs for multiple instances (prompt size, $p = 512$).
    }
    \label{fig:per_instance_runtime_variance}
\end{figure*}

\textbf{Variation of runtimes across trials.}
To better quantify performance variability when using black-box APIs, we run multiple trials of synthetic queries where we control the size of the prompt and the number of generated output tokens. Figure~\ref{fig:per_instance_runtime_variance} shows per-trial runtimes for different model offerings from the same model provider (AI21). Unless otherwise noted, all experiments in this paper were run in September or October 2022 with the latest API versions available at the time.

We see discernible performance variance across multiple trials for different models, across prompt sizes and number of generated output tokens. 
Certain models experience higher performance variability: Figure~\ref{fig:per_instance_runtime_variance} shows \jurassicgrande~has much higher performance variance than \jurassiclarge~or \jurassicjumbo (larger spread among points for a query of given size). \jurassicgrande~has an average coefficient of variation of about 0.55 compared to much smaller coefficients of variation ($\sim$0.2) for the other AI21 models. Even for models with lower spreads (e.g., \jurassiclarge), we see that outlier points can have as much as 3$\times$ higher latency.

\begin{figure}[t!]
    \centering
    \includegraphics[keepaspectratio=1.0,width=0.75\columnwidth]{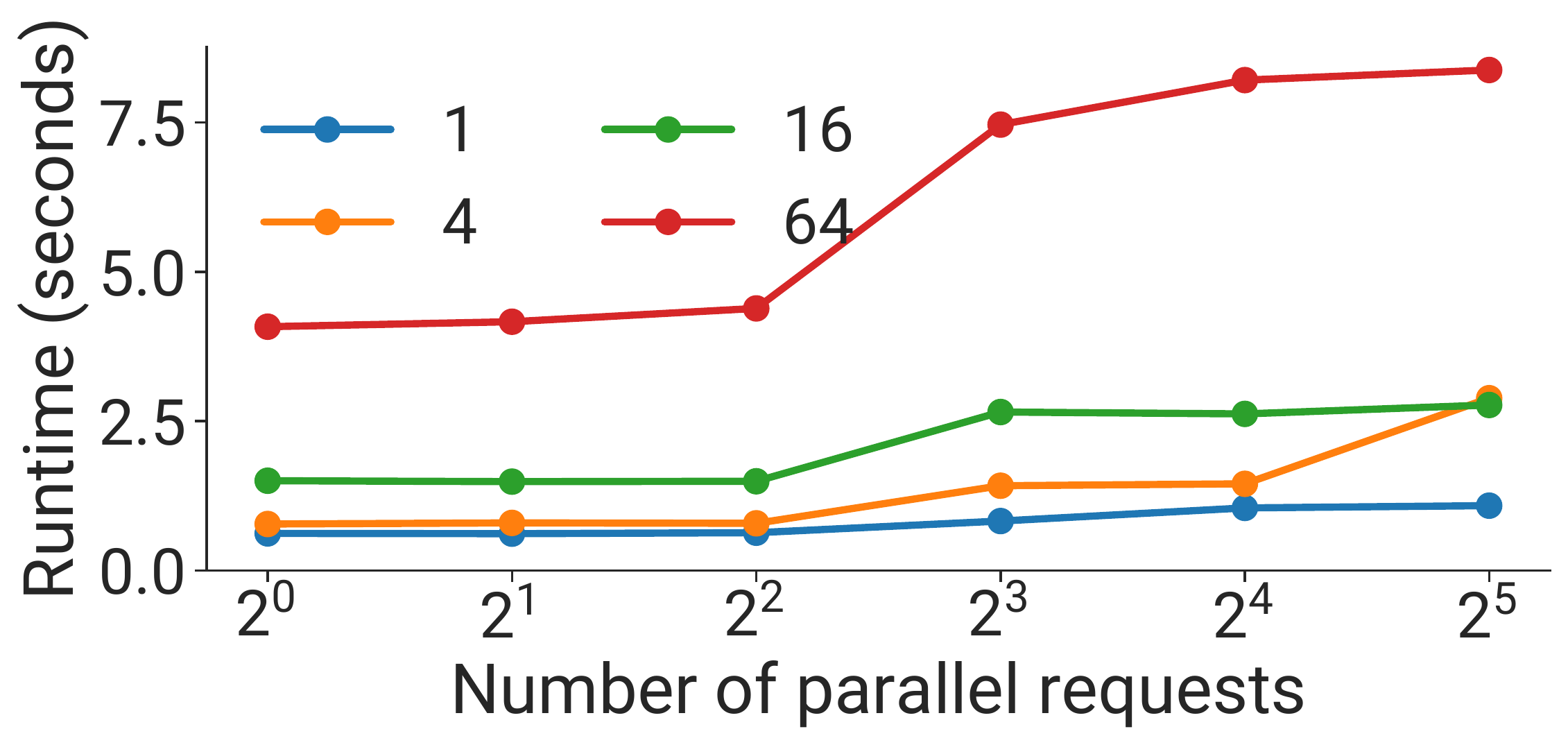}
    \caption{
        Minimum runtime across 10 trials as number of parallel queries increases for the \anthropic model. Prompt size is 512 tokens and the number of output tokens is varied (shown in legend). Experiment was run in 10/2022.
    }
    \label{fig:runtime_vs_load}
\end{figure}

\textbf{Variation of runtimes with load.}
To understand the impact of load on performance contention and end-to-end runtime, we measured query runtime as we increase the number of queries sent in parallel to the various black-box APIs. Figure~\ref{fig:runtime_vs_load} shows runtime versus number of parallel queries for different numbers of output tokens and a fixed prompt size of 512 tokens for the \anthropic model. We observe as much as a 2$\times$ increase in runtime, indicating that load can lead to increased contention on API servers and consequently increased observed runtime.

\section{Idealized and Denoised Metrics}
\label{sec:metrics}

In this section, we propose two concrete parameterizations of Equation~\ref{eqn:runtime_as_linear_function} that result in two runtime metrics that can be used for different types of downstream analyses. These metrics can also be used to derive other metrics in terms of dollar cost or consumed energy.

\subsection{Runtime Metrics}

We can find the underlying performance parameters in Equation~\ref{eqn:runtime_as_linear_function} in a couple of different ways, yielding different runtime metrics. 

\textbf{Idealized runtime.} The runtime using a uniform hardware and software implementation (e.g., NVIDIA A100 GPUs and Megatron respectively), allowing for the inference efficiency of models to be directly compared with each other.
\begin{eqnarray}
& t^\text{idealized}_{(m, s, h)}(\text{prompt size } p, \text{number of output tokens } o) = \nonumber \\
& \textsf{\small prompt\_encoding\_time}^\text{idealized}_{(m, s, h)}(p) + (o - 1) \cdot g^\text{idealized}_{(m, s, h)}. \nonumber
\end{eqnarray}

\textbf{Denoised runtime.} In an attempt to test whether our idealized runtime metric is accurate, we also propose a runtime metric that factors out the noise from performance variation. We call this the denoised runtime; we assume use of the same hardware and software used by the API provider.
\begin{eqnarray}
& t^\text{denoised}_{m \text{ on API } a}(\text{prompt size } p, \text{number of output tokens } o) = \nonumber \\
& \textsf{\small prompt\_encoding\_time}^\text{denoised}_{m \text{ on API } a}(p) + \nonumber (o - 1) \cdot g^\text{denoised}_{m \text{ on API } a}. \nonumber
\end{eqnarray}

To estimate denoised runtime, we profile the models through the provided black-box APIs directly using synthetic prompts with pre-configured sizes, as outlined in \S\ref{sec:estimation}. We see higher variance in runtimes when using black-box APIs relative to dedicated hardware. Since the performance noise is a random variable $\eta \geq 0$, we can run multiple trials in the profiling step and perform the linear regression using the \emph{minimum} obtained runtime (i.e., the runtime with minimum \emph{variable} overhead) across trials for each prompt size and number of generated tokens.

We observe that the following inequality should hold for any model $m$ on API $a$ for a prompt of size $p$ and number of output tokens $o$, as long as the idealized runtime is computed for software $s^*$ and hardware $h^*$ that are \emph{at least as fast} than those used to back the original API $a$:
\begin{eqnarray}
   t^\text{raw}_{m \text{ on API } a}(p, o) &\geq& t^\text{denoised}_{m \text{ on API } a}(p, o) \nonumber \\
                                            &\geq& t^\text{idealized}_{(m, s^*, h^*)}(p, o). \nonumber
\end{eqnarray}
This is by construction (software $s^*$ and hardware $h^*$ are assumed to be at least as fast as that used by the API provider) and since the denoised runtime is the raw runtime with performance variation factored out.

\subsection{Incorporating Scale}

Larger models often require more accelerators just to fit the model in accelerator memory. As a result, just comparing runtimes between two models does not accurately capture the cost of running inference for the model. We propose two metrics that explicitly take into account scale. Both metrics are derived from the idealized runtime by multiplying with the number of accelerators used and a metric-specific scaling factor (e.g., cost per hour or power draw of an A100 GPU). Unfortunately, we cannot similarly modify the denoised runtime since we do not know the tyoe of hardware and the number of chips used by the model provider.

\textbf{Idealized dollar cost.}
We can compute the idealized dollar cost as follows:
\begin{equation}
    t^\text{idealized}_{(m, s, h)} \text{ (secs)} \times n_\text{accelerator $h$} \times c_\text{accelerator $h$} \text{ (\$/sec)}. \nonumber
\end{equation}
$n_\text{accelerator $h$}$ is the number of accelerators used at a time to serve a single request ($1$ if not using model parallelism, $> 1$ otherwise), and $c_\text{accelerator $h$}$ is the per-unit-time cost of the hardware $h$ (e.g., if $h$ is NVIDIA A100 GPUs, then $c_\text{accelerator}$ could then be the per-hour cost of renting an NVIDIA A100 GPU in the cloud like on AWS). The idealized dollar cost is then the cost of serving the model on A100 GPUs on AWS.

\textbf{Idealized energy cost.}
Similar to work that has examined the energy cost of training~\cite{cao2020towards, henderson2020towards, strubell2019energy, patterson2021carbon}, we can estimate the idealized energy cost as follows:
\begin{equation}
    t^\text{idealized}_{(m, s, h)} \text{ (secs)} \times n_\text{accelerator $h$} \times p_\text{accelerator $h$} \text{ (W)}. \nonumber
\end{equation}
$p_\text{accelerator $h$}$ is the power draw of hardware $h$. We can compare the idealized energy cost of running a specific inference query to the energy cost of training a full model end-to-end to better understand the number of inference queries needed to amortize the significant overhead of training models.

\begin{figure}[t!]
    \centering
    \begin{subfigure}[c]{\columnwidth}
        \centering
        \includegraphics[keepaspectratio=1.0,width=0.85\columnwidth]{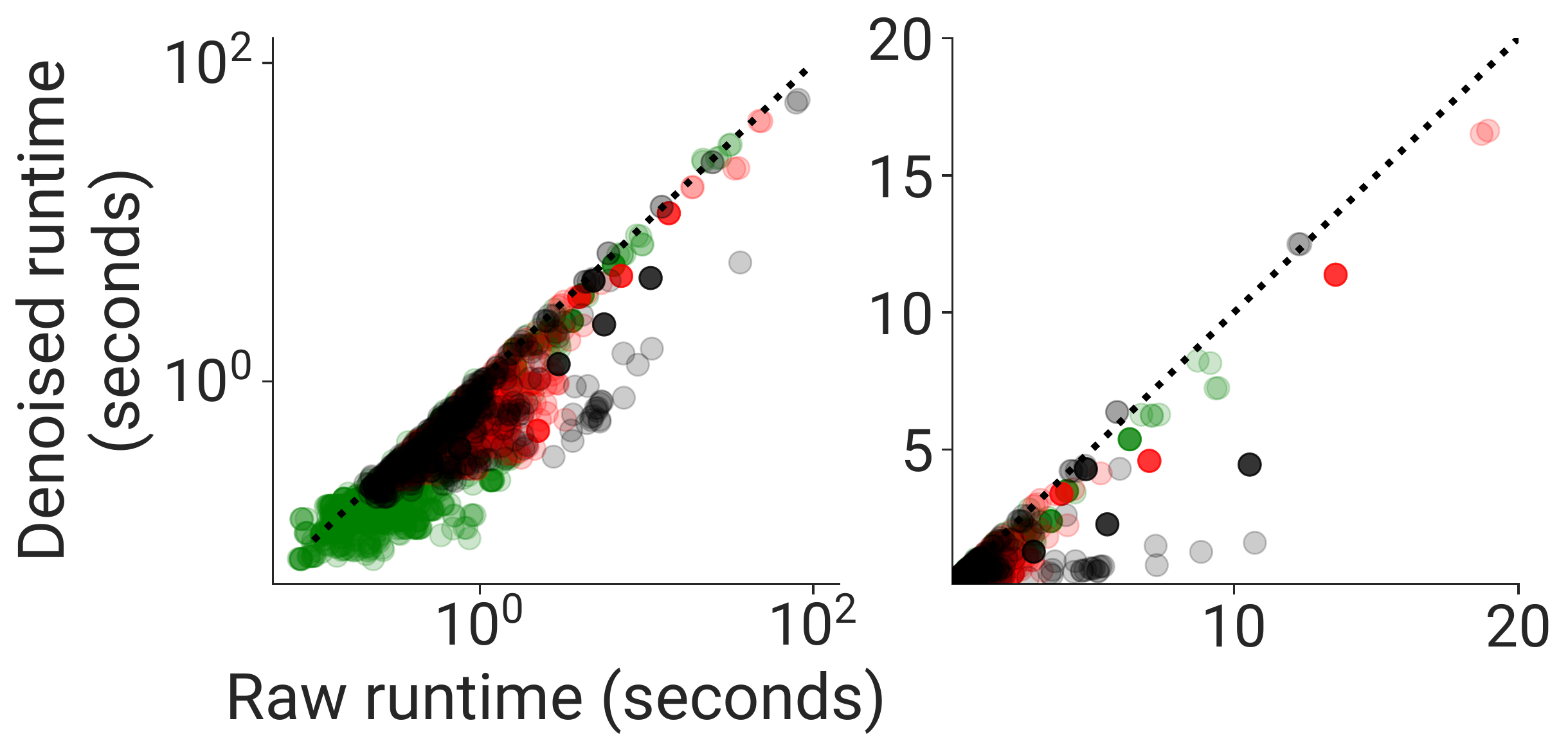}
        \caption{Denoised runtime vs. raw runtime.}
        \label{fig:denoised_runtime_vs_actual_runtime}
    \end{subfigure}
    \centering
    \begin{subfigure}[c]{\columnwidth}
        \centering
        \includegraphics[keepaspectratio=1.0,width=0.85\columnwidth]{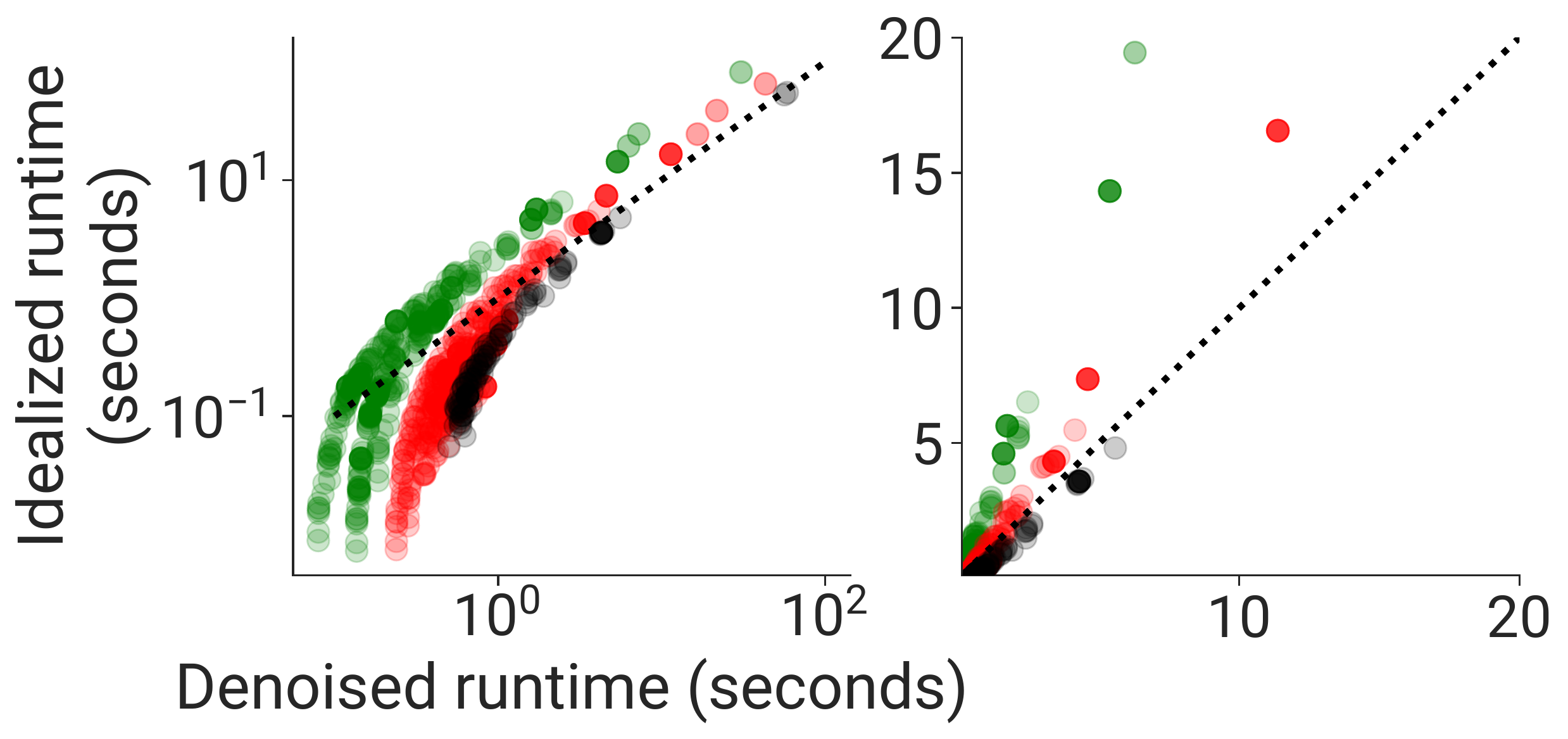}
        \caption{Idealized runtime vs. denoised runtime.}
        \label{fig:idealized_runtime_vs_denoised_runtime}
    \end{subfigure}
    \caption{
        Denoised vs. raw runtime and idealized vs. denoised runtime for various models across a range of queries along with a dotted $y=x$ line. Points corresponding to OpenAI models are shown in green, points corresponding to AI21 Labs models are shown in red, and points corresponding to all remaining models are shown in black.
    }
    \vspace{-0.1in}
    \label{fig:estimated_runtime_vs_actual_runtime}
    \vspace{-0.05in}
\end{figure}

\begin{table}[t!]
\centering
\begin{tabular}{lc}
\toprule
Model (owner/name) & $R^2$ \\
\toprule
\gptdavinci & 0.985 \\ 
\midrule 
\jurassiclarge  & 0.990 \\
\jurassicgrande & 0.917 \\
\jurassicjumbo  & 0.995 \\
\midrule
\coherexl & 0.997 \\
\midrule
\anthropic & 0.924 \\
\bottomrule
\end{tabular}
\caption{Models and coefficient of determination ($R^2$) of time estimates for end-to-end text generation for various models using black-box APIs.}
\label{table:r_squared_real_systems}
\end{table}

\begin{table*}[t]
\centering
\begin{tabular}{llcc}
\toprule
Model (owner/name)       & Metric  & $\textsf{\small prompt\_encoding\_time}$ & Per-output-token \\
                         &         & $(p=512 / 1024 / 1536)$ in secs & generation time ($g$) in secs \\
\toprule
\gptdavinci      & $t_{(m, \text{ Megatron}, \text{ A100})}^\text{idealized}$ & 0.178 / 0.323 / 0.476 & 0.081 \\
                & $t_{m}^\text{denoised}$       & 0.045 / 0.033 / 0.142 & 0.030 \\
\midrule
\jurassicgrande  & $t_{(m, \text{ Megatron}, \text{ A100})}^\text{idealized}$ & 0.097 / 0.190 / 0.298 & 0.038 \\
                 & $t_{m}^\text{denoised}$       & 0.172 / 0.351 / 0.519 & 0.021 \\
\midrule
\jurassicjumbo  & $t_{(m, \text{ Megatron}, \text{ A100})}^\text{idealized}$ & 0.164 / 0.310 / 0.465 & 0.064  \\
                & $t_{m}^\text{denoised}$       & 0.268 / 0.463 / 0.655 & 0.042 \\
\midrule
\anthropic      & $t_{(m, \text{ Megatron}, \text{ A100})}^\text{idealized}$ & 0.108 / 0.189 / 0.279 & 0.054 \\
                & $t_{m}^\text{denoised}$       & 0.193 / 0.191 / 0.380 & 0.057 \\
\bottomrule
\end{tabular}
\caption{Models and estimated prompt encoding times / per-output-token generation times for $t_{(m, \text{ Megatron}, \text{ A100})}^\text{idealized}$ and $t_{m}^\text{denoised}$.}
\label{table:learnt_parameters_comparison}
\end{table*}

\section{Results}

In this section, we seek to empirically answer the following:
\begin{itemize}
    \item Is the proposed methodology to estimate inference runtime of autoregressive Transformer models accurate?
    \item Is it efficient compared to exhaustive profiling?
    \item Can this method reveal interesting insights about models' efficiency-capability tradeoffs?
\end{itemize}

\subsection{Evaluated Models}

We evaluate 10 different models, ranging in size from 6 to 530 billion parameters (see Table~\ref{table:models} for more details), and focus on the few-shot evaluation setting, similar to other benchmarks for LLMs like BIG-Bench~\cite{srivastava2022beyond} and HELM~\cite{liang2022holistic}. The covered models are available in different ways: some were public via a commercial API (e.g., \gptdavinci, \jurassicjumbo), some were private but the model owner provided research access for this effort (\anthropic, \mtnlg), and some were public and free (e.g., \yalm, \bloom) and were run using the Together Open Models API\footnote{\url{https://www.together.xyz/}.}. We do not evaluate models with publicly unavailable model architecture details (including OpenAI's \texttt{ChatGPT} and \texttt{GPT-4}).

\subsection{Accuracy of Runtime Estimation Procedure}

Table~\ref{table:r_squared_real_systems} shows the coefficients of determination for runtimes using black-box APIs. Despite performance variance, we see that the estimated runtimes using the methodology based on linear regression outlined in \S\ref{sec:estimation} are fairly accurate, lending credence to the accuracy of our closed-form expressions for autoregressive inference runtime of Transformer models.

Figure~\ref{fig:denoised_runtime_vs_actual_runtime} compares denoised runtimes to raw runtimes for a range of prompt sizes and number of generated output tokens. We observe that raw runtimes for the most part (96.6\% of points) are greater than the estimated denoised runtimes (below the $y=x$ dotted line), indicating that the denoised runtimes in practice are a good lower bound for actual runtime obtained using black-box APIs.
Figure~\ref{fig:idealized_runtime_vs_denoised_runtime} is similar, but shows idealized runtime with A100 GPUs and NVIDIA's Megatron~\cite{shoeybi2019megatron} versus denoised runtime. In a number of cases, the idealized runtime is much lower than the denoised runtime, since the relevant API uses slower hardware and / or software implementations. For AI21 Labs models, idealized runtimes are greater than denoised runtimes 15.7\% of the time. For OpenAI models, idealized runtimes are greater than denoised runtimes 64.2\% of the time. For all other models, idealized runtimes are always lower than the denoised runtimes, indicating that our hardware and software stack assumptions were fairly accurate for other model providers.

Table~\ref{table:learnt_parameters_comparison} compares the learnt performance parameters for $t_{(m, \text{ Megatron}, \text{ A100})}^\text{idealized}$ and $t_{m}^\text{denoised}$ for a subset of the considered models. As noted above, the estimated ``(Megatron, A100) idealized'' parameters for the AI21 Labs and OpenAI models are higher than the estimated denoised parameters, indicating that both these providers have implemented optimizations not present in the software stacks we considered.

\subsection{Efficiently Evaluating Other Hardware}

\begin{figure*}[ht!]
    \centering
    \includegraphics[keepaspectratio=1.0,width=1.03\textwidth]{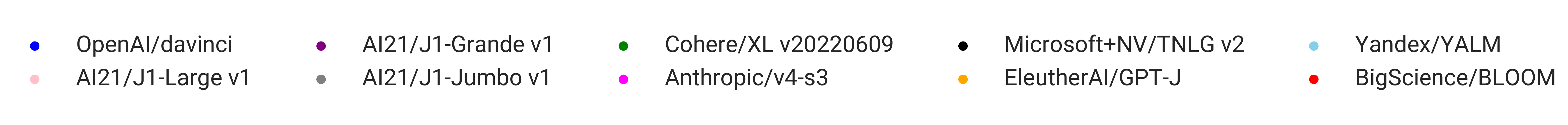}
    \begin{subfigure}[c]{0.49\columnwidth}
        \centering
        \includegraphics[keepaspectratio=1.0,width=\columnwidth]{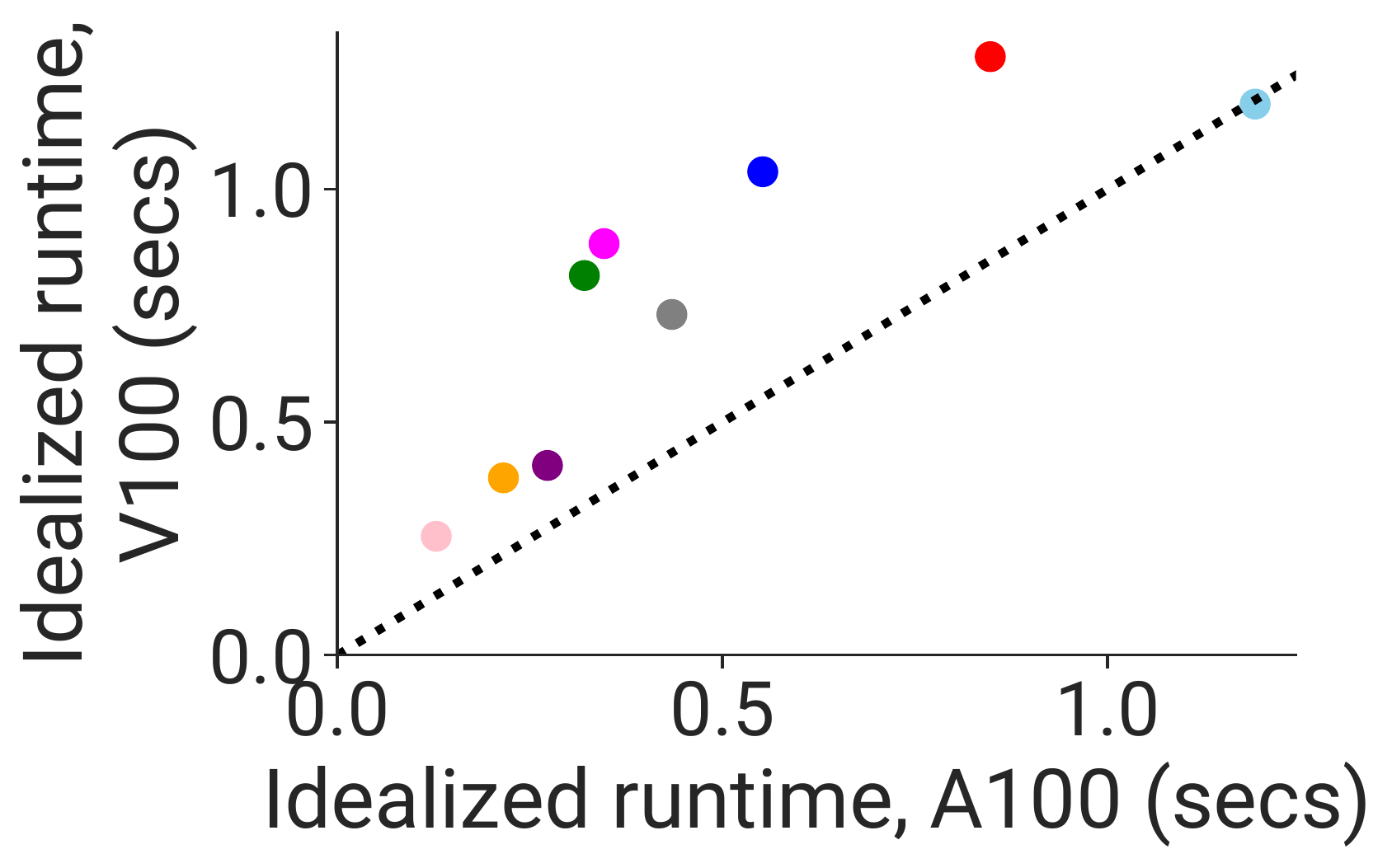}
        \includegraphics[keepaspectratio=1.0,width=\columnwidth]{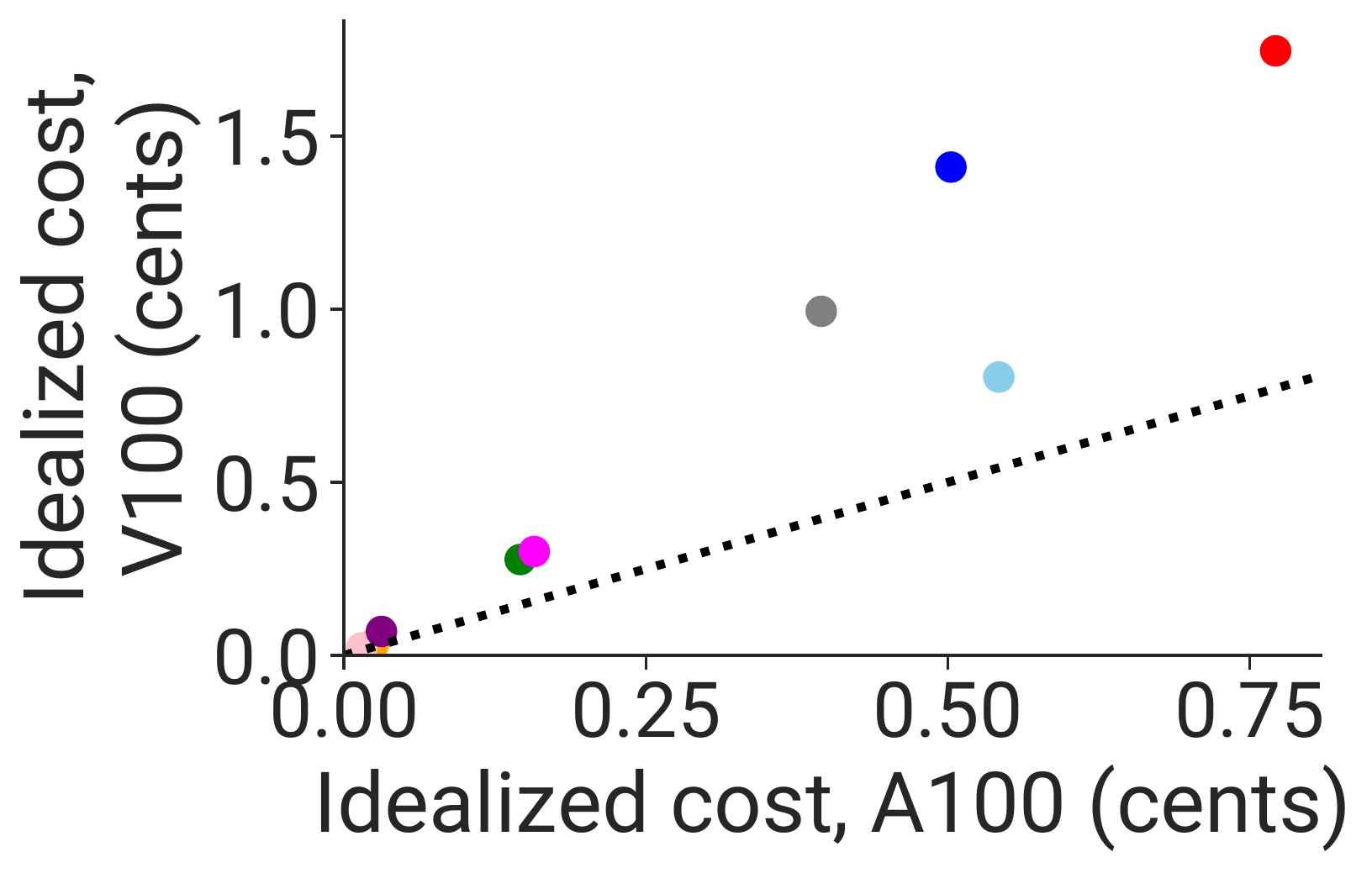}
        \caption{IMDB.}
    \end{subfigure}
    \centering
    \begin{subfigure}[c]{0.49\columnwidth}
        \centering
        \includegraphics[keepaspectratio=1.0,width=\columnwidth]{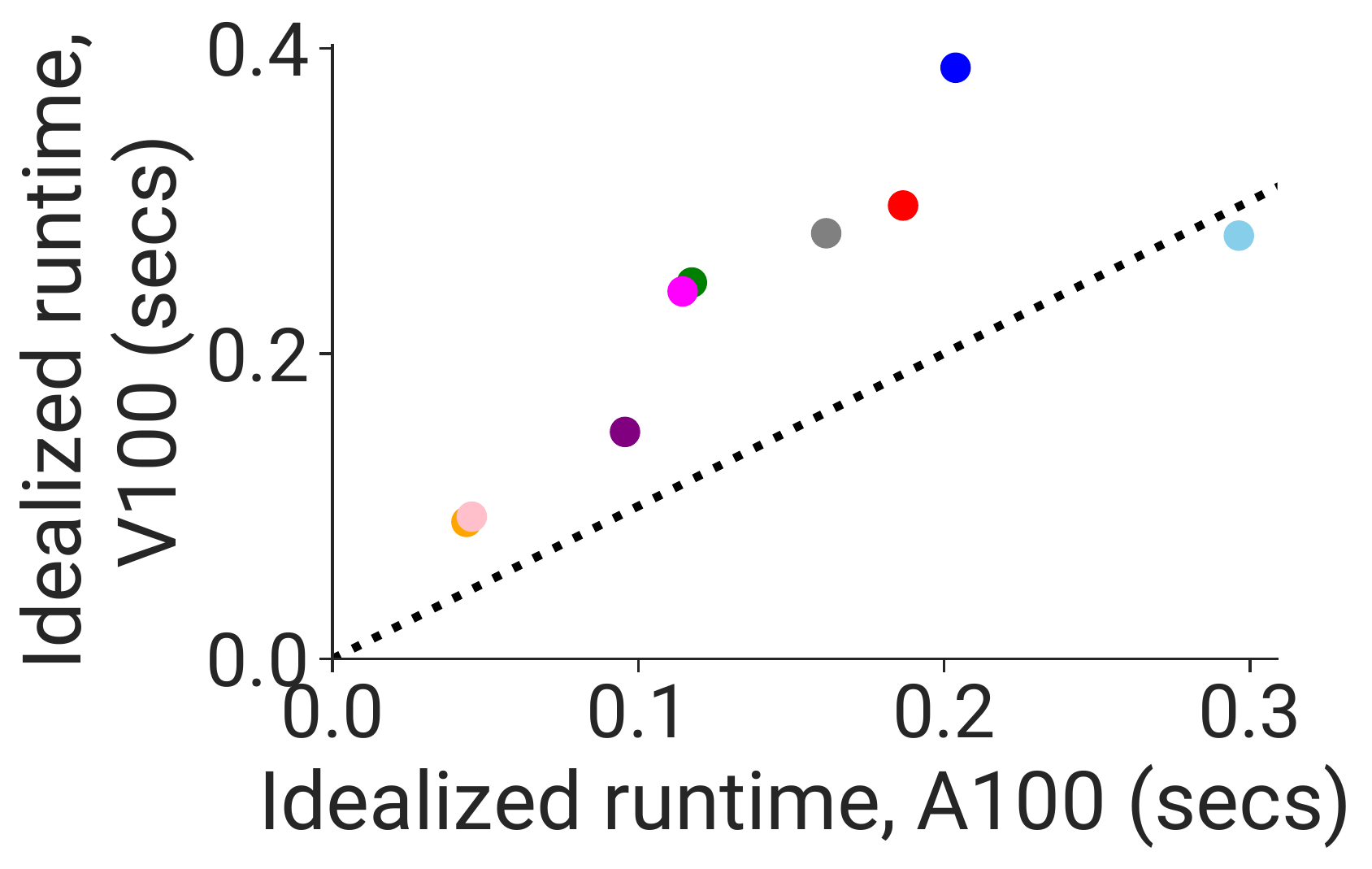}
        \includegraphics[keepaspectratio=1.0,width=\columnwidth]{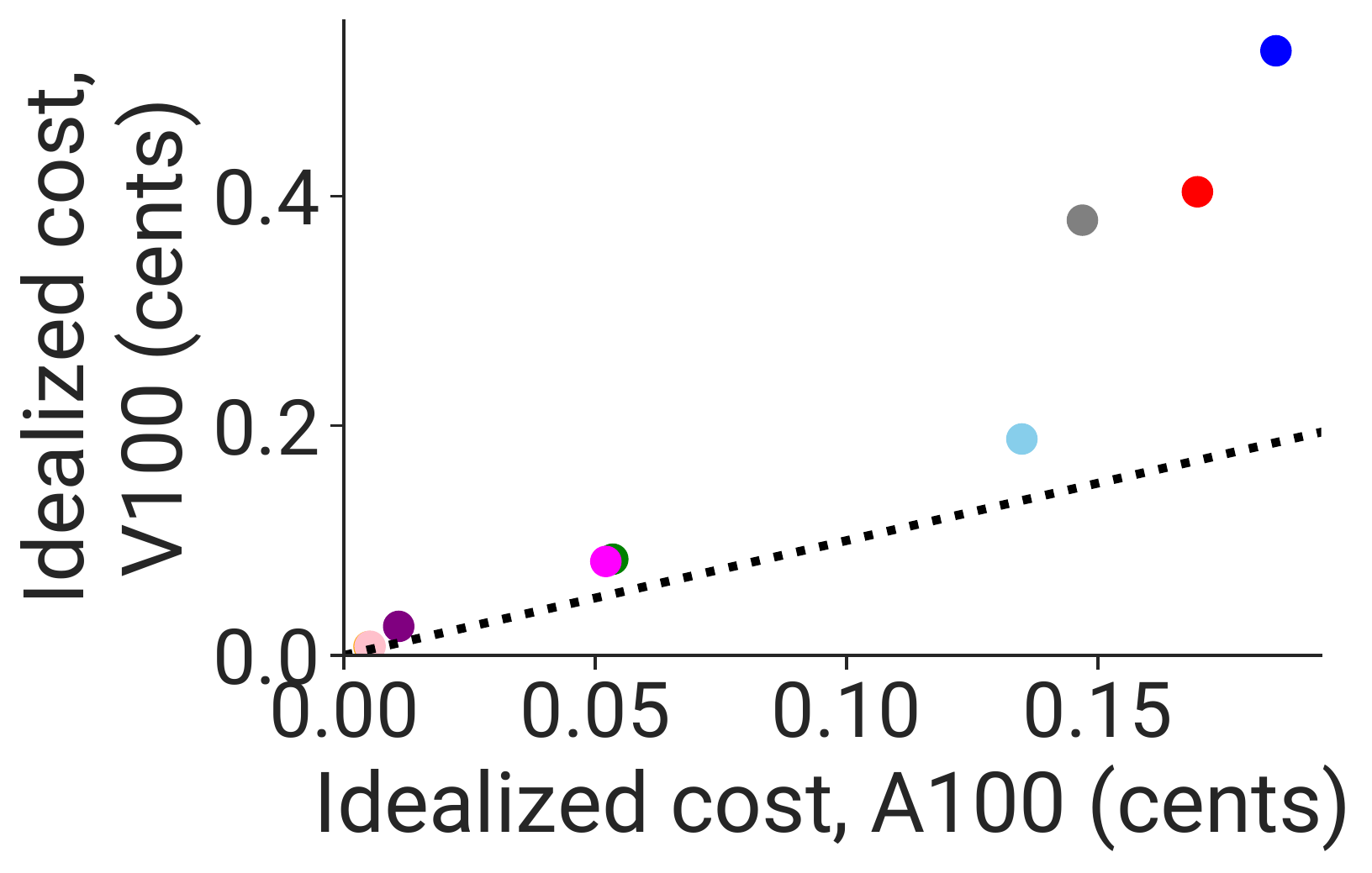}
        \caption{MMLU, college chemistry.}
    \end{subfigure}
    \centering
    \begin{subfigure}[c]{0.49\columnwidth}
        \centering
        \includegraphics[keepaspectratio=1.0,width=\columnwidth]{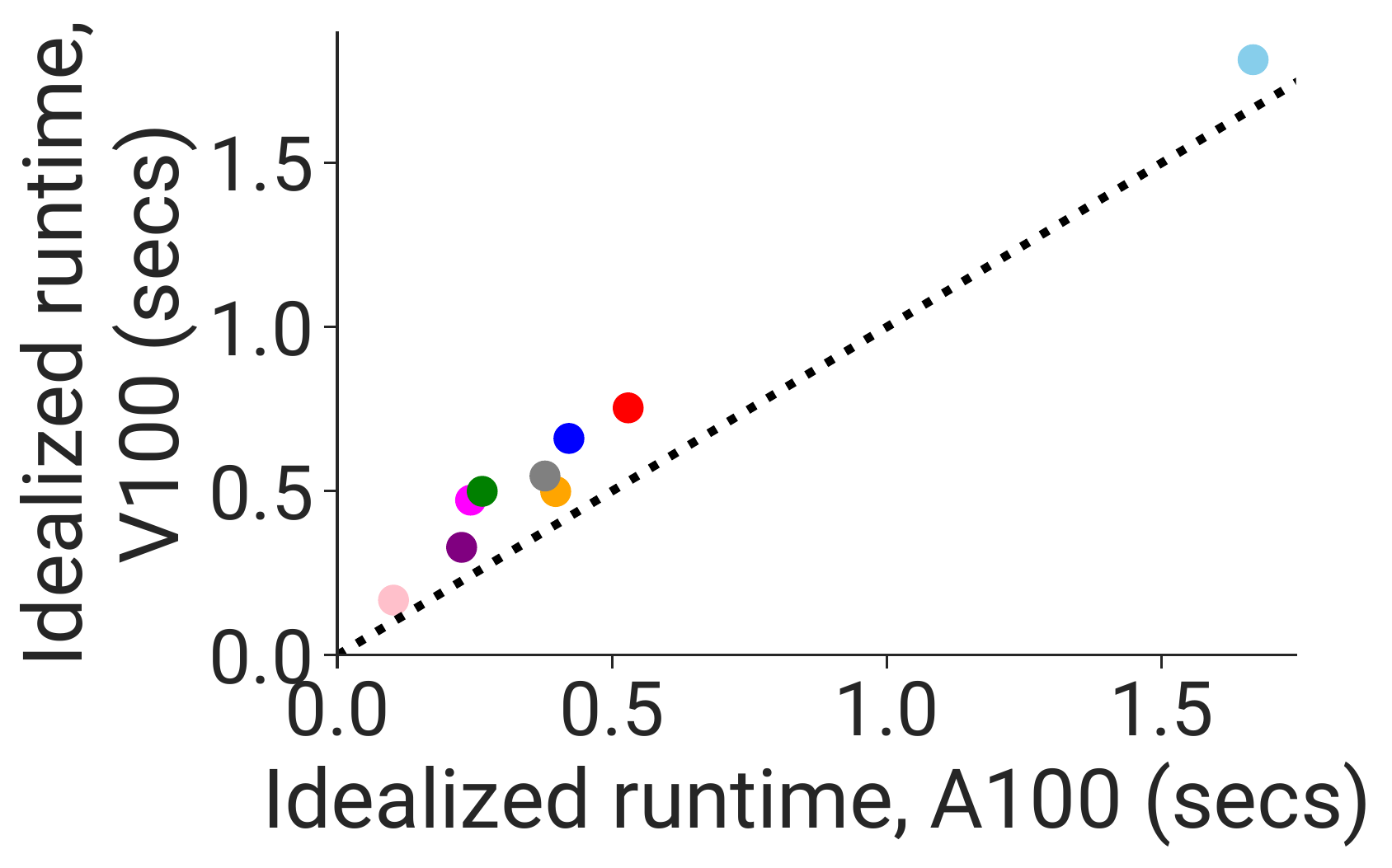}
        \includegraphics[keepaspectratio=1.0,width=\columnwidth]{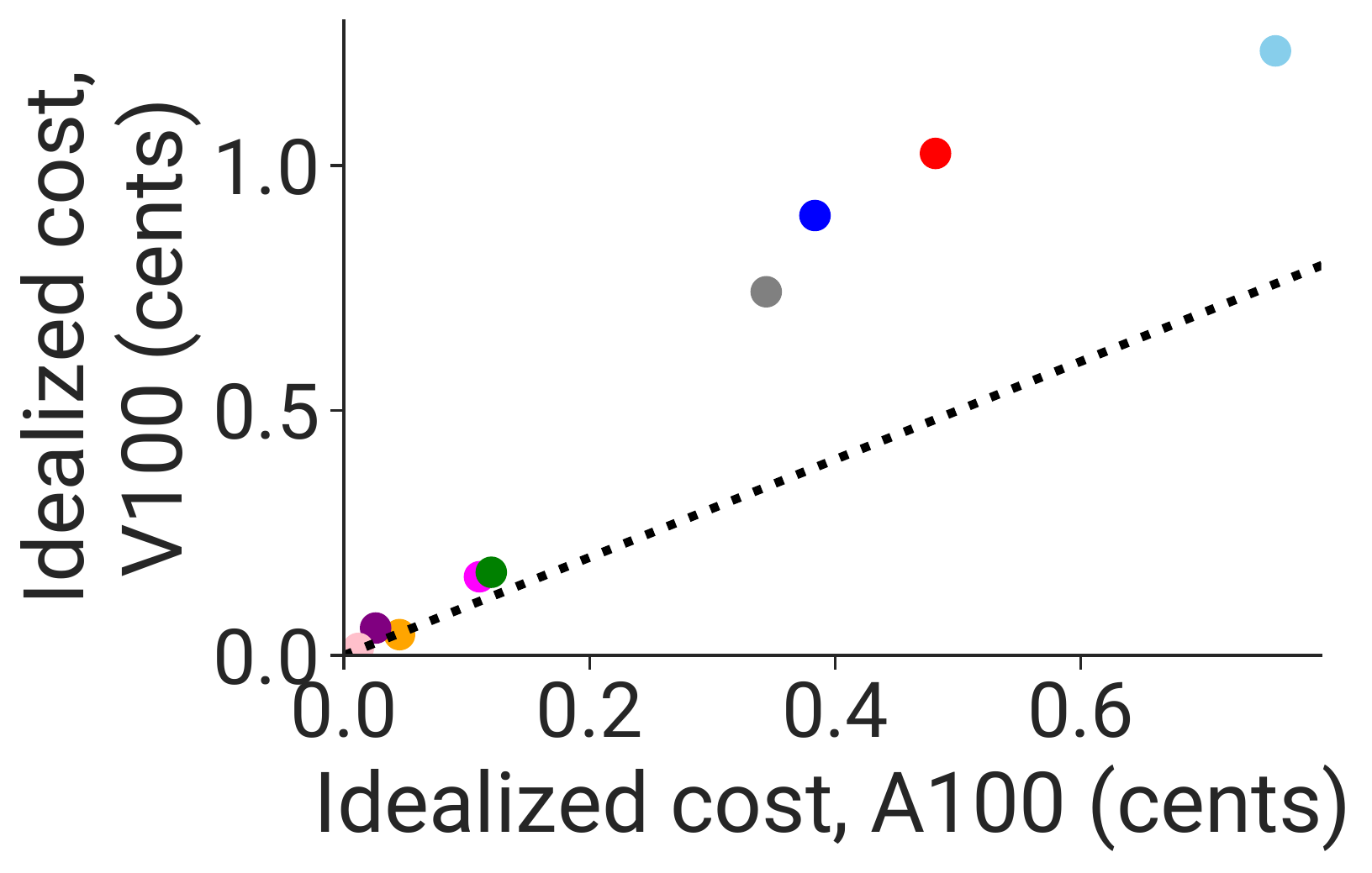}
        \caption{RAFT, terms of service.}
    \end{subfigure}
    \centering
    \begin{subfigure}[c]{0.49\columnwidth}
        \centering
        \includegraphics[keepaspectratio=1.0,width=\columnwidth]{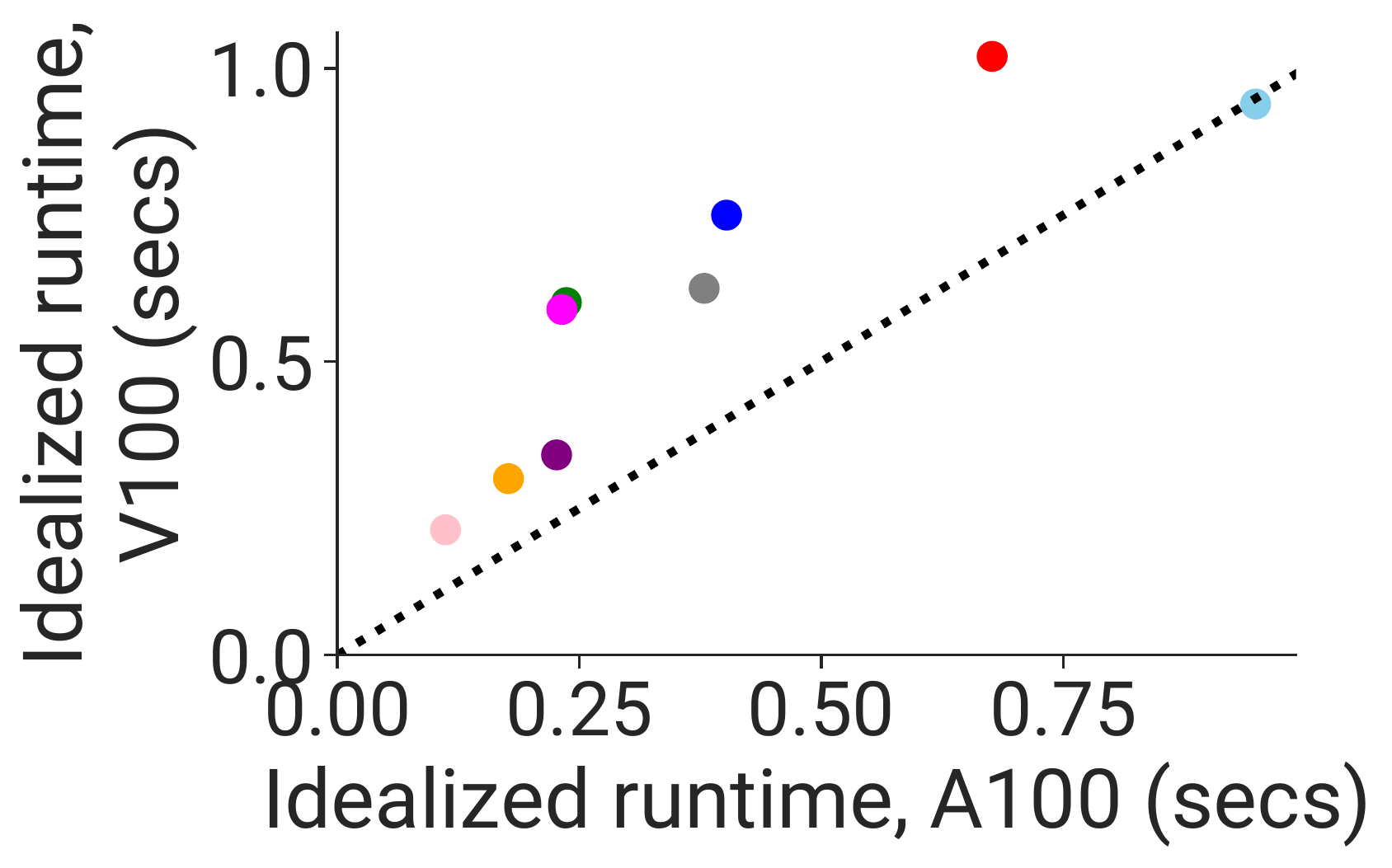}
        \includegraphics[keepaspectratio=1.0,width=\columnwidth]{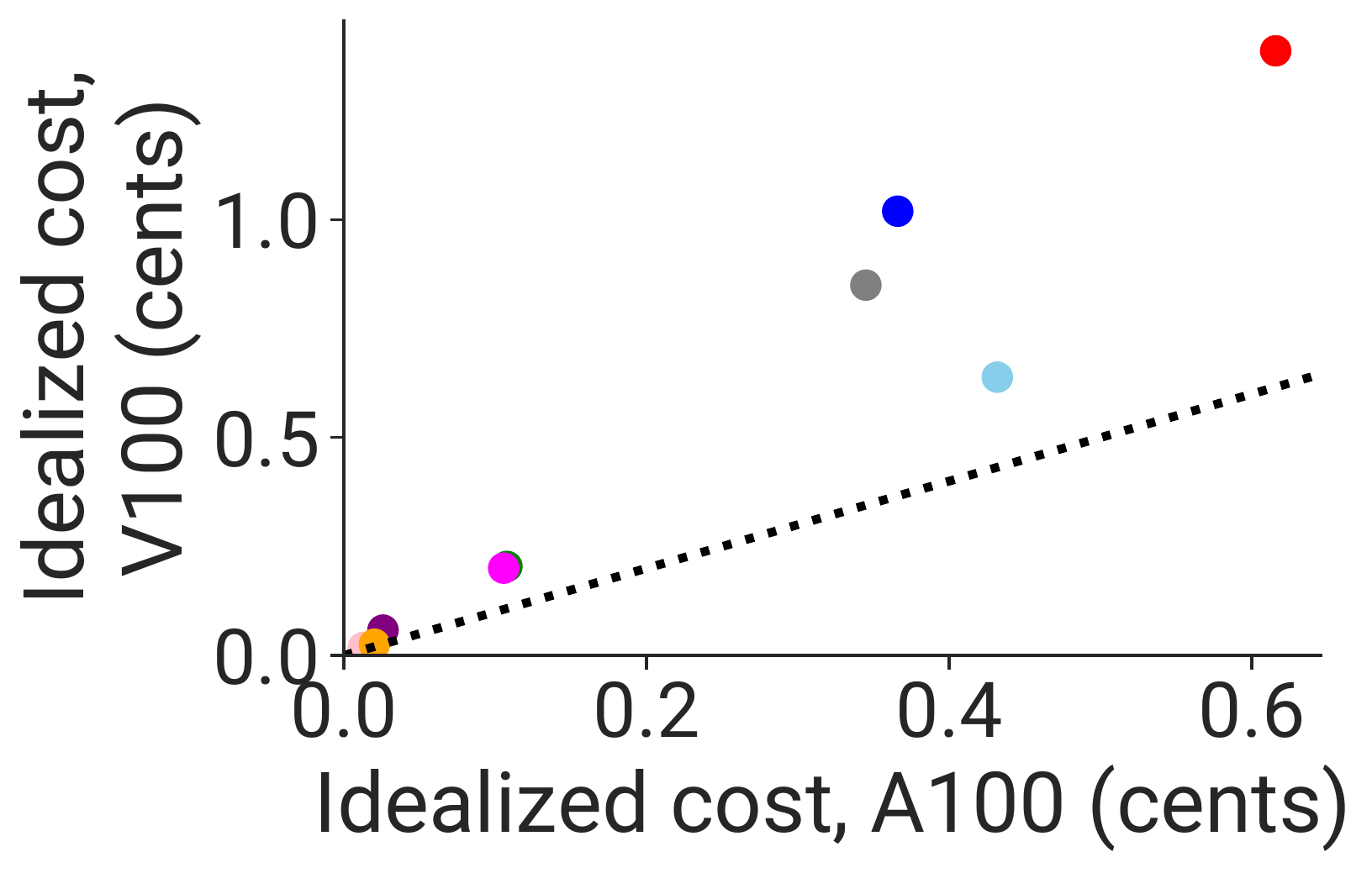}
        \caption{BoolQ.}
    \end{subfigure}
    \caption{
        Comparison of idealized metrics estimated on different hardware.
    }
    \label{fig:idealized_comparison}
\end{figure*}

We can use the methodology proposed in this paper to evaluate the efficacy of other hardware and software solutions for serving of autoregressive Transformer models. For example, Figure~\ref{fig:idealized_comparison} shows a comparison between Megatron on Nvidia A100 GPUs (the default configuration in this paper) to Megatron on Nvidia V100 GPUs (an older generation of Nvidia GPUs). While we expect these GPUs to be slower, we can also reasonably expect them to be cheaper (due to cheaper per-hour costs~\cite{awspricing}). In practice, we find that this is \emph{not the case}, suggesting V100 GPUs are both slower and more expensive. This is partially because we often have to use double the GPUs to fit the model parameters in GPU memory, since V100 GPUs only have 32GB of device memory compared to 80GB on the A100 GPUs.

This differential analysis with our methodology requires profiling on the order of hours ($<2$ hours for most models, depending on the number of $(p, o)$ values profiled) \emph{once}, compared to hours \emph{per benchmark} (depending on number of queries in the benchmark) for exhaustive profiling.

Our analyses are not constrained to evaluating how fast inference queries could be processed on other types of hardware accelerators (e.g., TPUs). We can perform similar analysis for different software stacks as well (e.g., Nvidia Triton or Megatron with FlashAttention~\cite{dao2022flashattention} enabled).

\subsection{Efficiency-Capability Tradeoffs}
\label{sec:tradeoffs}

\begin{figure*}[ht!]
    \centering
    \includegraphics[keepaspectratio=1.0,width=1.03\textwidth]{figures/efficiency_vs_capability/idealized_runtime/legend.pdf}
    \begin{subfigure}[c]{0.49\columnwidth}
        \centering
        \includegraphics[keepaspectratio=1.0,width=\columnwidth]{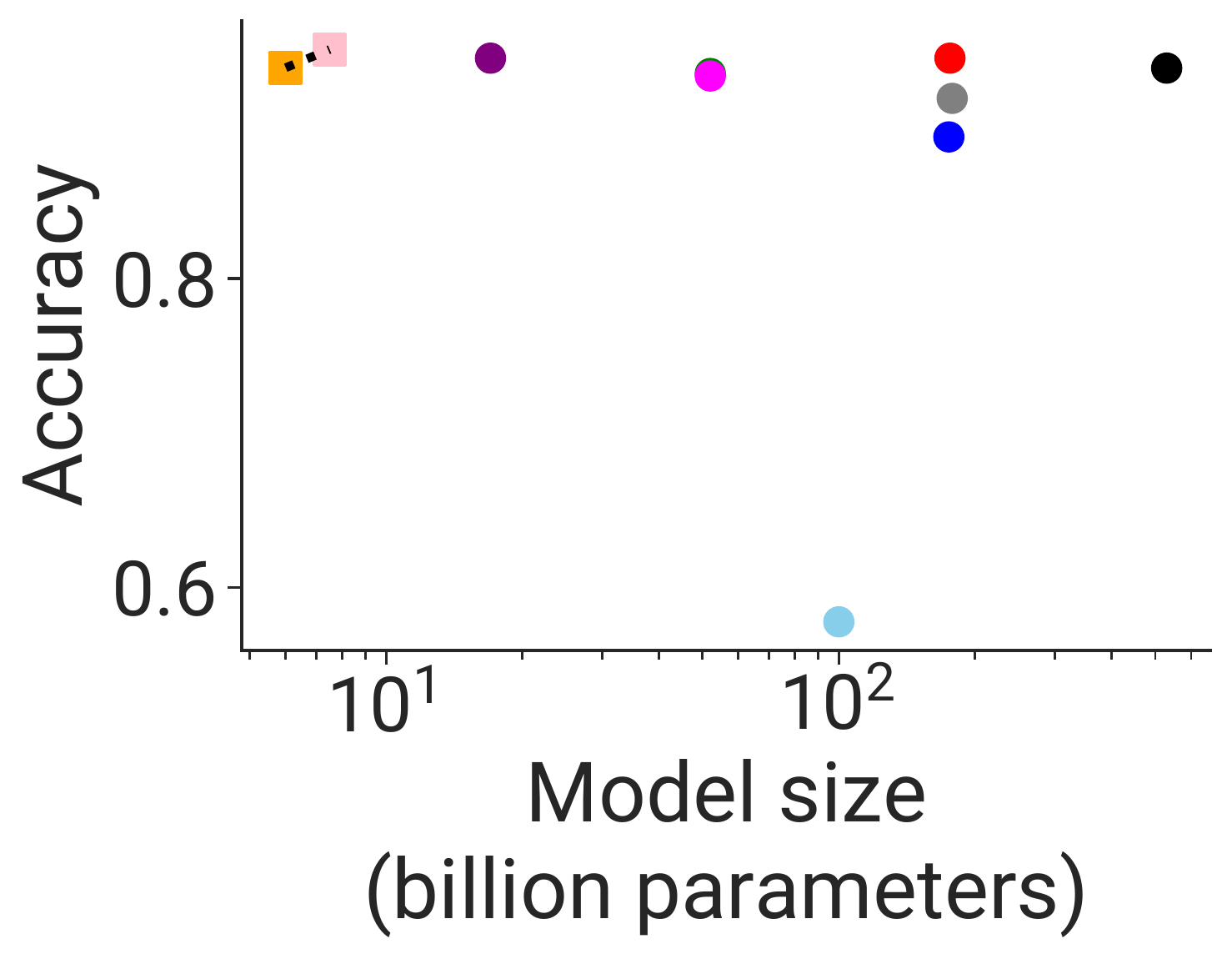}
        \includegraphics[keepaspectratio=1.0,width=\columnwidth]{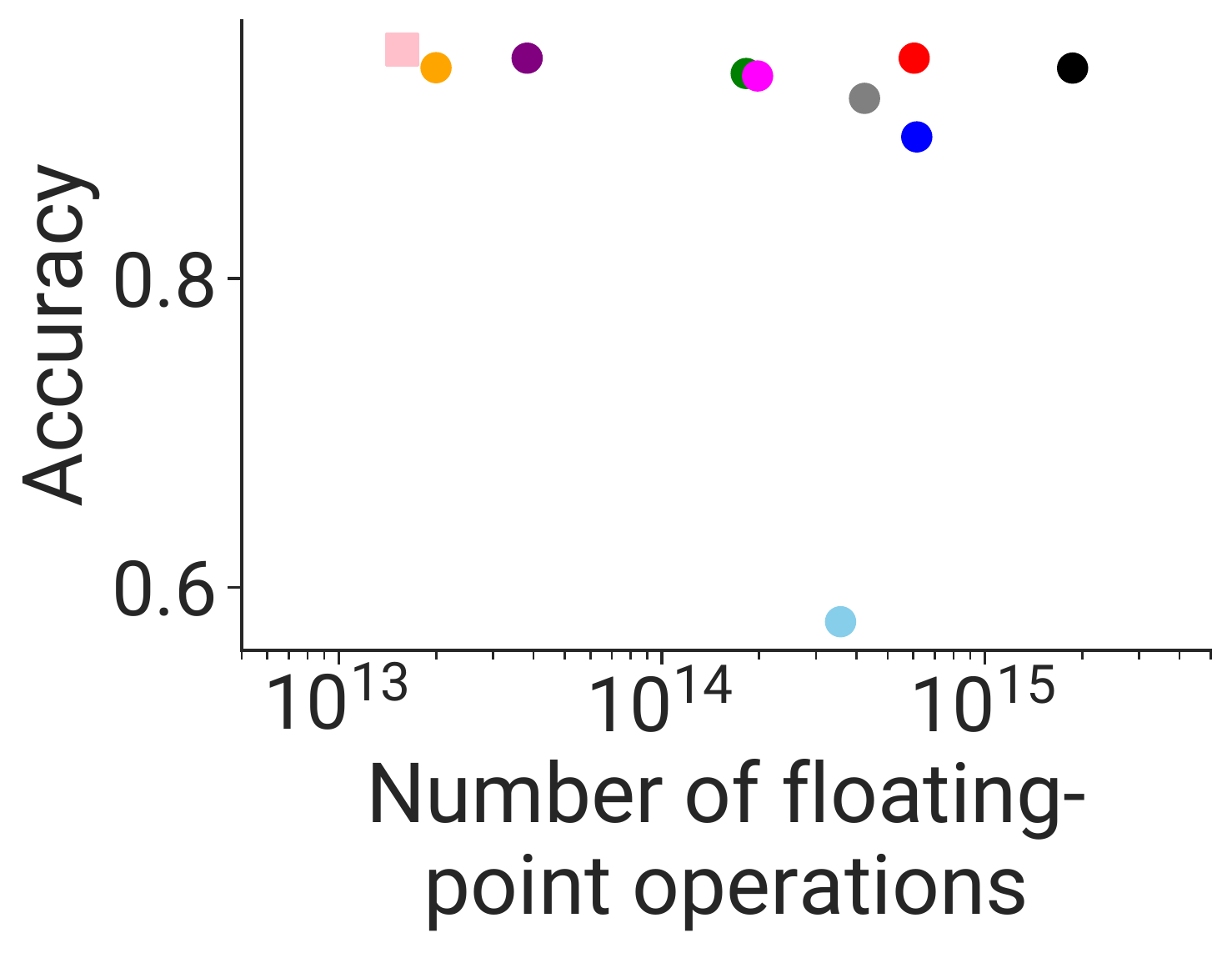}
        \includegraphics[keepaspectratio=1.0,width=\columnwidth]{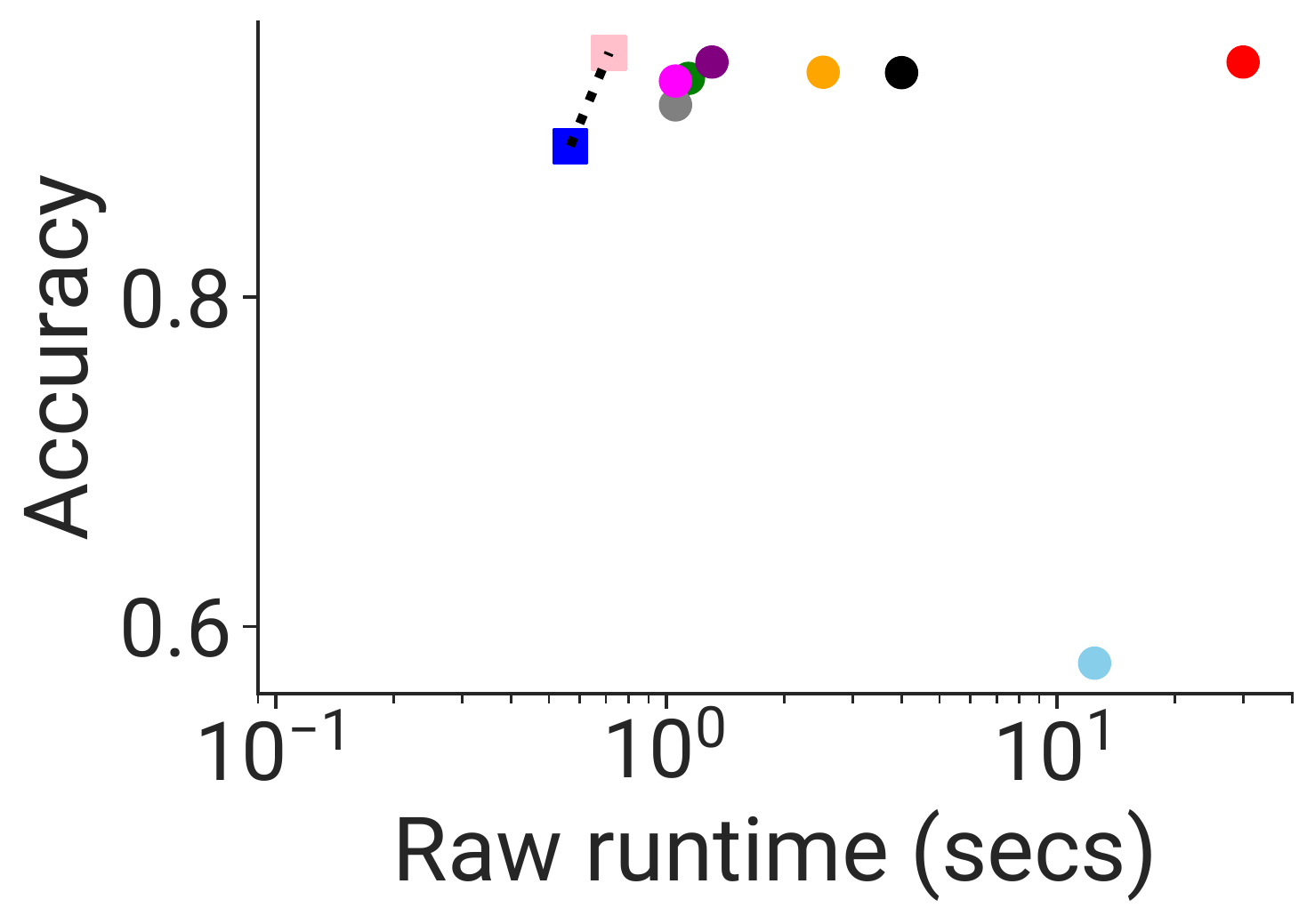}
        \includegraphics[keepaspectratio=1.0,width=\columnwidth]{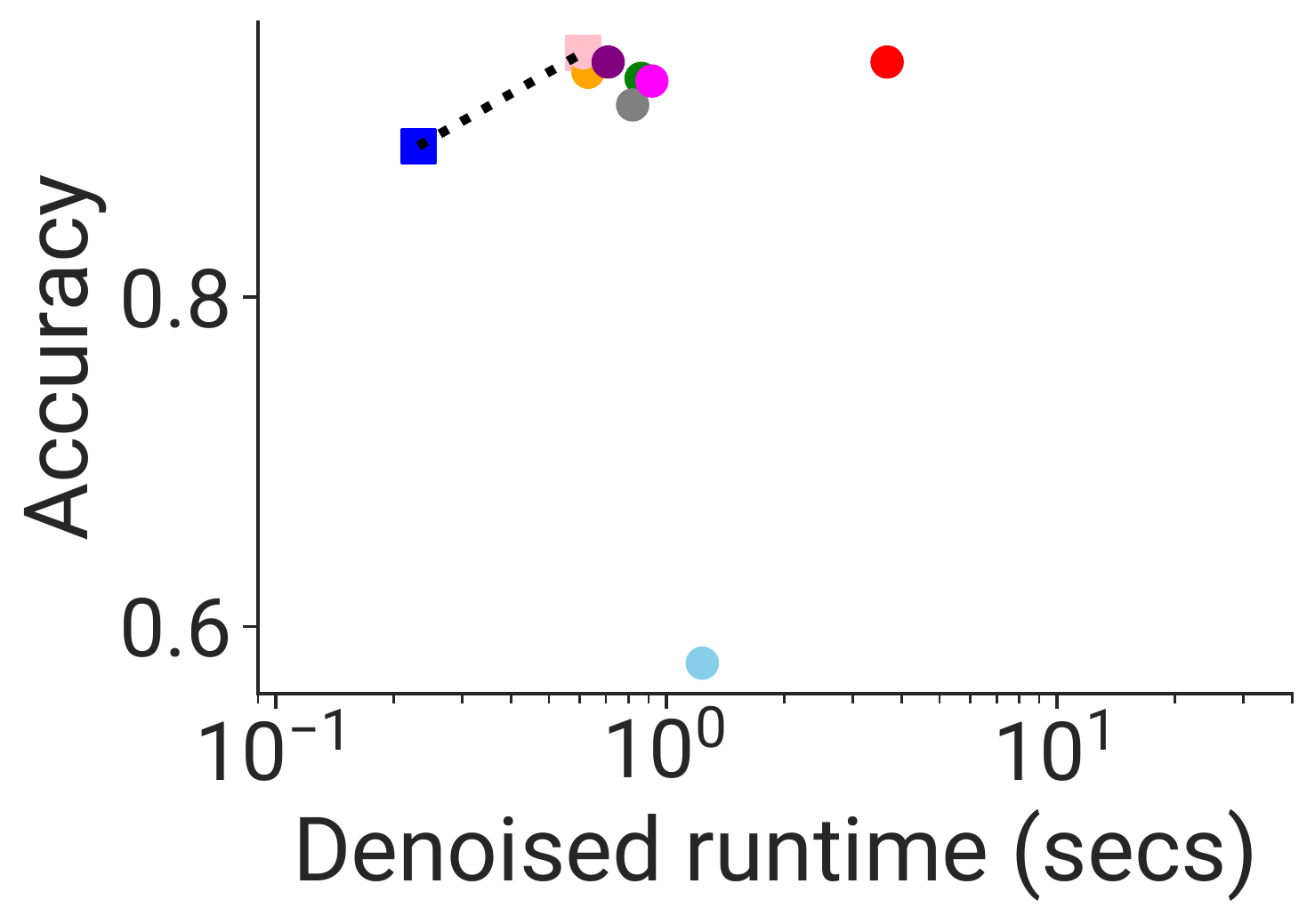}
        \includegraphics[keepaspectratio=1.0,width=\columnwidth]{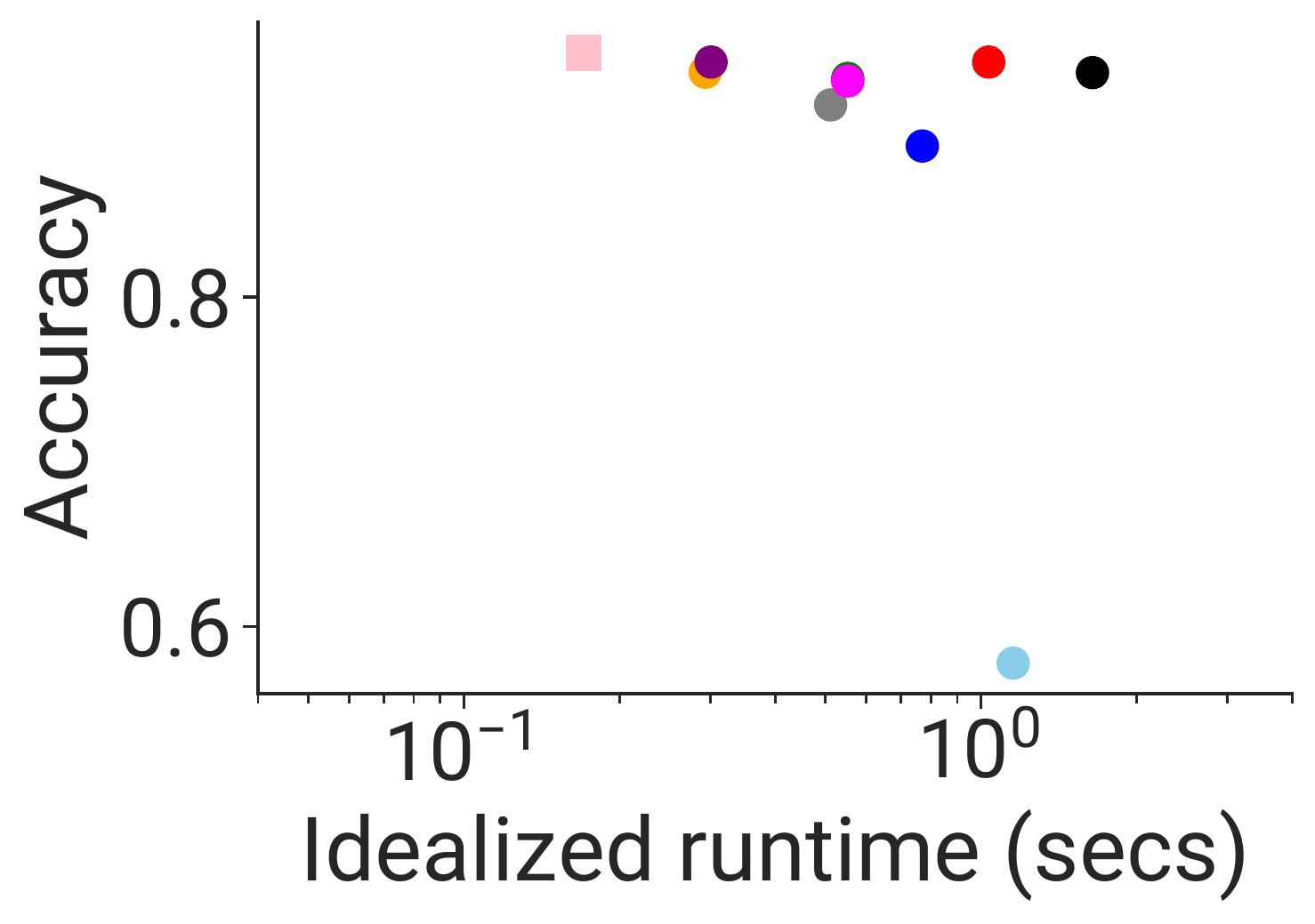}
        \includegraphics[keepaspectratio=1.0,width=\columnwidth]{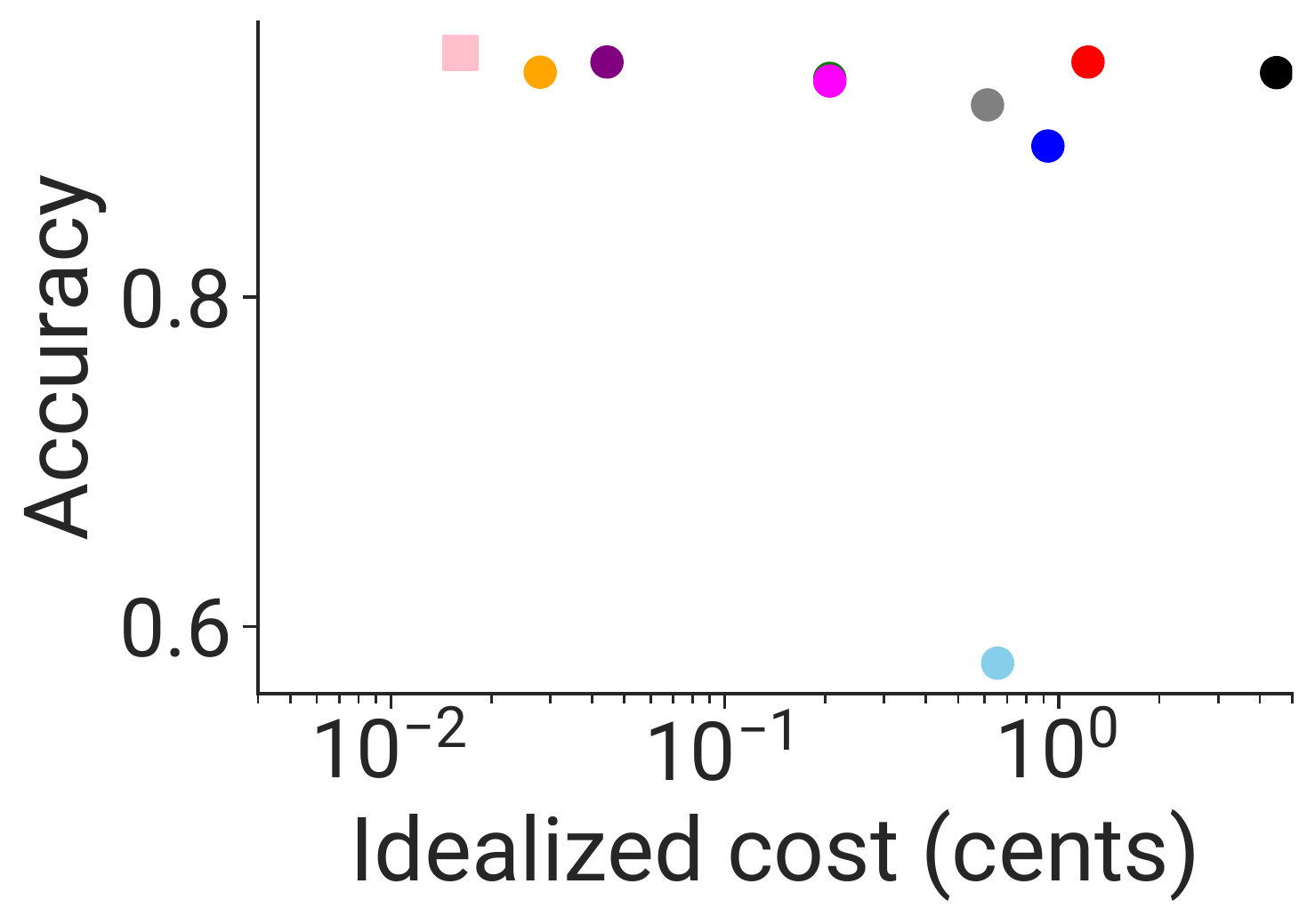}
        \caption{IMDB.}
    \end{subfigure}
    \centering
    \begin{subfigure}[c]{0.49\columnwidth}
        \centering
        \includegraphics[keepaspectratio=1.0,width=\columnwidth]{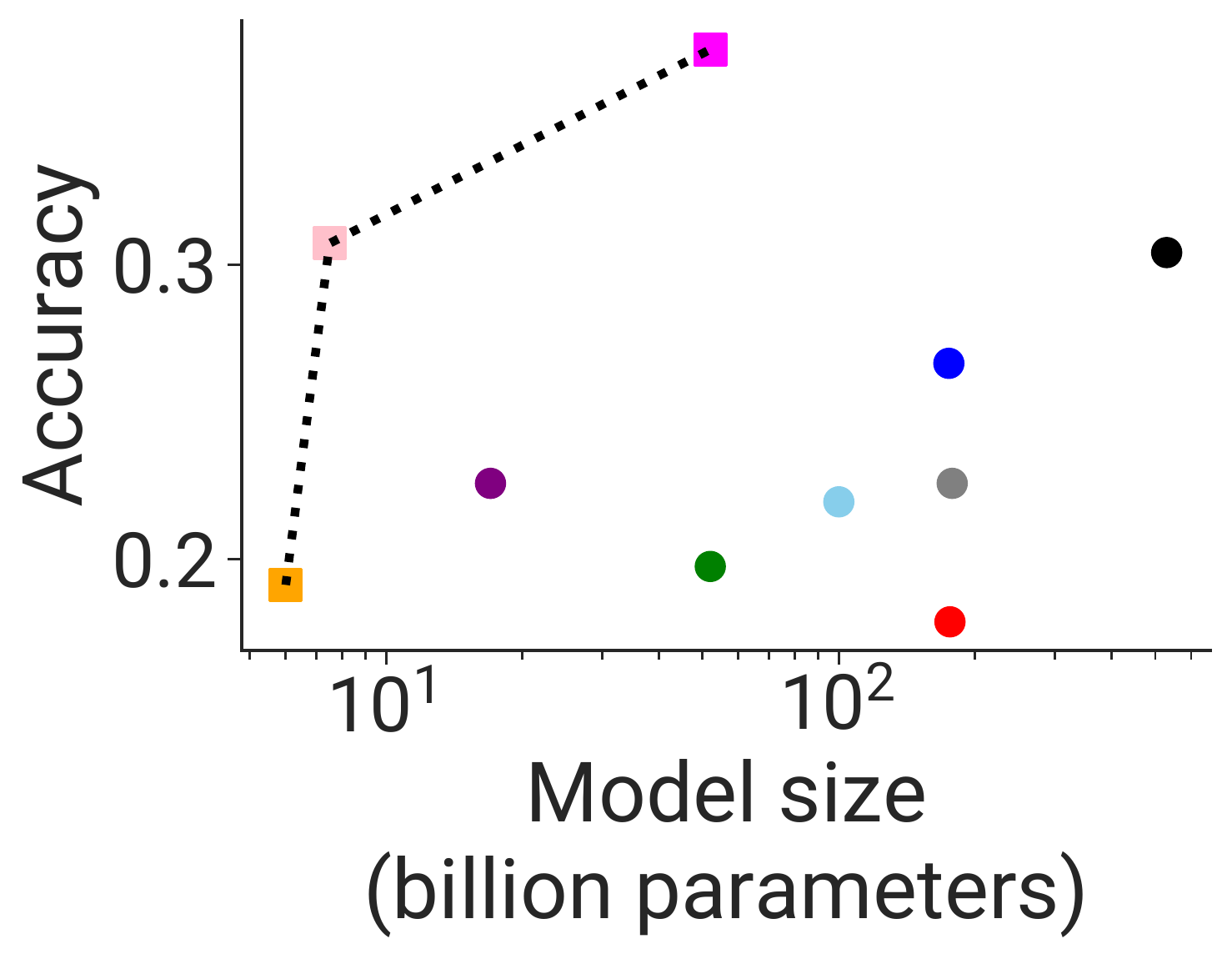}
        \includegraphics[keepaspectratio=1.0,width=\columnwidth]{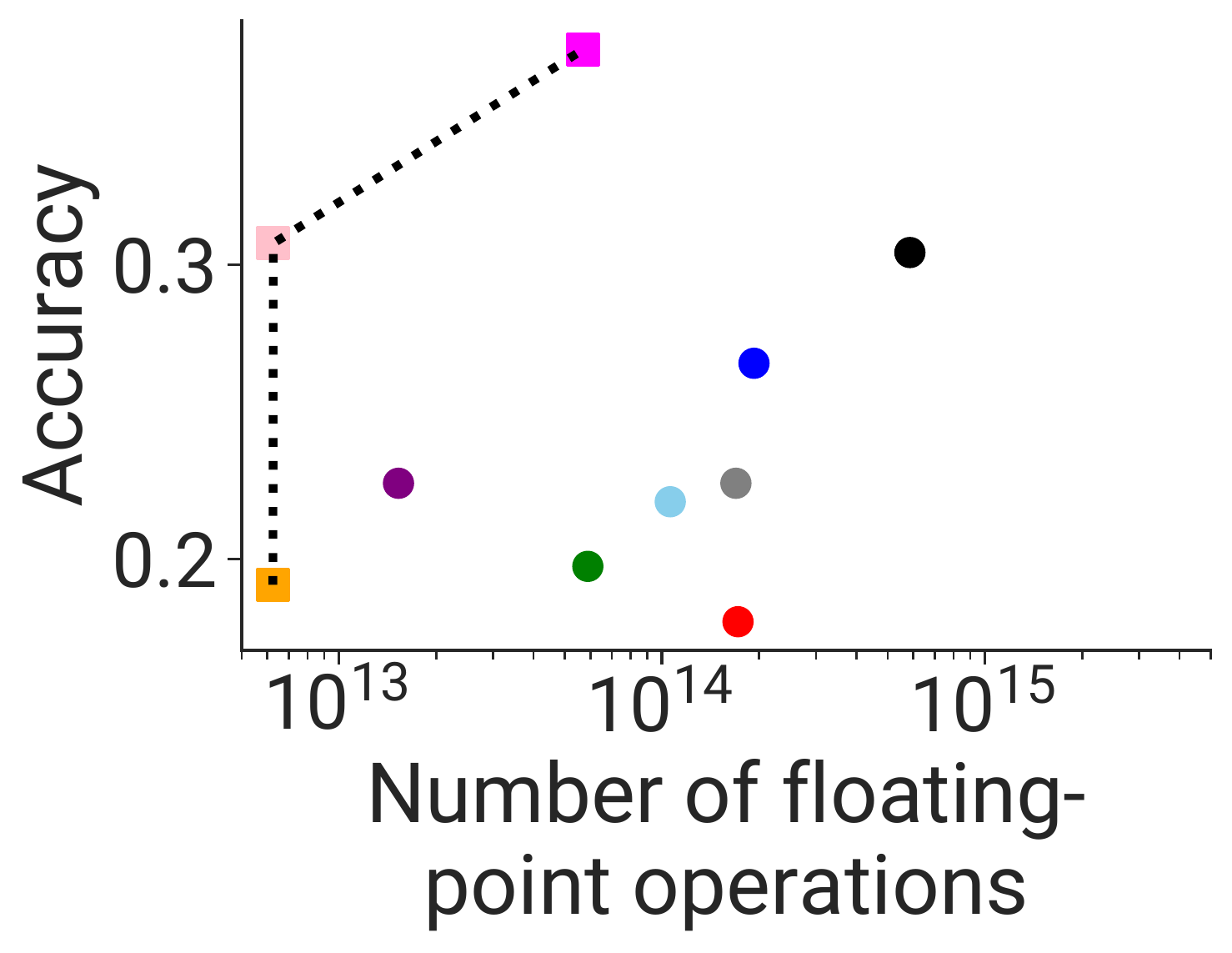}
        \includegraphics[keepaspectratio=1.0,width=\columnwidth]{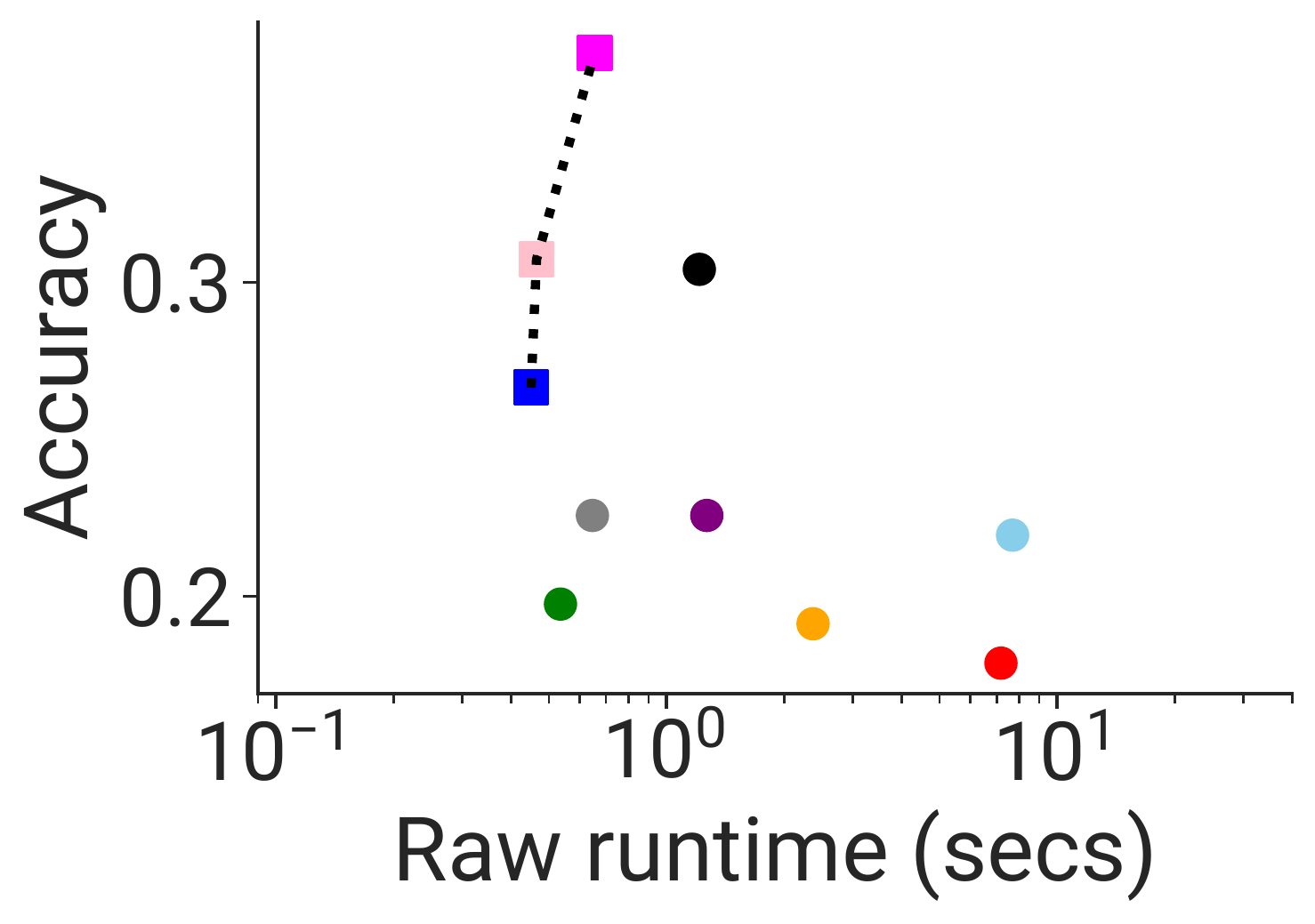}
        \includegraphics[keepaspectratio=1.0,width=\columnwidth]{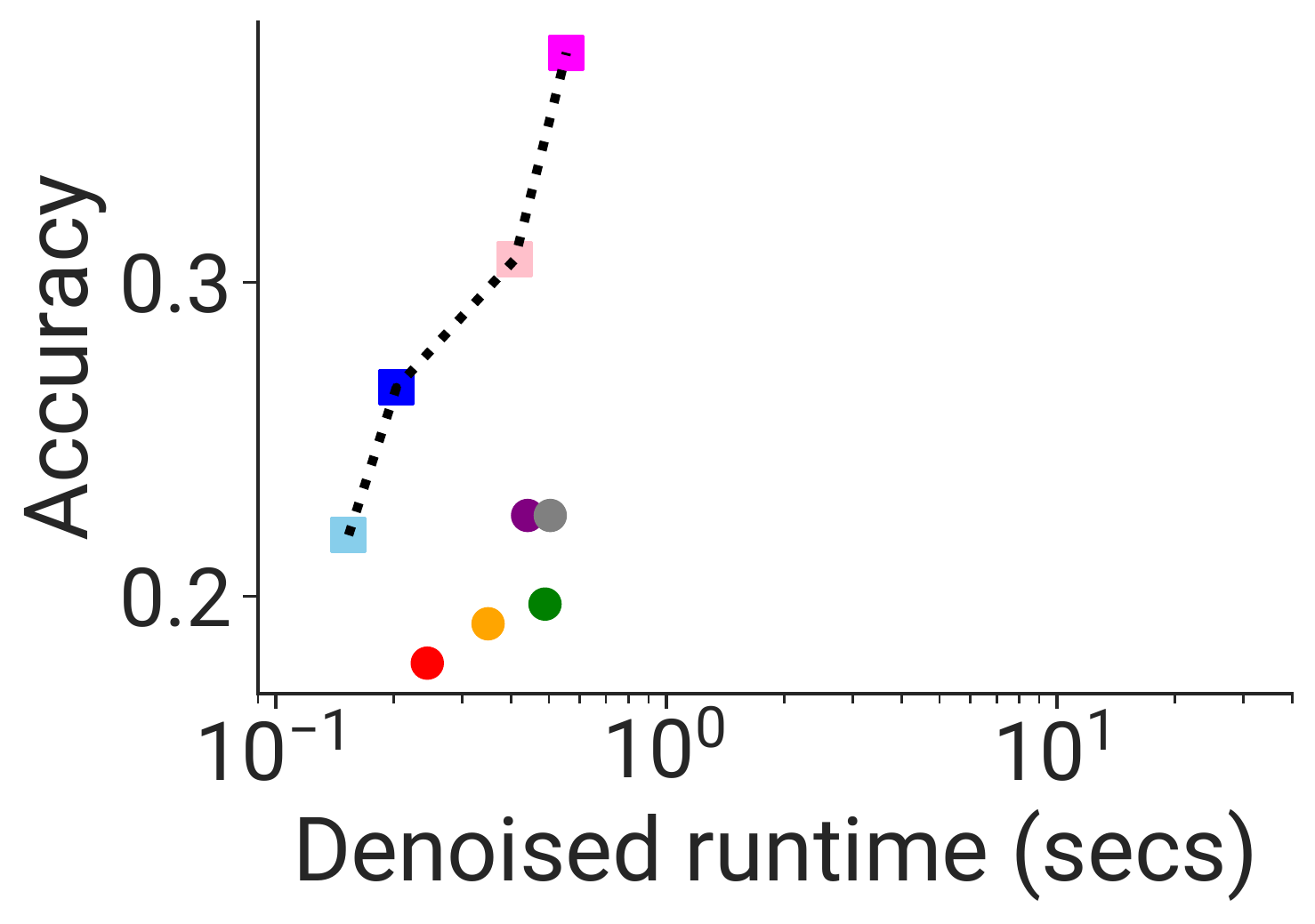}
        \includegraphics[keepaspectratio=1.0,width=\columnwidth]{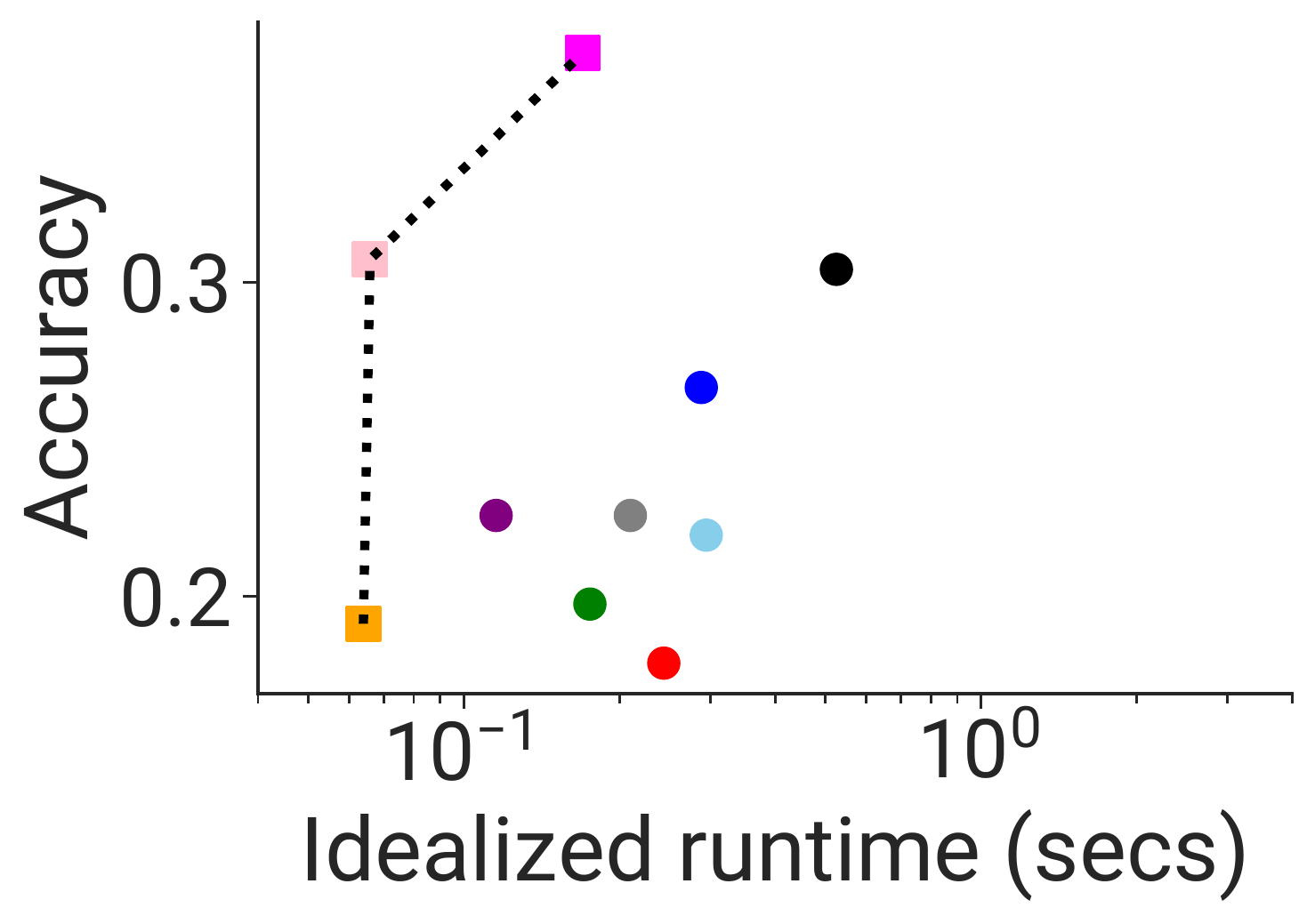}
        \includegraphics[keepaspectratio=1.0,width=\columnwidth]{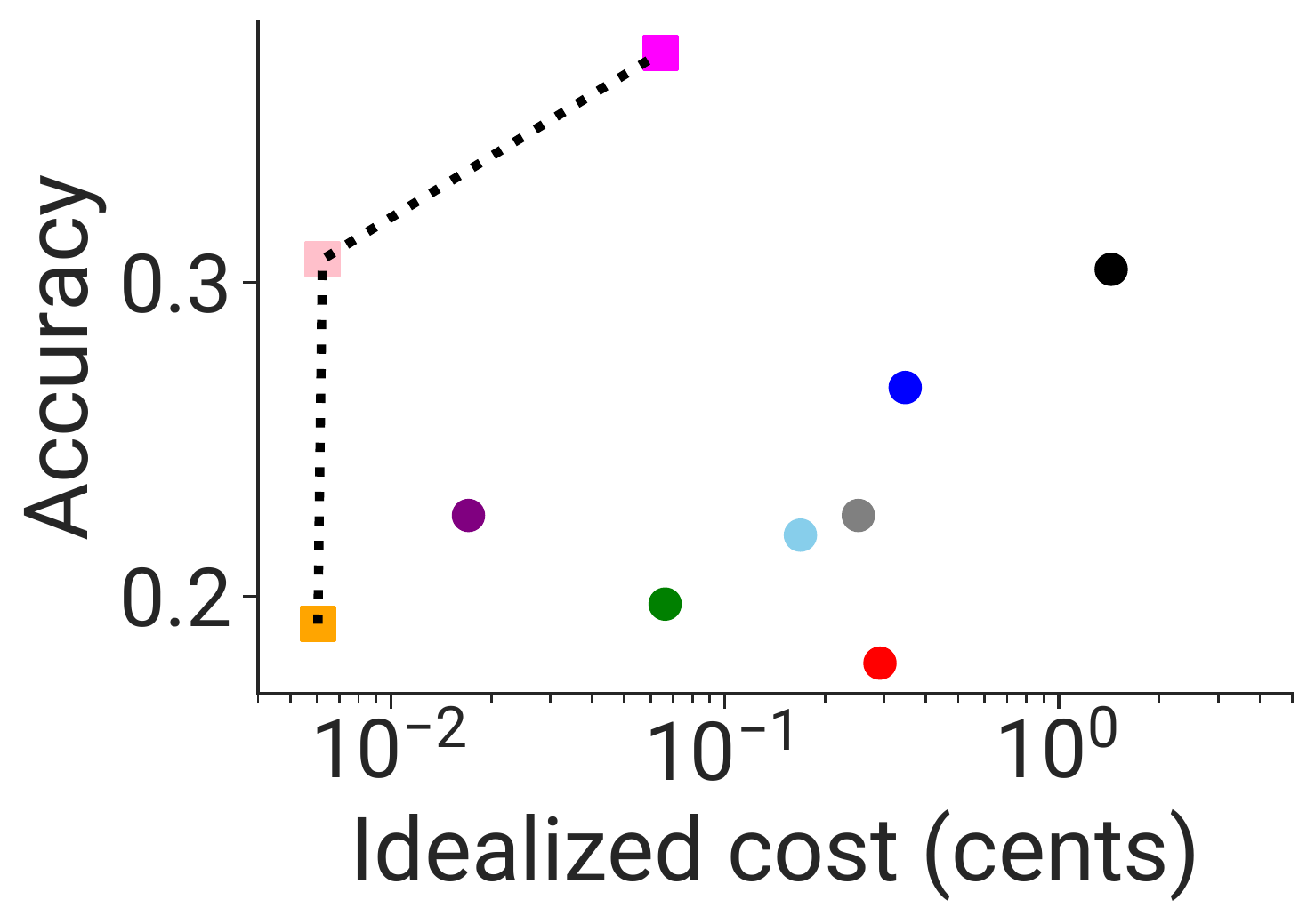}
        \caption{MMLU, college chemistry.}
    \end{subfigure}
    \centering
    \begin{subfigure}[c]{0.49\columnwidth}
        \centering
        \includegraphics[keepaspectratio=1.0,width=\columnwidth]{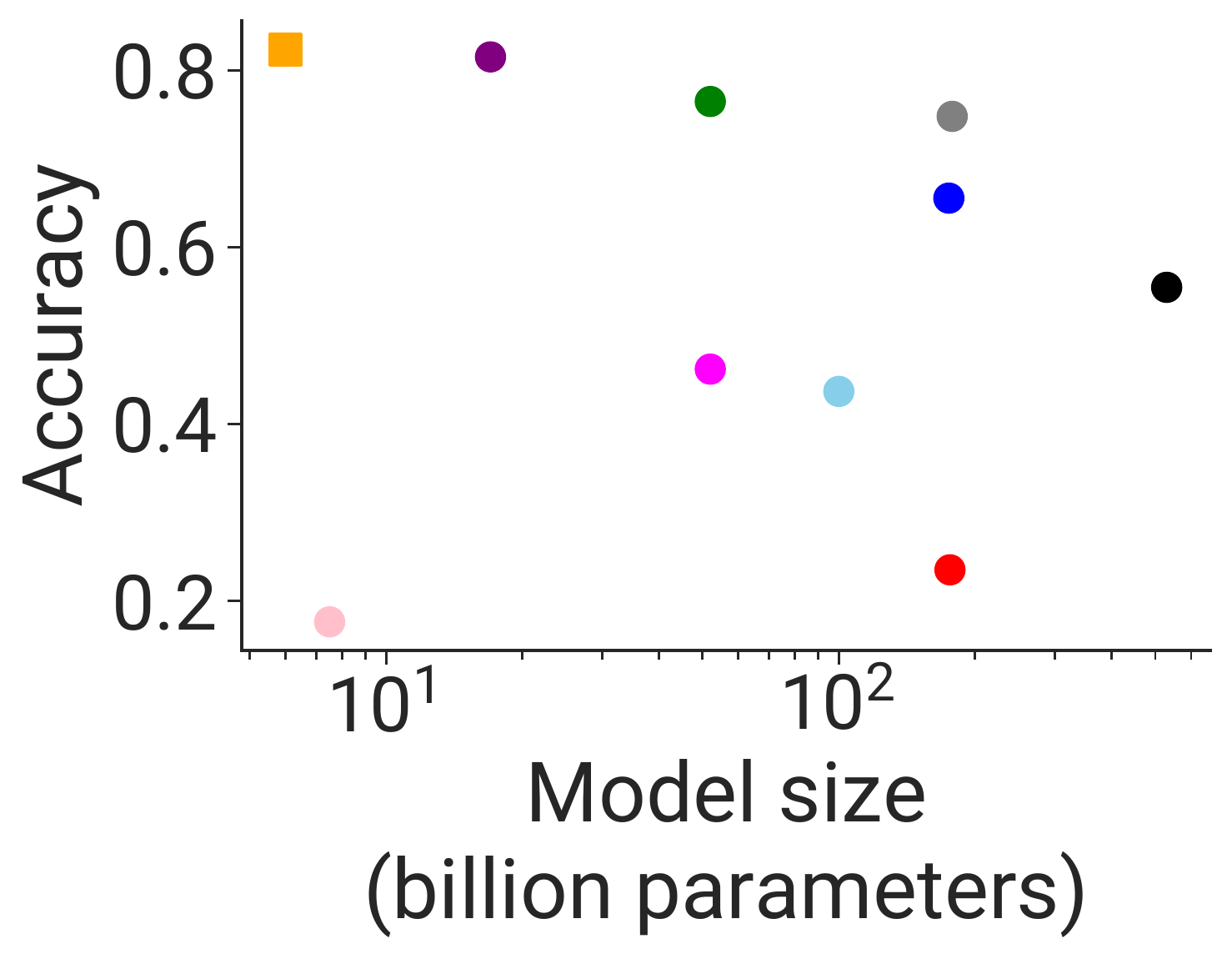}
        \includegraphics[keepaspectratio=1.0,width=\columnwidth]{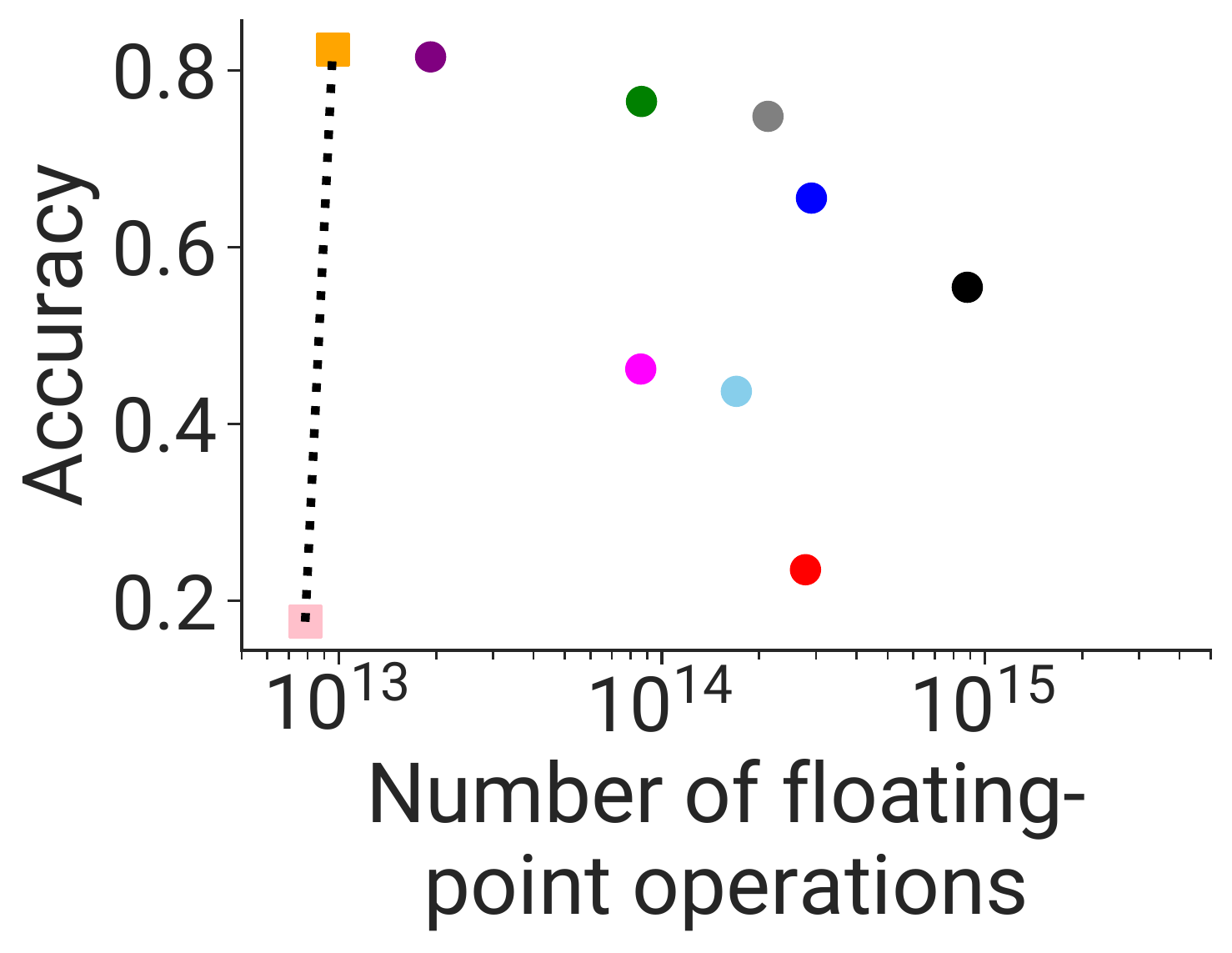}
        \includegraphics[keepaspectratio=1.0,width=\columnwidth]{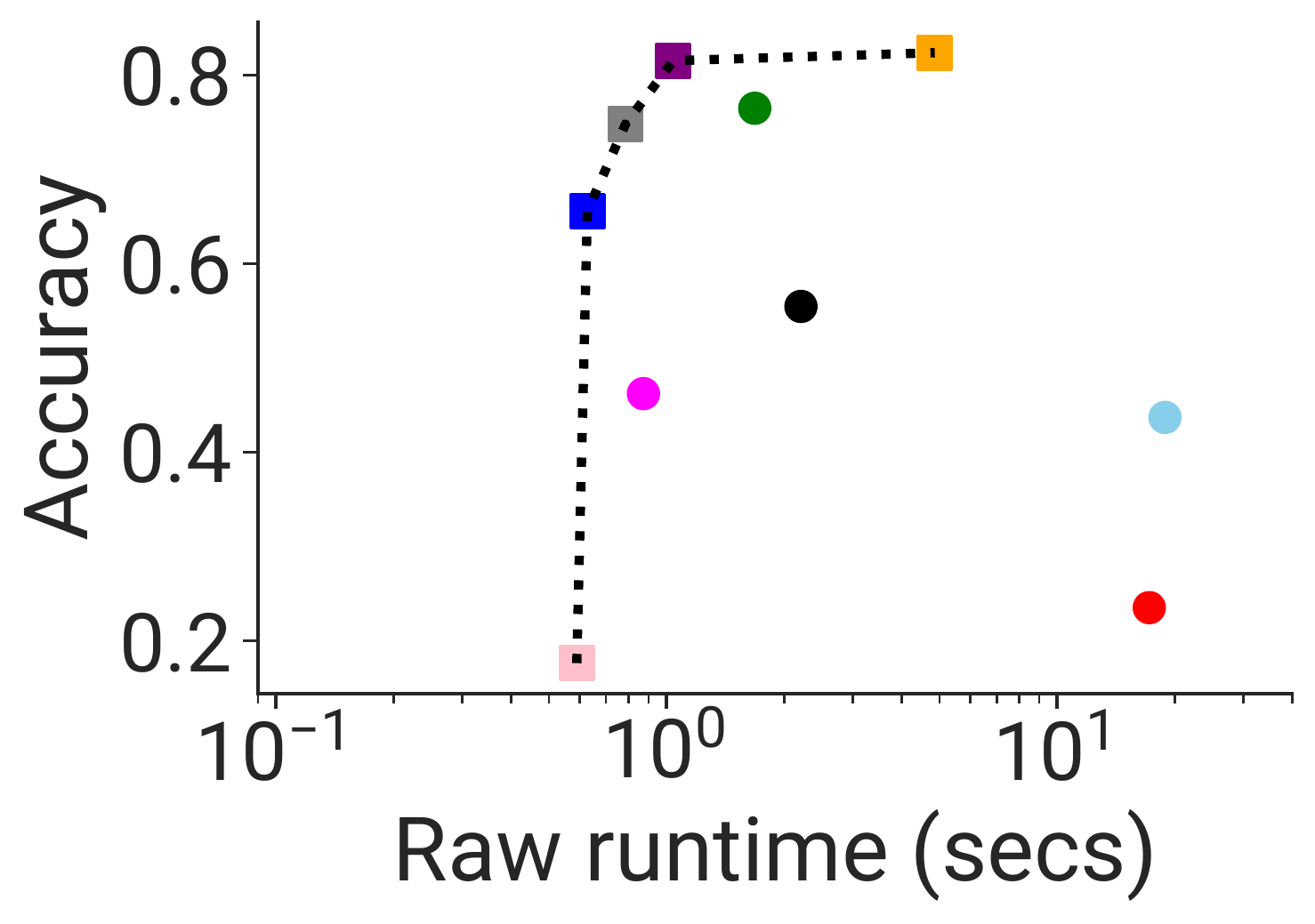}
        \includegraphics[keepaspectratio=1.0,width=\columnwidth]{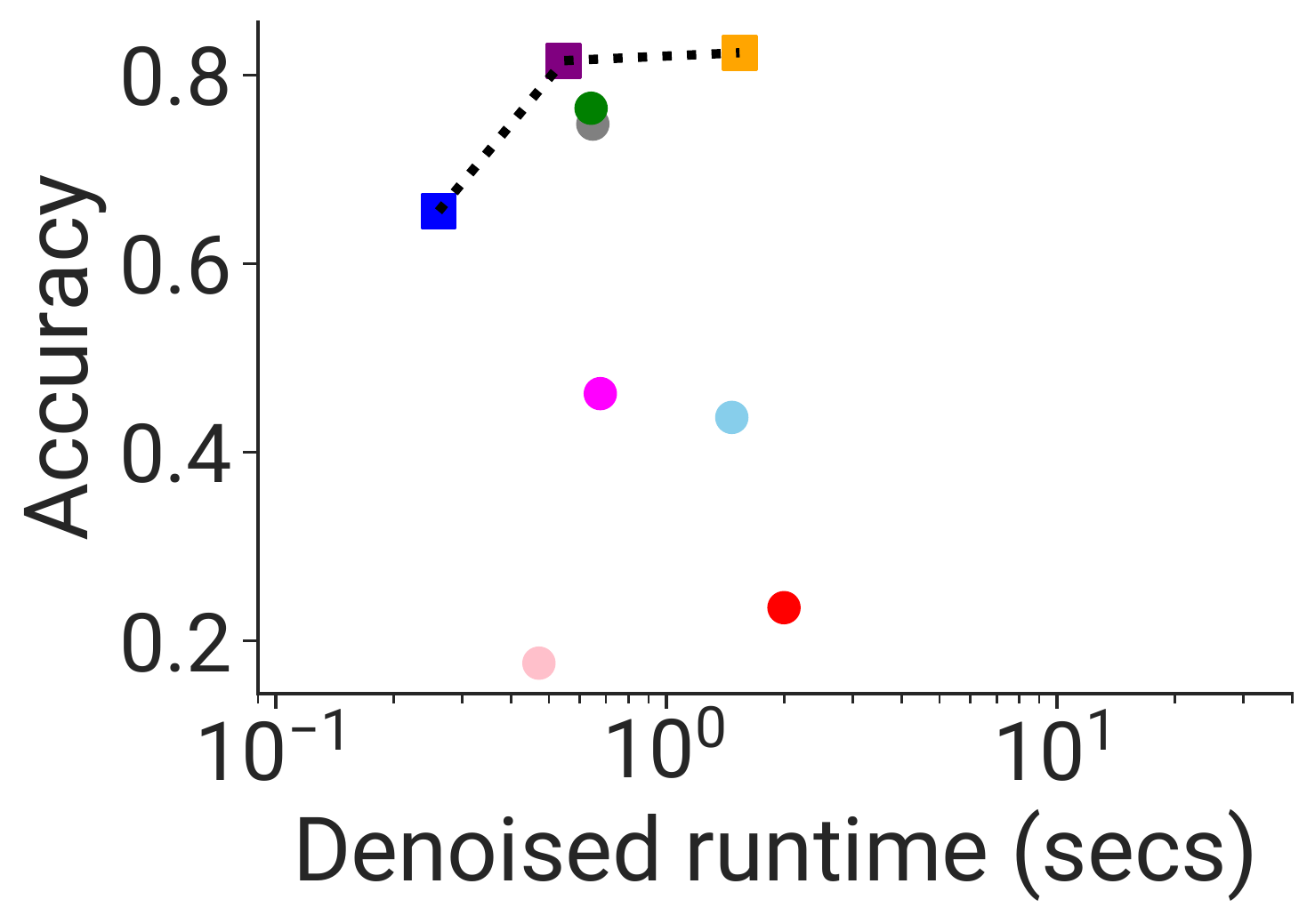}
        \includegraphics[keepaspectratio=1.0,width=\columnwidth]{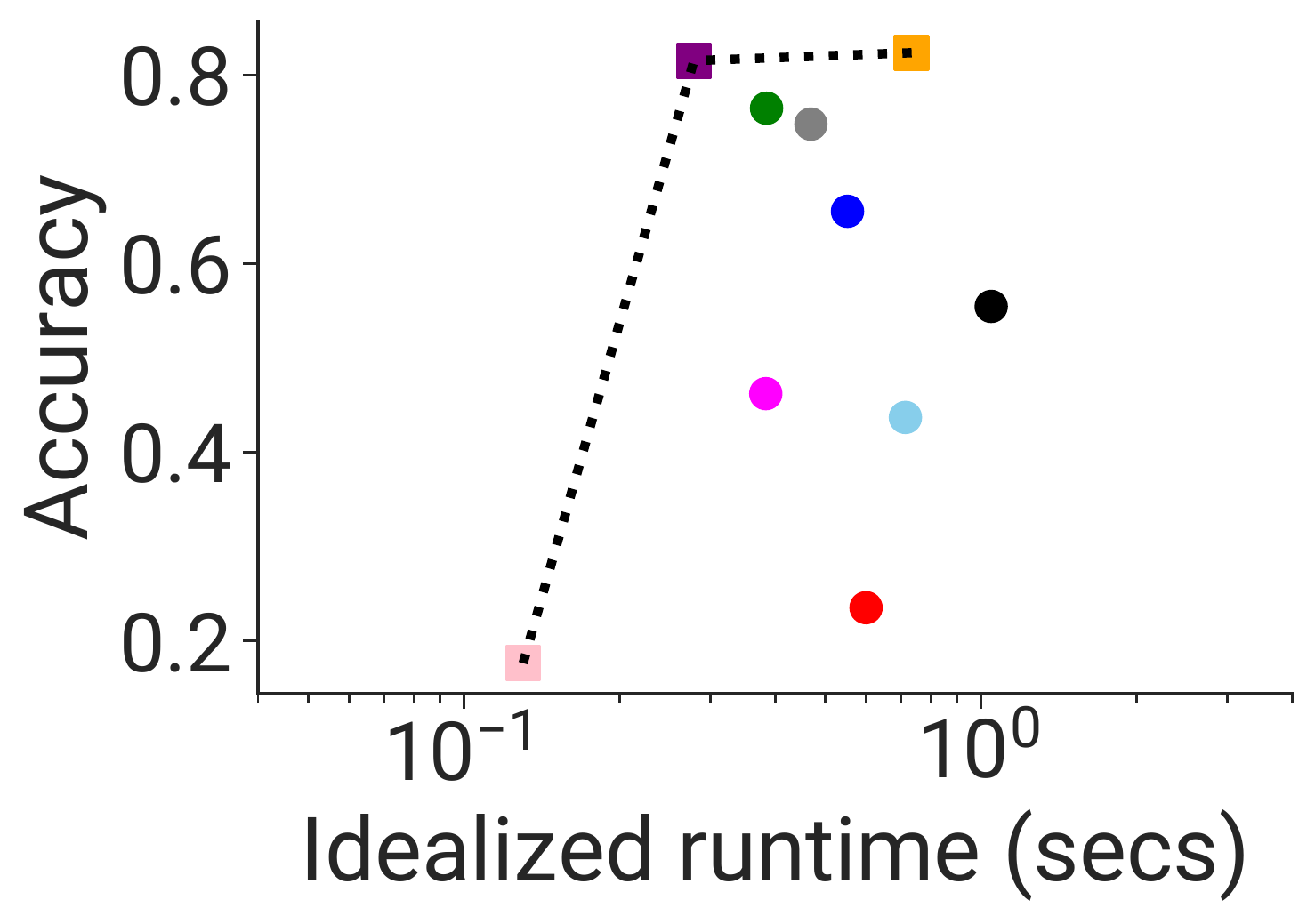}
        \includegraphics[keepaspectratio=1.0,width=\columnwidth]{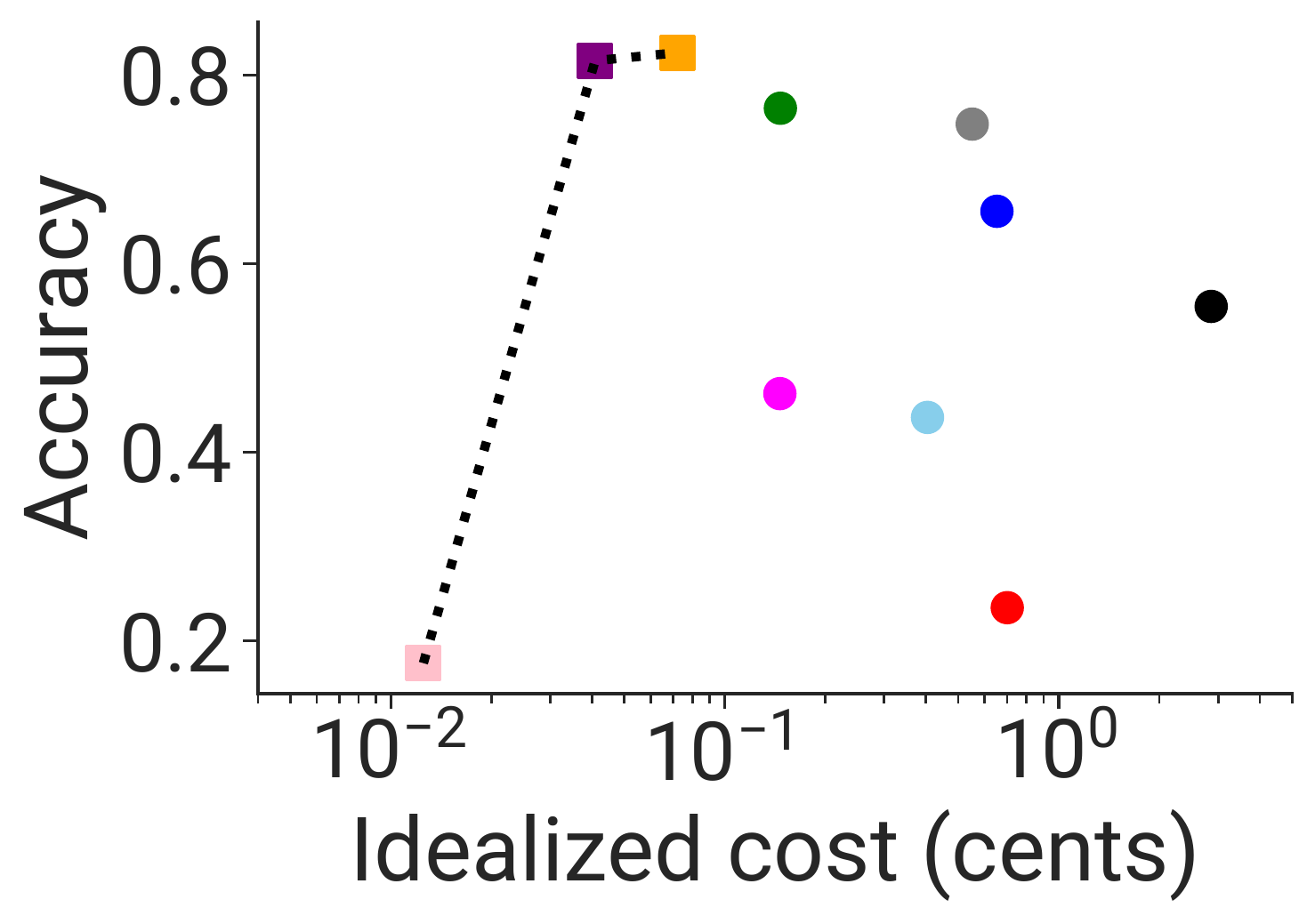}
        \caption{RAFT, terms of service.}
    \end{subfigure}
    \centering
    \begin{subfigure}[c]{0.49\columnwidth}
        \centering
        \includegraphics[keepaspectratio=1.0,width=\columnwidth]{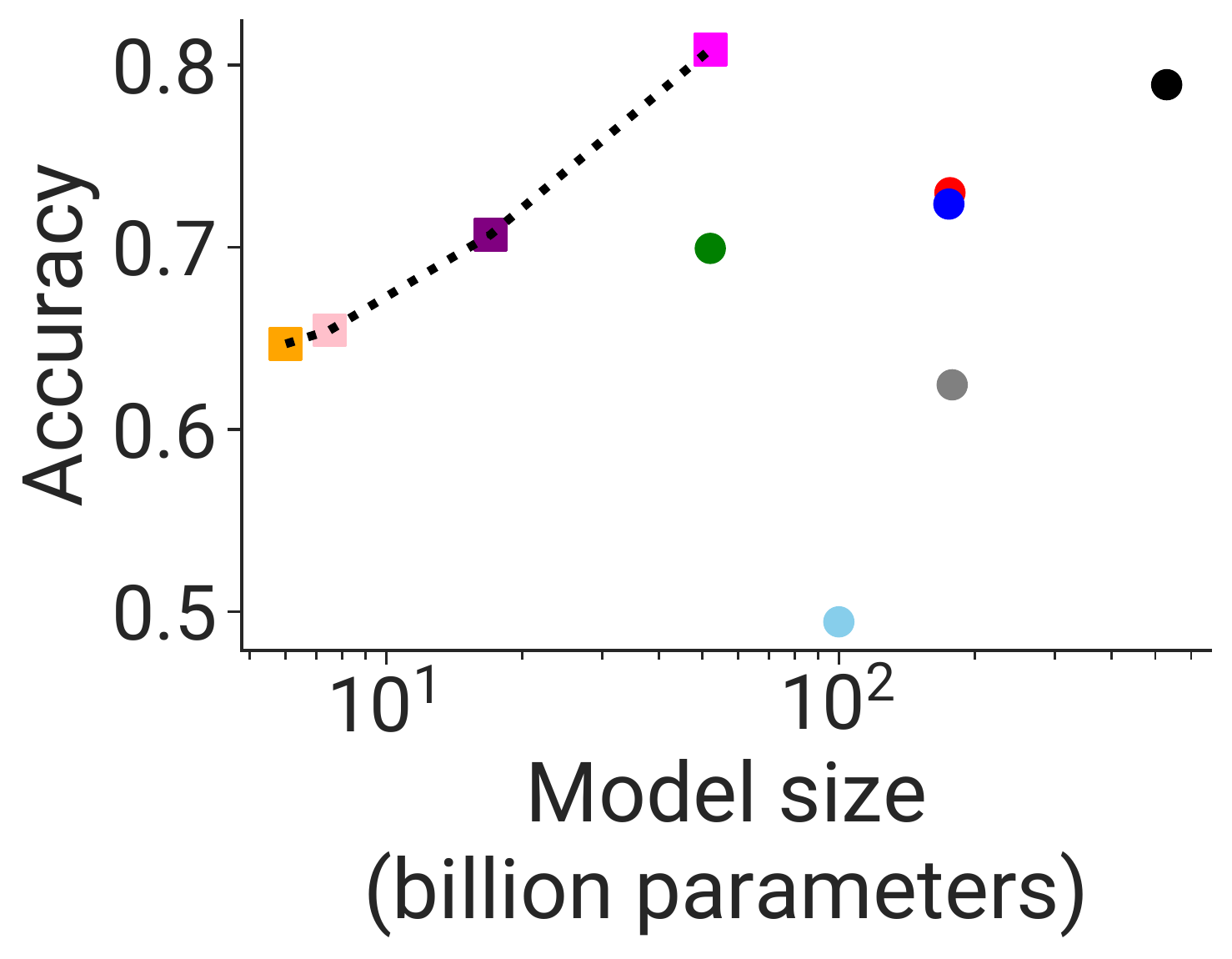}
        \includegraphics[keepaspectratio=1.0,width=\columnwidth]{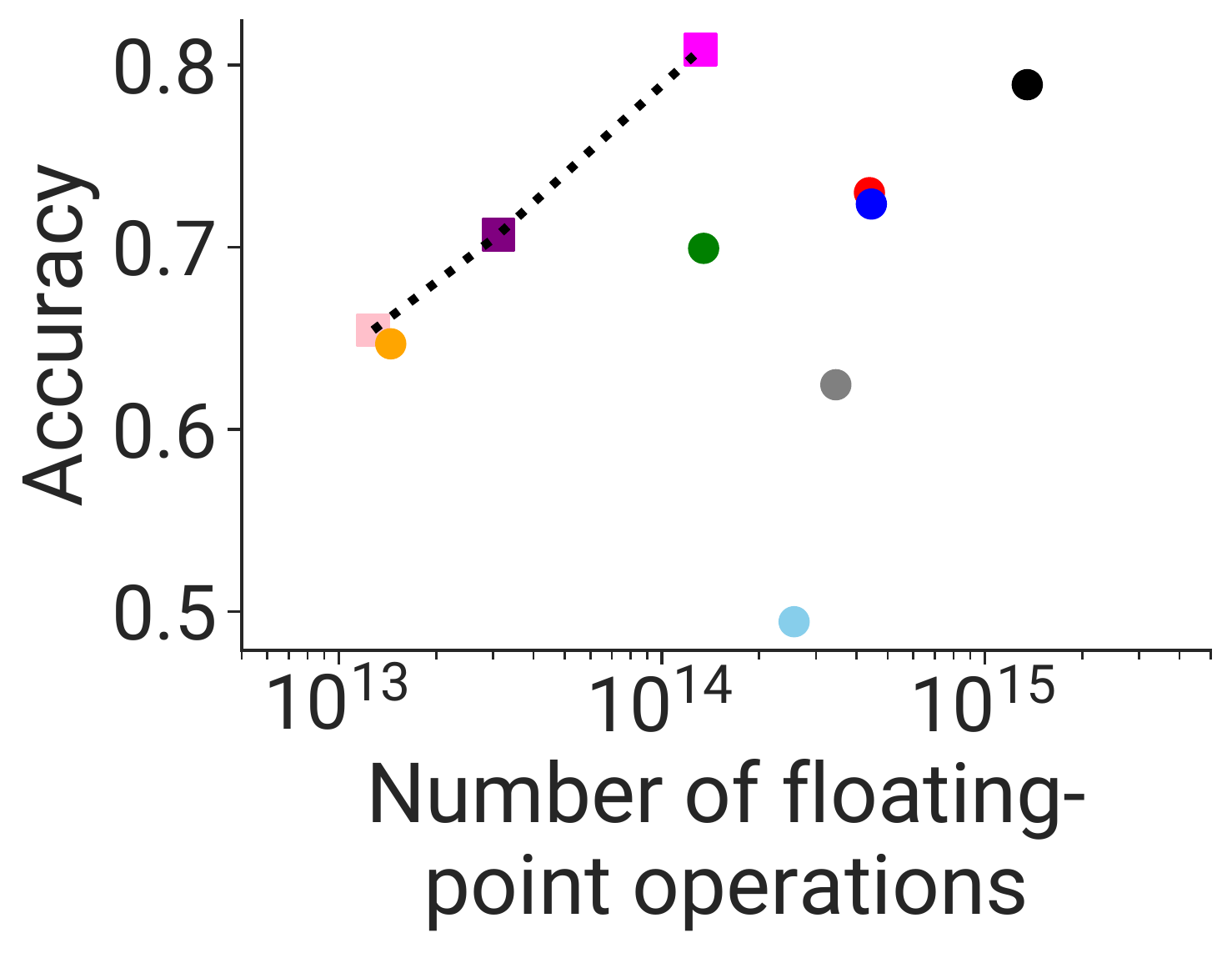}
        \includegraphics[keepaspectratio=1.0,width=\columnwidth]{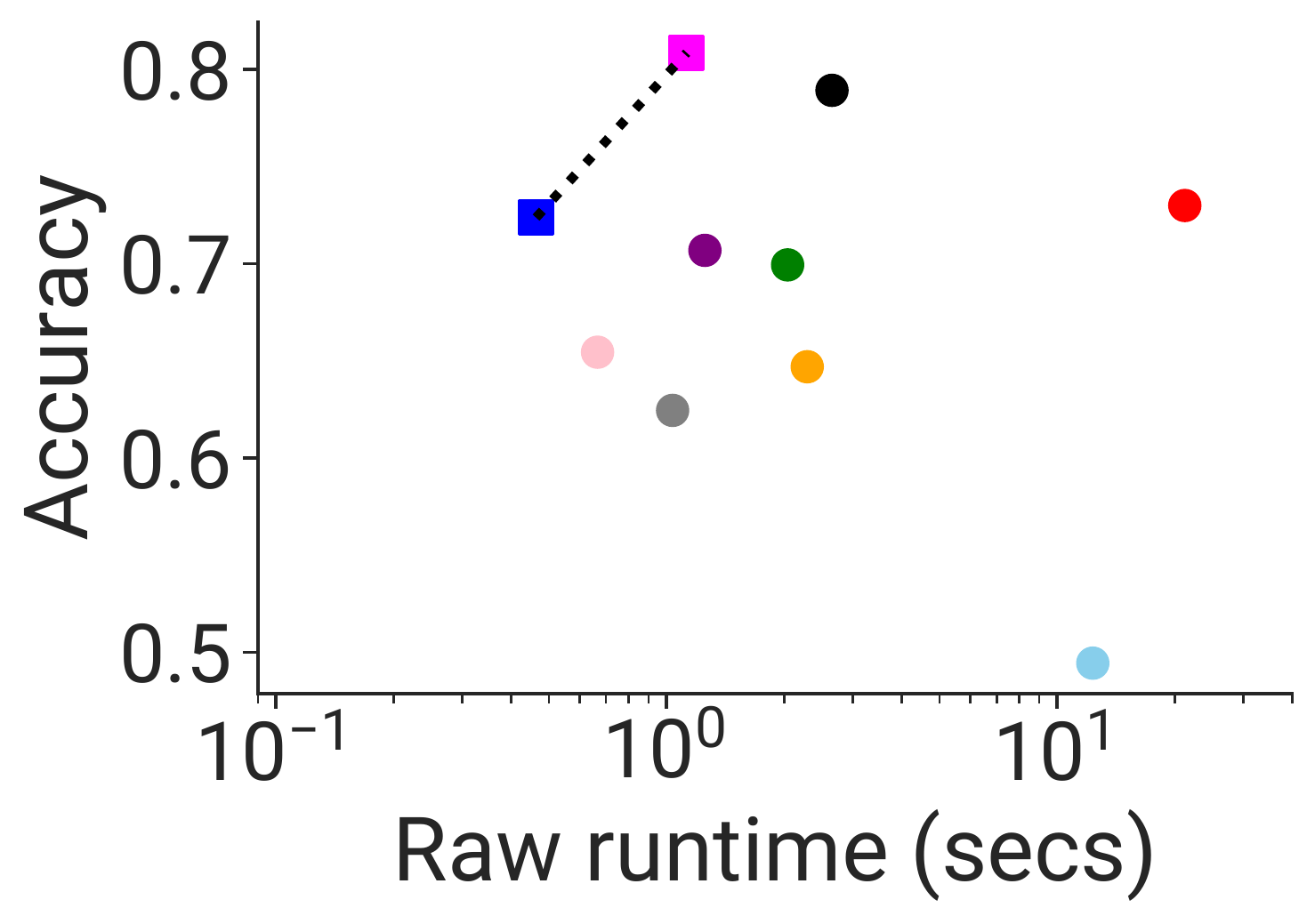}
        \includegraphics[keepaspectratio=1.0,width=\columnwidth]{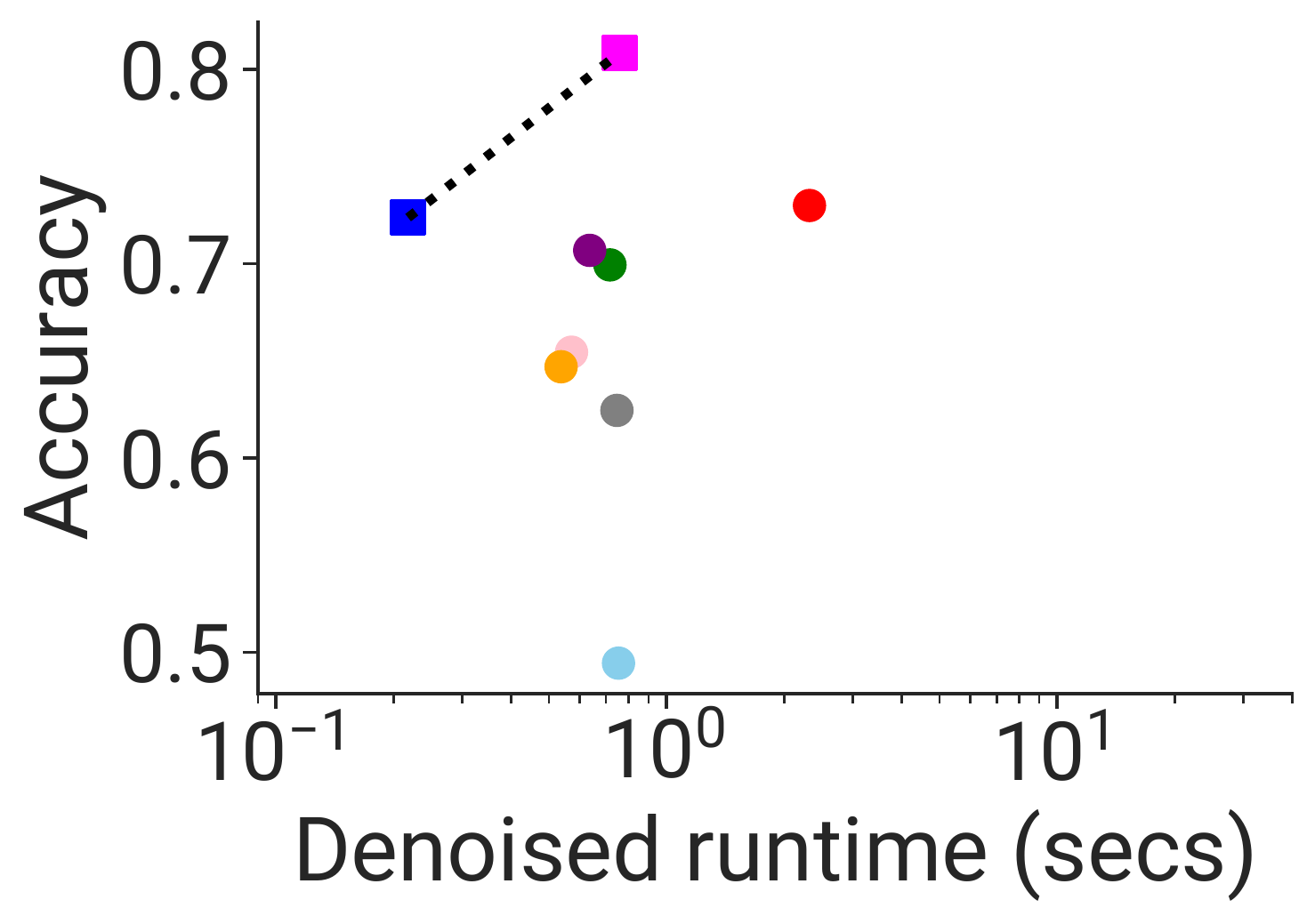}
        \includegraphics[keepaspectratio=1.0,width=\columnwidth]{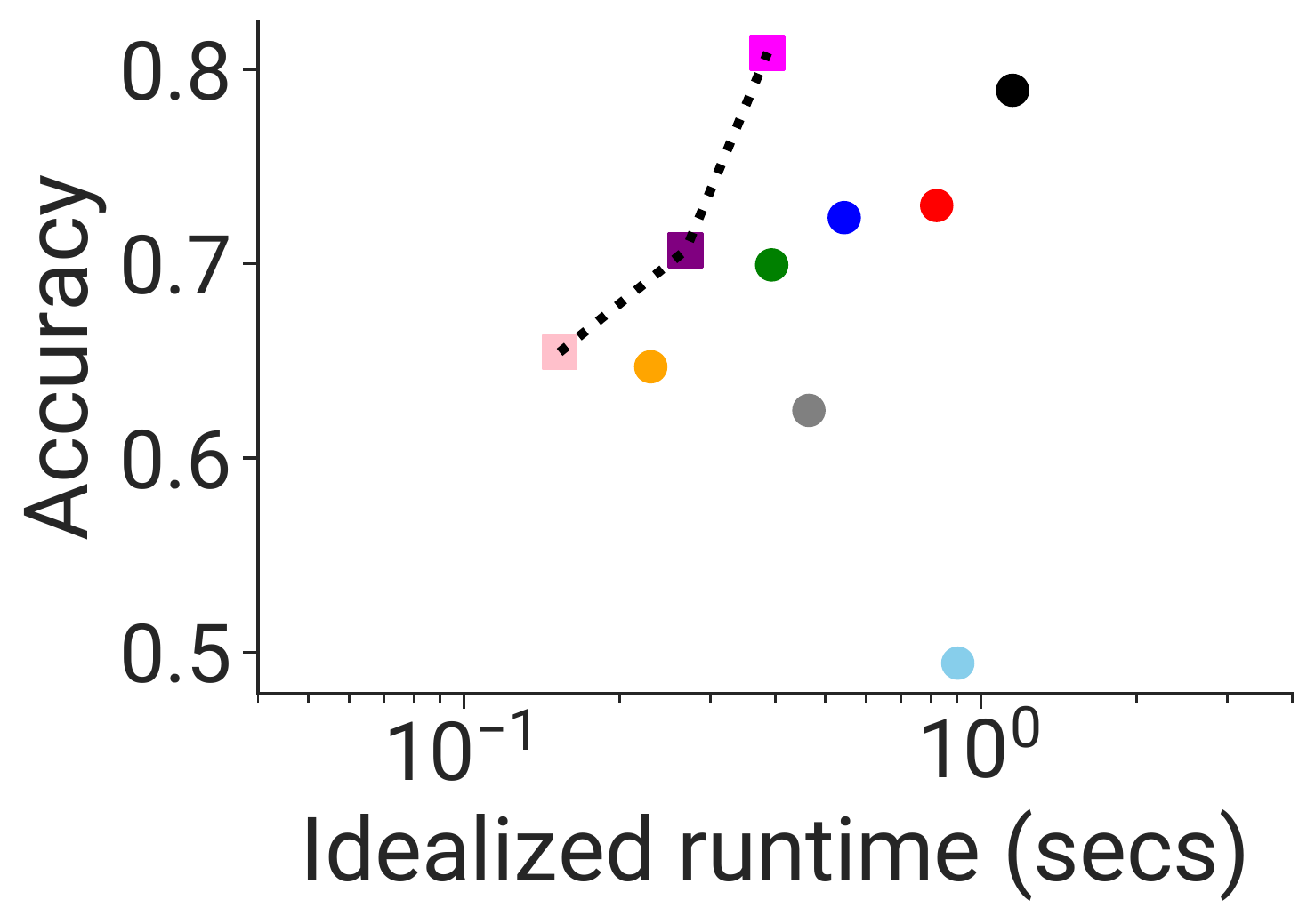}
        \includegraphics[keepaspectratio=1.0,width=\columnwidth]{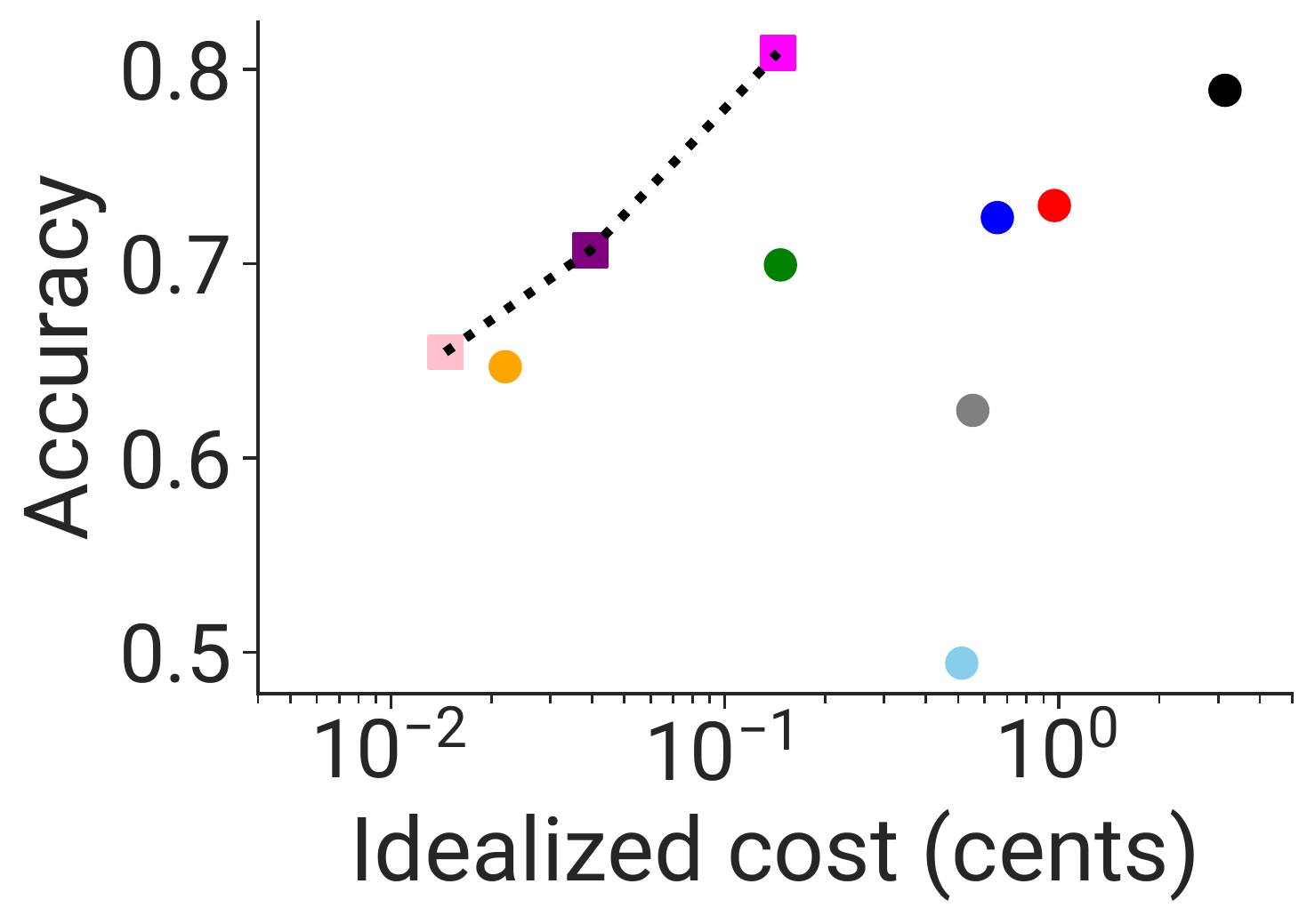}
        \caption{BoolQ.}
    \end{subfigure}
    \caption{
        Capability vs. efficiency tradeoff graphs. Capability is shown as accuracy on the target task. Six efficiency metrics are shown: model size (billions of parameters), per-query number of floating-point operations (FLOPs), raw runtime, denoised runtime, idealized runtime (all in seconds), and idealized cost (in cents). Metrics are averaged over all instances in a scenario. Models on the Pareto efficiency frontier are shown as squares with a black dotted line connecting the points (if Pareto frontier has greater than 1 point). 
    }
    \label{fig:capability_vs_accuracy}
\end{figure*}

We can now use the metrics proposed in this paper to evaluate the efficiency-capability tradeoffs of various language models accessible through black-box APIs.

We consider four diverse tasks in HELM~\cite{liang2022holistic}: a sentiment analysis task (IMDB), two question answering tasks (MMLU [college chemistry]~\cite{hendrycks2020measuring} and BoolQ~\cite{clark2019boolq}), and a classification task (RAFT [terms of service]~\cite{alex2021raft}).

Figure~\ref{fig:capability_vs_accuracy} presents the results, with each row of graphs comparing average accuracy to a different efficiency metric (model size, FLOPs, raw runtime, denoised runtime, idealized runtime, and idealized cost in order from top to bottom). Data points on the Pareto frontier of each graph are shown as squares; all other data points are shown as circles. We highlight a few takeaways.

\textbf{Effect of scale.} We observe that only a subset of the evaluated models fall on a Pareto frontier, with different models on the Pareto frontier for different tasks. This suggests that scale alone does not predict model capabilities. Scaling laws do not capture such nuances in capability differences, especially across model families; rigorous \emph{empirical} evaluation of LLMs is also needed.

\textbf{Inconsistent optimizations.} The \gptdavinci model appears in the Pareto frontier for each benchmark when using raw runtimes but not the idealized metrics. This suggests that the OpenAI API implementation is more optimized than others: this could be due to a number of factors, such as query caching or better resilience to high load. Comparing these models on a level footing (same software and hardware) requires metrics that can factor out the effect of performance optimizations orthogonal to the model, such as idealized runtime. 

\textbf{Model architecture design.} The relative positions of \bloom and \yalm on the idealized cost and FLOPS (+ model size) graphs are sometimes reversed: while \bloom achieves cheaper idealized cost (which takes into account the lower number of GPUs that \yalm requires), \yalm uses fewer floating-point operations. \bloom's improved performance can be at least partially attributed to a more thorough search through model architectures for minimum runtime with a given number of floating-point operations in the forward pass~\cite{scao2022what}.

\textbf{Run-to-run variance.} \jurassicgrande ~often achieves worse raw runtime than \jurassicjumbo ~despite having 10$\times$ fewer parameters, since the Grande model experiences higher performance variance (Figure~\ref{fig:per_instance_runtime_variance}). The idealized metrics factor out run-to-run variance, making it easier to see the true efficiency-capability tradeoffs.

\begin{figure}
    \centering
    \includegraphics[width=0.9\columnwidth]{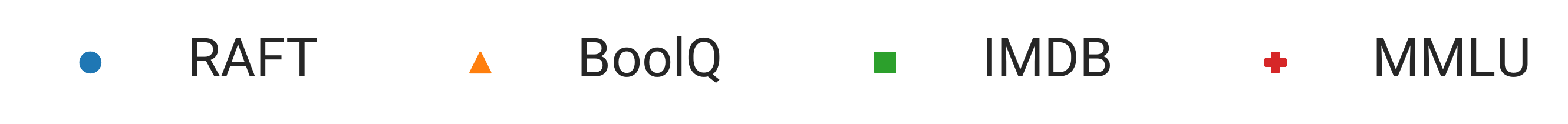}
    \begin{subfigure}{\columnwidth}
        \centering
        \includegraphics[width=0.9\columnwidth]{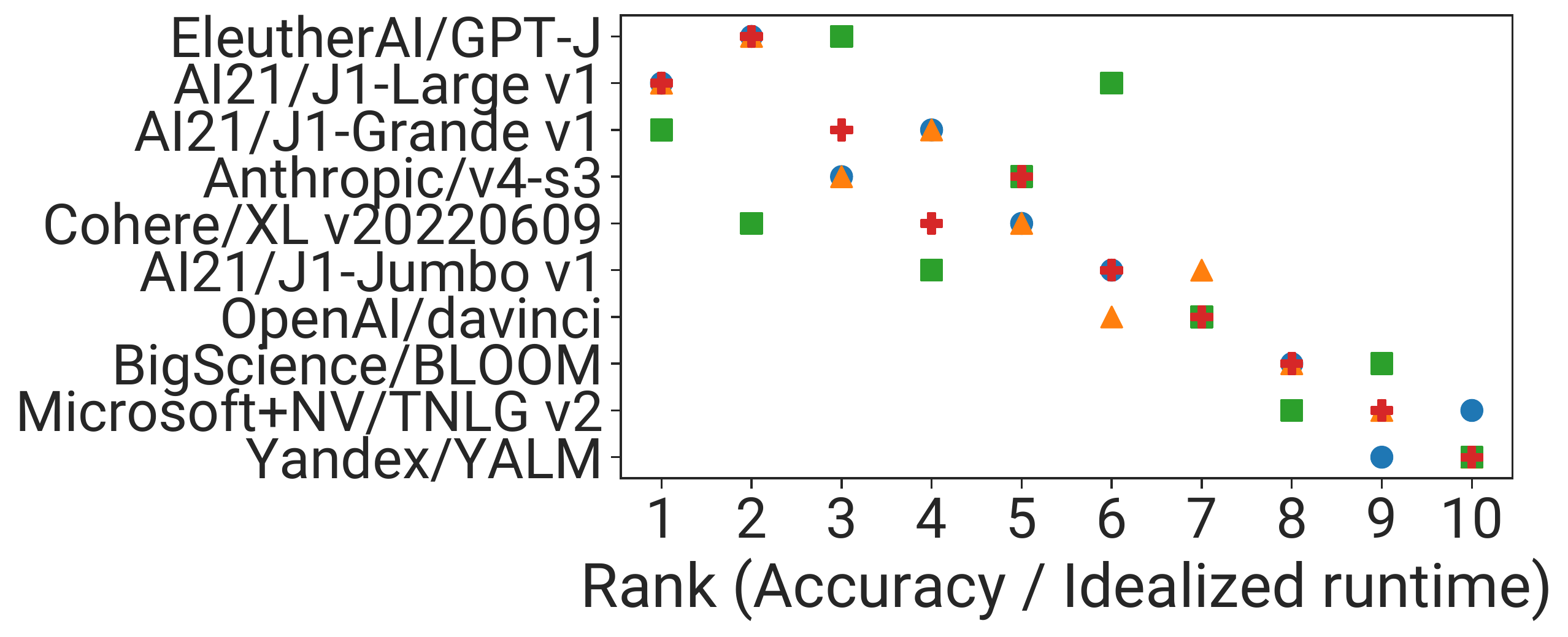}
        \caption{$f_1(\text{accuracy}, \text{idealized runtime}) = \frac{\text{accuracy}}{\text{idealized runtime}}$.}
        \label{fig:acc_vs_idealized_runtime} 
    \end{subfigure}
    \vspace{-0.05in}
    \begin{subfigure}{\columnwidth}
        \centering
        \includegraphics[width=0.9\columnwidth]{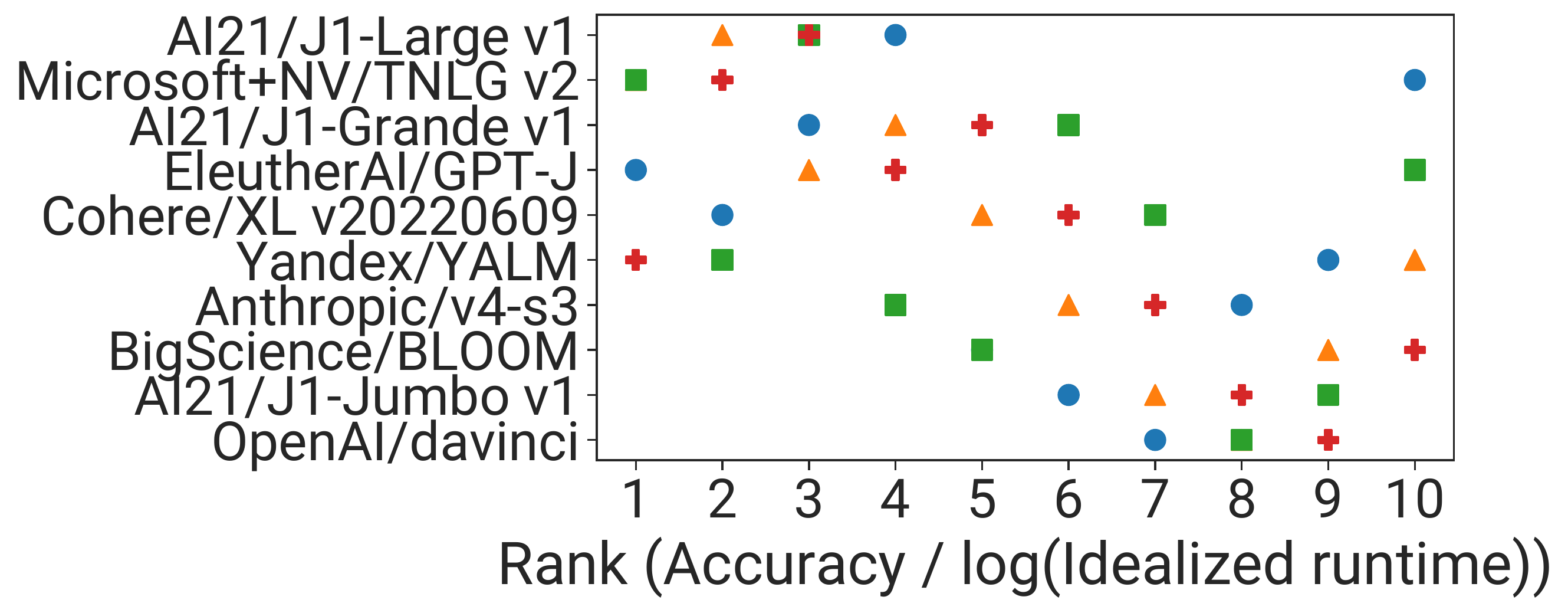}
        \caption{$f_2(\text{accuracy}, \text{idealized runtime}) = \frac{\text{accuracy}}{\log(\text{idealized runtime})}$.}
        \label{fig:acc_vs_log_idealized_runtime}
    \end{subfigure}
    \caption{Each model's relative rank when ordered by accuracy / idealized runtime across different benchmarks.}
    \label{fig:efficiency_vs_capability_ranks}
    \vspace{-0.2in}
\end{figure}

\textbf{Cost comparison.} We can also compare these esimtated \emph{inference} costs to the costs charged by the black-box API provider. We observe that they are up to an order of magnitude lower than the charged actual costs. However, we note that these reported costs \emph{do not} incorporate the significant cost of training models, which presumably gets amortized into the cost users pay with black-box APIs.

\textbf{Variation in relative performance.} We can use objective functions combining accuracy and an inference efficiency metric to rank the models. Figure~\ref{fig:efficiency_vs_capability_ranks} plots the rank of each model using an objective function $f_1(\text{accuracy}, \text{idealized runtime}) = \frac{\text{accuracy}}{\text{idealized runtime}}$. For this particular objective, we observe that each model achieves similar ranks across benchmarks. However, a modified objective function $f_2(\text{accuracy}, \text{idealized runtime}) = \frac{\text{accuracy}}{\log(\text{idealized runtime})}$ increases variation across benchmarks and impacts relative model ordering (e.g., \mtnlg's average rank significantly improves) as this objective de-emphasizes the importance of inference efficiency compared to accuracy. None of the models we evaluated dominated across scenarios and objective functions. Studying the implications of different objective functions in detail is interesting future work.

\section{Related Work}

A large body of work has looked at studying the impact of model scale on model capabilities along different dimensions. We summarize this work here.

\textbf{Scaling laws and other benchmarking efforts.}
Recent work has proposed ``scaling laws''~\cite{kaplan2020scaling}, which show how model size affects the training and validation loss of these models by fitting a curve to dozens of training runs.
While these scaling laws are instructive, we also care about the capabilities of models along other axes beyond validation loss (e.g., are models robust; do they exhibit stereotypes?).
Moreover, large language models have been shown to exhibit \emph{emergent behavior} that cannot easily be expressed as a continuous function of scale~\cite{wei2022emergent}.
Similarly, even though model size is used as a proxy for both training and inference runtime performance, it is not interpretable when trying to answer questions like ``Can model $X$ meet a latency SLO of 100 milliseconds?'' or ``How much will it cost to use model $X$ in this concrete application with the following characteristics?''.
Consequently, we need to fall back on empirical analysis to fully understand the capabilities of these models.

Various empirical analyses that focus on quantifying the capabilities of LLMs along various dimensions, including papers introducing new models like PaLM~\cite{chowdhery2022palm} and Gopher~\cite{rae2021scaling}, and more ambitious benchmarking efforts like BIG-bench~\cite{srivastava2022beyond}, also use model size as a proxy for scale and runtime performance. These comparisons are often useful within a family of models (e.g., OpenAI Instruct series of models), but are less useful when trying to compare model families.

\textbf{Floating-point operations and other proxy metrics for efficiency.}
The number of floating-point operations (FLOPs) required for the forward pass of a model has also often been used to estimate inference efficiency.
While this is a fine approximation, it is not ideal for a couple different reasons.
First, runtime does not correlate exactly with the number of FLOPs required~\cite{scao2022what}. In particular, two operators with the same number of FLOPs could be executed with different efficiencies if one of the operators involves more memory accesses, preventing execution at peak device throughput.
Second, as with model size, the number of FLOPs is hard to interpret. LLMs are often part of larger applications, and the performance requirements of these applications impose runtime constraints on LLM inference. It is hard to translate FLOPs to something actionable.

Similarly, even though raw runtime from black-box APIs accurately represents the behavior API consumers observe, it has various issues as outlined earlier: black-box APIs can run models on unknown hardware and can be subject to performance contention. We show quantitatively that both of these metrics can lead to incorrect conclusions when examining fundamental efficiency-capability tradeoffs (\S\ref{sec:tradeoffs}).

\textbf{Inference runtime estimation for other types of models.}
Typically, inference for ML models is straightforward: an input of a particular size is passed through the model, in the process generating intermediate outputs and eventually a final prediction from a \emph{single forward pass}. The sizes of intermediate outputs do not change from input to input, resulting in negligible runtime variance. This consequently makes inference runtime estimation easy. However, LLMs are different: while the hidden size does not change from input to input, the prompt size (in number of tokens) can be different for various inputs. Additionally, with token-at-a-time generation, inference happens in two phases, with multiple forward passes often needed depending on the number of output tokens generated. This makes runtime estimation in this setting much more challenging.

\textbf{Carbon costs of ML computation.}
Many papers~\cite{canziani2016analysis, cao2020towards, henderson2020towards, strubell2019energy, bender2021dangers, patterson2021carbon} have discussed the importance of quantifying the cost of training models, both from an energy and emitted CO$_\text{2}$ perspective. This is often possible because model providers are open about details on training necessary to compute these metrics~\cite{black2022gpt, patterson2021carbon}. While recent work has emphasized the need for considering inference-time efficiency~\citep[][\S5.3]{henderson2020towards,bommasani2021opportunities}, information on inference-time costs of LLMs is more scant 
for a multitude of reasons (e.g., runtime performance of a black-box API might be part of a company's competitive advantage). 
This makes it harder to measure such metrics without some assumptions as well as profiling, as demonstrated in our work.
\section{Conclusion}

This work presents a systematic study of inference efficiency for autoregressive Transformer models accessible through black-box APIs. We showed both analytically and empirically that the inference runtime for these models is the sum of a piecewise linear function of the prompt size and a linear function of the number of output tokens, and designed a new idealized runtime metric that can be estimated efficiently with minimal extra profiling.

We are hopeful that our work provides a step forward in consistent and comparable analyses of efficiency-capability trade-offs for Transformer models served via black-box APIs, and helps model creators make better informed decisions about long-term model investments, considering both training and inference costs.

\bibliography{main}
\bibliographystyle{mlsys2023}

\clearpage
\appendix
\section{Operators in Transformer Layer}

We use the same notation as before: $b$ is the microbatch size (number of sequences) and $h$ is the hidden size of the model. In practice, the self-attention layer computation described in \S\ref{sec:transformer_computation_training} is performed with different parameter matrices $W_i^K$, $W_i^V$ and $W_i^Q$. This is called running the self-attention layer with multiple \emph{attention heads}. We assume that the Transformer model has $n$ attention heads.

\subsection{Training}
\label{sec:transformer_operators_in_training}

$s$ is the sequence length in terms of number of tokens. Inputs $X$ to the Transformer layer have shape $(b, s, h)$. The Transformer layer's computation during training can then be reduced to the following matrix multiplication operations.
\begin{itemize}
    \item Attention key, value, query transformations: These can be expressed as a single matrix multiplication of size: $(bs, h) \times (h, 3h)$. Output is of size $(bs, 3h)$.
    \item Attention score computation: $bn$ batched matrix multiplications (BMMs), each of size $(s, h/n) \times (h/n, s)$. Output is of size $(bn, s, s)$.
    \item Attention over value computation: $bn$ batched matrix multiplications of size $(s, s) \times (s, h/n)$. Output is of size $(bn, s, h/n)$.
    \item Post-attention linear projection: a single matrix multiplication of size $(bs, h) \times (h, h)$ to coalesce outputs of $n$ attention heads to a single per-sequence vector of size $h$. Output is of total size $(bs, h)$.
    \item Matrix multiplications in the MLP layer of size $(bs, h) \times (h, 4h)$ and $(bs, 4h) \times (4h, h)$. Outputs are of size $(bs, 4h)$ and $(bs, h)$.
\end{itemize}

Using the fact that a $(m, n) \times (n, k)$ matrix multiplication needs $2mnk$ floating-point operations, the total number of compute operations is to complete the forward pass through a Transformer layer during training is $24bsh^2\left(1 + \frac{s}{6h}\right)$. A Transformer model typically has $l$ Transformer layers, resulting in a total of $24bsh^2l\left(1 + \frac{s}{6h}\right)$ floating-point operations.

\subsection{Autoregressive Inference}
\label{sec:transformer_operators_in_inference}

We can similarly compute the number of floating-point operations needed to generate a single output token during autoregressive inference. $i$ is the number of tokens generated so far (i.e., the $(i+1)^\text{th}$ token, including the prompt, needs to be generated next). The operators to be run in each Transformer layer in this phase are:
\begin{itemize}
    \item Attention key ($K$), value ($V$), query ($Q$) transformations: These can be expressed as a single matrix multiplication of size $(b, h) \times (h, 3h)$.
    \item Attention score computation: $bn$ batched matrix multiplications (BMMs), each of size $(1, h/n) \times (h/n, i)$ (only $Q$ value for the latest token is used; $K$ and $V$ values accumulated over all tokens so far).
    \item Attention over value computation: $bn$ batched matrix multiplication of size $(1, i) \times (i, h/n)$.
    \item Post-attention linear projection: a single matrix multiplication of size $(b, h) \times (h, h)$.
    \item Matrix multiplications in the MLP layer of size $(b, h) \times (h, 4h)$ and $(b, 4h) \times (4h, h)$.
\end{itemize}

Consequently, the total number of compute operations needed to generate the $(i+1)^\text{th}$ token is $24bh^2l + 4bihl = 24bh^2l\left(1 + \frac{i}{6h}\right)$.

\end{document}